\documentclass{article}

\usepackage{microtype}
\usepackage{graphicx}
\usepackage{booktabs}
\usepackage{array}
\usepackage[table]{xcolor} % table option enables \rowcolor (used to shade our-method rows)
\usepackage{xspace}
\usepackage{pifont}
\usepackage{makecell}

\usepackage{booktabs} % for professional tables
\usepackage{multirow}
\usepackage{subcaption}
\usepackage{siunitx}
\usepackage{amsmath}
\usepackage{amssymb}
\usepackage{url}
\usepackage{float}
\usepackage{placeins}
\usepackage{tikz}
\usepackage{hyperref}

\usepackage{hyperref}

\usepackage[accepted]{mlsys2025}
\makeatletter
\renewcommand{\mlsys@appearing}{Preprint.}
\makeatother
\mlsystitlerunning{Failure Prediction and Classification in Vision-Language-Action Models Via Calibrated Latent Probing}

\newcommand{\ours}{\textsc{VLA-Tug}\xspace} %Name of framework
\newcommand{\tprearly}{\mathrm{TPR}_{\mathrm{early}}}
\newcommand{\tfail}{t_{\mathrm{fail}}}
\newcommand{\tdet}{t_{\mathrm{det}}}
\newcommand{\fmode}{f_{\mathrm{mode}}}
\newcommand{\Dcal}{\mathcal{D}_{\mathrm{cal}}}
\usepackage[most]{tcolorbox}
\usepackage{xspace}

\definecolor{taskOrange}{RGB}{222,142,70}
\definecolor{sigGreen}{RGB}{74,140,100}
\definecolor{skillNavy}{RGB}{62,84,150}

\newcommand{\newtask}{{\textcolor{taskOrange}{\footnotesize \textsc{new-task}\xspace}}}
\newcommand{\newsig}{{\textcolor{sigGreen}{\footnotesize \textsc{new-signature}\xspace}}}
\newcommand{\newskill}{{\textcolor{skillNavy}{\footnotesize \textsc{new-skill}\xspace}}}

\newtcolorbox{takeawaybox}{
    enhanced,
    colback=red!4!white,
    colframe=red!70!black,
    boxrule=0pt,
    titlerule=0pt,
    leftrule=2.5pt,
    rightrule=2.5pt,
    top=2pt, bottom=2pt, left=6pt, right=6pt,
    fonttitle=\bfseries,
    title=Takeaways,
    coltitle=black,
    colbacktitle=red!4!white,
    boxsep=2pt,
    arc=0pt,
    outer arc=0pt,
    before skip=8pt,
    after skip=8pt
}
\newtcolorbox{practicebox}{
    enhanced,
    colback=blue!4!white,
    colframe=blue!70!black,
    boxrule=0pt,
    titlerule=0pt,
    leftrule=2.5pt,
    rightrule=2.5pt,
    top=2pt, bottom=2pt, left=6pt, right=6pt,
    fonttitle=\bfseries,
    title=Practical Lessons,
    coltitle=black,
    colbacktitle=blue!4!white,
    boxsep=2pt,
    arc=0pt,
    outer arc=0pt,
    before skip=8pt,
    after skip=8pt
}

\newtcolorbox{promptbox}[1][]{
    colback=gray!10,
    colframe=gray!50,
    fonttitle=\bfseries\small,
    title=#1,
    left=6pt, right=6pt, top=2pt, bottom=2pt,
}

\begin{document}

\twocolumn[

% MLSYS VERSION
% \mlsystitle{%
%   \centering
%   \begin{tabular}[c]{@{}c@{\hspace{1.5em}}l@{}}
%     \multirow{3}{*}{\includegraphics[height=3.2em]{Figures/xkcd-vla-tug.png}} & 
%     \ours{}: Anticipating When and How \\ 
%     & Vision-Language-Action Models Fail \\ 
%     & via Calibrated Latent Probing
%   \end{tabular}%
% }

% ARXIV VERSION
\mlsystitle{%
  \centering
  \begin{tabular}[c]{@{}c@{\hspace{1.5em}}l@{}}
    \multirow{3}{*}{\includegraphics[height=3.2em]{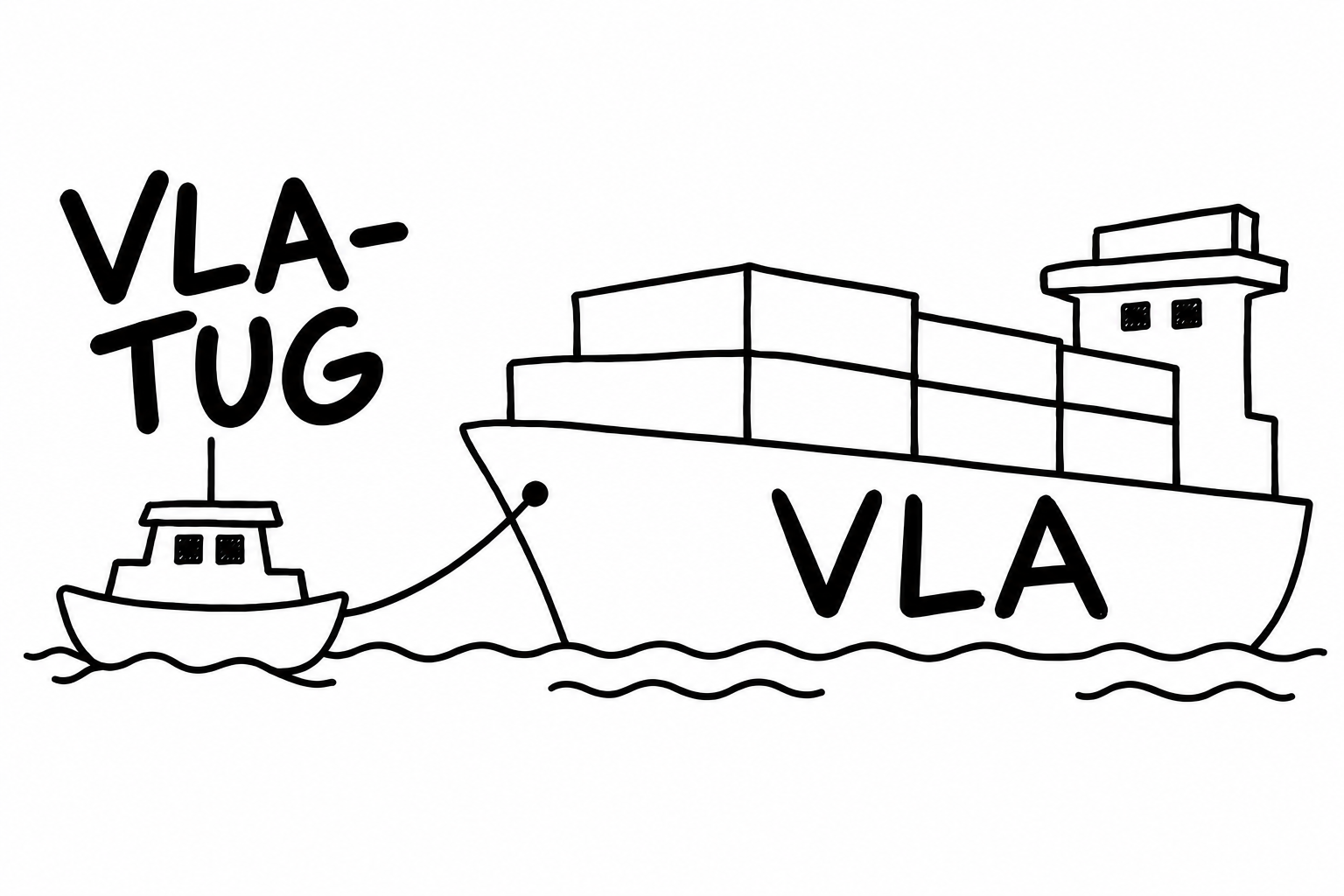}} & 
    Failure Prediction and Classification
      \\ & in Vision-Language-Action Models  \\
    & via Calibrated Latent Probing
  \end{tabular}%
}

% Author information is hidden automatically in the blind (non-[accepted]) version.
% Fill in the list below for the camera-ready version.
% \begin{mlsysauthorlist}
% \mlsysauthor{First Author}{inst}
% \end{mlsysauthorlist}
% \mlsysaffiliation{inst}{Department, University, City, Country}
% [ARXIV-EDIT START] authors added for arXiv version (original line below kept commented)
\begin{mlsysauthorlist}
\mlsysauthor{Michal P. Podolinsky}{inst1}
\mlsysauthor{Yunhao Yang}{inst1}
\mlsysauthor{Daniel Little}{inst2}
\mlsysauthor{Hong Lu}{inst2}
\mlsysauthor{Rohan Siva}{inst1}
\mlsysauthor{Olin Goudey}{inst2}
\mlsysauthor{Neel P. Bhatt}{inst4}
\mlsysauthor{Matthias Scheutz}{inst2}
\mlsysauthor{Ufuk Topcu}{inst1}
\end{mlsysauthorlist}
\mlsysaffiliation{inst1}{University of Texas at Austin}
\mlsysaffiliation{inst2}{Tufts University}
% \mlsysaffiliation{inst3}{Tufts University, Department of Computer Science}
\mlsysaffiliation{inst4}{University of Texas at Dallas}
\mlsyscorrespondingauthor{Michal P. Podolinsky}{michal.podolinsky@utexas.edu}
% ORIGINAL: \mlsyscorrespondingauthor{Anonymous}{anonymous@example.com}
% [ARXIV-EDIT END]

\mlsyskeywords{Machine Learning, MLSys, Vision-Language-Action Models, Failure Detection}

\vskip 0.3in

\begin{abstract}
Vision-Language-Action models (VLAs) enable general-purpose embodied intelligence, but their performance is brittle under distribution shift, where unanticipated failures can damage equipment or endanger people. Deploying VLAs in real-world systems therefore requires a runtime monitor that is \emph{timely}, \emph{calibrated}, and \emph{informative} while being \emph{efficient} enough to run on the edge. Existing methods fail to fulfill these requirements. We present \ours{}, a lightweight probe over the latent layers of a VLA backbone that improves early failure detection and bounds the missed-detection and false-alarm rates, with minimal inference overhead. We show that supervising on the pre-failure segment of a trajectory increases early failure detection at matched false-positive rates by up to $30\%$ over the strongest learned baseline. Under shift to unseen tasks, the bounds on missed-detection and false-alarm rates remain valid in up to $86\%$ of runs, versus at most $21\%$ for baselines. Unlike existing methods, \ours{} predicts the failure mode to enable corrective actions, raising top-3 accuracy on unseen tasks by up to \textcolor{black}{$44\%$} over a majority baseline. \ours{} enables real-time monitoring where the well-timed and informative probe supports lightweight steering that improves task success. We release a dataset of $3{,}180$ annotated manipulation episodes to support future work on latent failure detection in VLAs at \href{https://vla-tug.github.io}{vla-tug.github.io}.
\end{abstract}
]

\printAffiliationsAndNotice{}

\begin{figure*}[t]
\centering
\includegraphics[width=\linewidth]{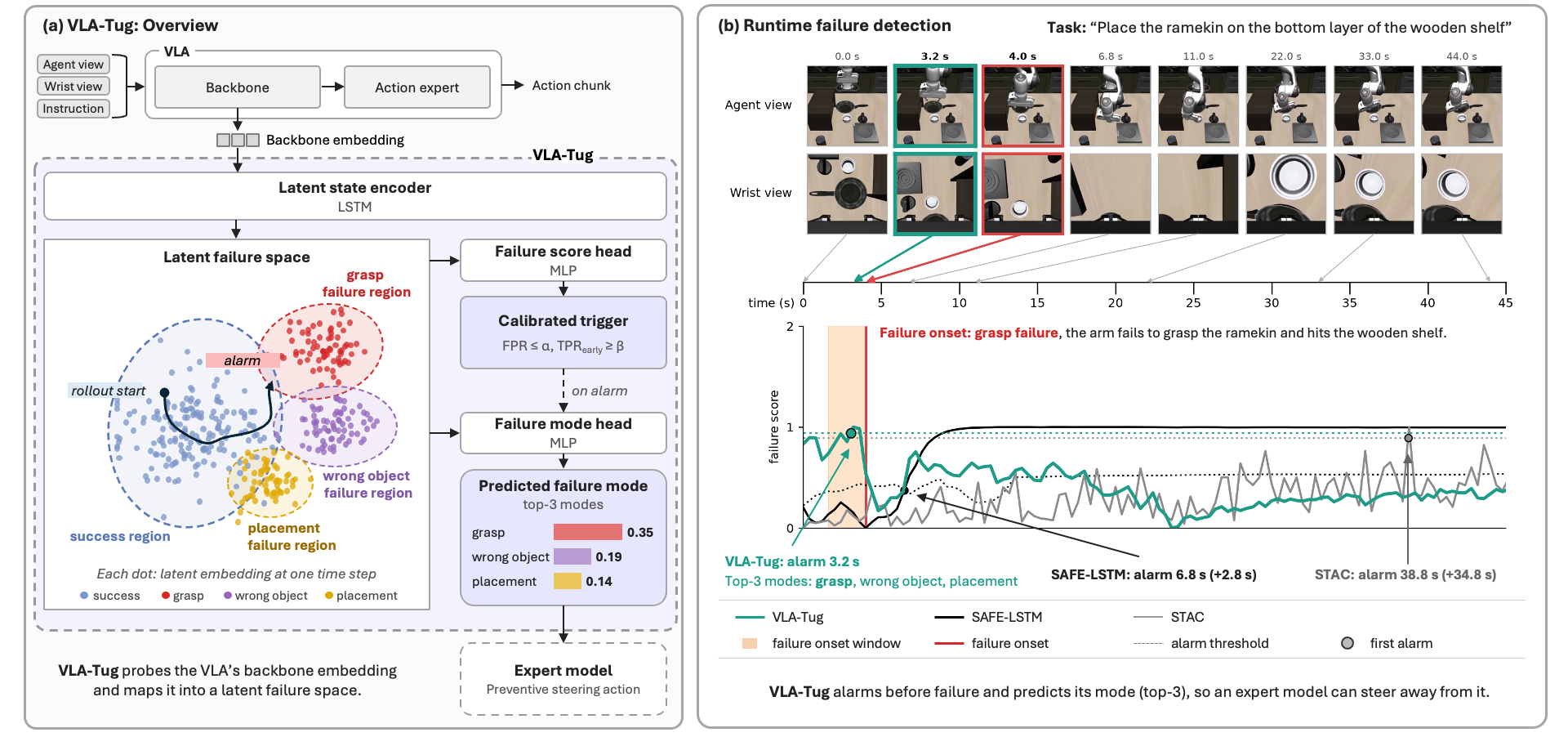}
\caption{We introduce \ours{}, a probe on the VLA backbone embeddings that predicts failures and their mode. (a) A recurrent encoder maps the embedding into a latent failure space; a score head and a calibrated trigger raise the alarm, and a mode head then predicts the failure mode. (b) \ours{} alarms before the failure, allowing early stopping or intervention by an expert model.}
\label{fig:overview}
\end{figure*}

\section{Introduction}
Vision-Language-Action models (VLAs) \cite{intelligence__05_2025, intelligence__07_2026, nvidia_gr00t_2025} are strong candidates for general-purpose embodied intelligence. By pairing a VLM backbone pre-trained on internet-scale data with an action expert fine-tuned on demonstration trajectories, they excel at complex dexterous tasks such as tightening screws or assembling cardboard boxes. Yet their performance is brittle and can degrade significantly under distribution shift, such as changes in lighting, novel textures, or rare object combinations \cite{wang_libero-x_2026, fei_libero-plus_2025}. A VLA can therefore fail unexpectedly, especially as task and environment novelty increases, and such failures can be costly, damaging materials and robots, or outright dangerous to humans. This brittleness has slowed the adoption of VLAs in factories, homes, and public spaces.

Among multiple solutions, one solution that the robotics community is exploring is VLA steering. The idea of steering comes from research on LLMs, where activations are selectively modified to nudge a model towards a desired output \cite{turner_steering_2024}. Steering VLAs is particularly interesting because it can unlock behaviors that cannot be reached through prompting alone \cite{buurmeijer_observing_2026, swann_sparse_2026}. Various steering methods have recently been proposed \citep{li_towards_2026, liu_vls_2026, tan_rl2-vla_2026, swann_sparse_2026, cui_dreamsteer_2026, jeong_learning_2026, liVLAsAreConfined2025}, but they rarely use a calibrated probe to decide \emph{when} to steer. Some steer continuously \cite{li_towards_2026, liu_vls_2026, cui_dreamsteer_2026}, some rely on human-specified offline triggers \cite{swann_sparse_2026, liVLAsAreConfined2025}, and others are gated by a one-sided calibrated trigger \cite{tan_rl2-vla_2026, jeong_learning_2026}. Overall, they lack an efficient and principled trigger that would make them deployable in real-world systems.

Research on failure detection for VLAs is becoming increasingly mainstream \cite{gu_safe_2025, agia_unpacking_2024, ma_failurespot_2026, huang_actprobe_2026}. In a nutshell, these methods compute a failure score from signals available in the VLA, often its latent variables \cite{gu_safe_2025} or predicted actions \cite{agia_unpacking_2024, huang_actprobe_2026, ma_failurespot_2026}, that encode how likely the policy is to fail at the current task. A detection threshold is then calibrated to give probabilistic guarantees on the detector's performance \cite{gu_safe_2025}. However, these guarantees often cover only the false-positive rate (FPR), i.e.\ the rate of successes misclassified as failures, and not a target true-positive rate (TPR), i.e.\ the rate of correctly classified failures, which is arguably even more important in safety-critical settings where a single failure could harm a person or damage costly equipment. These methods are also seldom evaluated systematically against ground-truth failure times, which we argue is crucial to assess whether they can prevent failures \emph{before} they happen and thus enable VLA steering. Finally, they cannot predict failure modes (grasping, pushing, opening), so steering cannot be targeted towards the root cause and would require human interventions.

We argue that steering would be best supported by a probe that is {\footnotesize (T.)}~\emph{timely}, detecting failures before they occur; {\footnotesize (I.)}~\emph{informative}, outputting semantic information beyond a numerical score; {\footnotesize (C.)}~\emph{calibrated}, achieving the desired error rates; and {\footnotesize (E.)}~\emph{efficient} to train, calibrate, and deploy.

We address this with \ours{}\footnote{Like a tugboat beside the vessel it guides, the probe is small next to the policy it monitors and acts only when needed.}, a well-behaved probe designed to support VLA steering in the real world. We train \ours{} on the latent layers of two VLAs \cite{intelligence__05_2025, nvidia_gr00t_2025} using \textcolor{black}{228} distinct tasks in two simulation environments and 8 real-world tasks, and evaluate it against ground-truth failure times. We show that supervision should focus on the pre-failure part of the trajectory to improve early failure detection, and that calibrating the detection threshold to guarantee both TPR and FPR simultaneously yields guarantees that hold far more often, including on unseen tasks. Building on these findings, we propose a real-time architecture that incorporates \ours{} to autonomously support the detection and prevention of VLA failures. Our contributions are as follows:

\begin{itemize}
    \item \textbf{\ours{}: A timely, informative, and efficient probe for \emph{early} failure detection.}  On unseen tasks, our probe improves the area under the early-detection curve by up to $30\%$ over the strongest learned baseline. \ours{} also predicts the failure mode, raising the top-3 accuracy on unseen tasks by up to \textcolor{black}{$44\%$} over a majority baseline, at under $1$\,ms per step.

    \item \textbf{Guarantees on false alarms and missed detections.} We certify FPR and $\textcolor{black}{\tprearly}$ targets jointly; both held in every certified run, compared to $57$--$71\%$ of runs for baselines. Under distribution shift to unseen tasks, the false-alarm target still held in $40$--$86\%$ of certified runs, compared to $0$--$21\%$ for baselines.

    \item \textbf{A deployable failure-monitoring system.} Through real-world experiments, we demonstrate that integrating the timely failure predictions and associated modes with steering strategies can lead to enhanced success rates.

    \item \textbf{An annotated VLA failure dataset.} We release a dataset of $\approx$289k policy-step VLA backbone embeddings from annotated $3{,}180$ episodes of diverse tasks.
\end{itemize}

\section{Related Work}

% \paragraph{Failure detection in VLAs.}

Existing methods can detect failures, but their alarms are not necessarily timely, informative, or sufficiently calibrated. Our method addresses these limitations through failure-aware representations, onset-based detection, and joint statistical guarantees.

\paragraph{Existing works and their limitations.}
Existing VLA failure detectors leverage different signals to identify potential failures, including external assessments from world models or LLM judges~\citep{liu_checkvla_2026,li_towards_2026}, action consistency across sampled or consecutive action chunks~\citep{agia_unpacking_2024,huang_actprobe_2026}, distributional deviations in policy inputs, outputs, or embeddings~\citep{xu_can_2025,gu_safe_2025}, and internal policy representations~\citep{gu_safe_2025,rajaprakash_safecast_2026,ma_failurespot_2026,chen_aegis_2026,park_hide-and-seek_2026,francis-meretzki_temporal_2026}.

These approaches have three key limitations.
\textbf{(1)} Detection performance is commonly evaluated at fixed time points using metrics such as AUROC or balanced accuracy~\citep{gu_safe_2025,huang_actprobe_2026}, without explicitly measuring whether alarms precede the actual failure onset.
Moreover, supervision ranges from episode-level failure labels assigned to all timesteps~\citep{gu_safe_2025}, to heuristic or human-refined timestep labels~\citep{ma_failurespot_2026}, to failure-free training~\citep{xu_can_2025}, without explicitly accounting for the timing of failures.
\textbf{(2)} Most detectors produce scalar risk scores without identifying the underlying failure modes. While semantic supervision and representation interpretability have been explored for embodied policies~\citep{xuVLMADEndtoEndAutonomous2024, swann_sparse_2026, buurmeijer_observing_2026}, they do not directly establish informative failure alarms.
\textbf{(3)} Existing calibration methods typically control a single criterion. Conformal methods provide false-positive guarantees for sequential failure detection~\citep{gu_safe_2025,rajaprakash_safecast_2026,park_hide-and-seek_2026}, calibrate anomaly or failure scores~\citep{xu_can_2025,ma_failurespot_2026}, or provide guarantees for one-shot policy compliance and embodied decision making~\citep{yang_repv_2025,chen_reconvla_2026,bhattKnowWhereYoure2025,yangUncertaintyGuidedEnhancementDriving2025}. However, these guarantees do not jointly account for false alarms and timely failure detection.

\paragraph{\ours{} addresses the limitations.}
We address these limitations by developing a failure detector that is \emph{timely, informative, and jointly calibrated}.
Following the internal-representation probing direction, we reuse the VLA's latent representations and align them with failure-mode semantics, enabling the detector to identify both potential failures and their types. To evaluate timeliness, we measure detection relative to annotated failure onsets and investigate supervision strategies that explicitly account for when failures occur. To provide statistical guarantees, we adopt Learn-then-Test~\citep{angelopoulosLearnThenTest2022} to jointly calibrate multiple detection criteria, including the false-positive rate (FPR) and early true-positive rate (\textcolor{black}{$\tprearly$}). Existing failure detectors primarily raise alarms without steering, while steering methods rely on predefined schedules or calibrated triggers. In contrast, to the best of our knowledge, \ours{} is the first framework to combine joint FPR and early-TPR calibration, failure-onset-aware evaluation, and failure-mode prediction for calibrated failure-triggered steering. \autoref{tab:related-work-comparison} compares \ours{} against the existing works.

\begin{table}[htb]
\centering
\caption{Comparison with related methods. \textcolor{black}{FPG / TPG: formal calibrated guarantee on the false-positive / true-positive rate.} EE: early evaluation, i.e.\ a per-episode criterion against annotated failure time; $\circ$ denotes partial use. FM: predicts the failure mode. Trigger: what decides when steering happens. A dash means n/a.}
\vspace{0.2cm}
\label{tab:related-work-comparison}
\scriptsize
\setlength{\tabcolsep}{3pt}
\begin{tabular}{@{}l l cccc@{}}
\toprule
Method & Trigger & FPG & \textcolor{black}{TPG} & EE & FM \\
\midrule
\multicolumn{6}{@{}l}{\textit{Failure detectors (no steering)}} \\
SAFE~\citep{gu_safe_2025}            & one-sided conformal & \checkmark & $\times$ & $\circ^{\dagger}$ & $\times$ \\
% SAFECAST~\citep{rajaprakash_safecast_2026} & one-sided conformal & \checkmark & $\times$ & $\times$ & $\times$ \\
ActProbe~\citep{huang_actprobe_2026} & split conformal & \checkmark & $\times$ & $\times^{\ddagger}$ & $\times$ \\
STAC~\citep{agia_unpacking_2024}     & one-sided conformal & \checkmark & $\times$ & $\times$ & $\times$ \\
FailureSpot~\citep{ma_failurespot_2026} & one-sided conformal & \checkmark & $\times$ & $\circ^{\dagger}$ & $\times$ \\
\midrule
\multicolumn{6}{@{}l}{\textit{Steering methods}} \\
VLA-Pilot~\citep{li_towards_2026}    & always on & -- & -- & -- & -- \\
VLS~\citep{liu_vls_2026} & \shortstack[l]{always on$^{\S}$} & -- & -- & -- & -- \\
DreamSteer~\citep{cui_dreamsteer_2026} & \shortstack[l]{fixed period} & -- & -- & -- & -- \\
SAE-VLA~\citep{swann_sparse_2026}    & human-specified & -- & -- & -- & -- \\
TLI~\citep{liVLAsAreConfined2025}              & human-specified & -- & -- & -- & -- \\
RL$^2$-VLA~\citep{tan_rl2-vla_2026}  & one-sided conformal & \checkmark & $\times$ & $\times$ & $\times$ \\
\shortstack[l]{LFP \citep{jeong_learning_2026}} & one-sided conformal & \checkmark$^{\mp}$ & $\times$ & $\times$ & $\times$ \\
\midrule
\textbf{\ours{} (Ours)} & \shortstack[l]{jointly calibrated} & \checkmark & \checkmark & \checkmark & \checkmark \\
\bottomrule
\end{tabular}

\vspace{2pt}
{\raggedright\scriptsize
$\dagger$ Annotated failure time used only in aggregate or qualitatively.
$\ddagger$ Timeliness measured via normalized detection time and early-window AUC relative to the episode length; no annotated failure time.
$\S$ Threshold on VLM-generated rewards.
$\mp$ Class-conditional bound on harmful steering. \par}
\end{table}

\section{Problem Statement}
\label{sec:problem_statement}

In this section, we start by giving several broad definitions used to describe how failure detectors for VLAs function. Next, we give a mathematical formulation of our problem and then introduce how \ours{} addresses it.

\subsection{Definitions}
To facilitate comparisons across methods, we give general definitions for the concepts of failure score, detection threshold, detection time, failure time, and failure mode.

\begin{itemize}
    \item \textbf{Failure score.} The score $s_t \in \mathbb{R}$, defined at every timestep $t$, encodes the probe's belief that the VLA is failing. It need not be a probability in the strict sense.
    \item \textbf{Detection threshold.} The threshold $h^\star$ over $s_t$ separates failures from successes.
    \item \textbf{Detection time.} The time $t_{\mathrm{det}} = \min\{t : s_t \geq h^\star\}$ is the first time the score crosses the threshold.
    \item \textbf{Failure time.} The time $t_{\mathrm{fail}}$, which we also call the failure onset, is when the policy makes an obvious mistake or its progress stalls to the point where an intervention, human or otherwise, is needed.
    \item \textbf{Failure mode.} The label $f_{\mathrm{mode}}$ is a high-level semantic category assigned to each failure, such as grasping failure, placement failure, or no progress.
\end{itemize}

\vspace{-10pt}

These definitions are general, and their exact form varies slightly across methods. The interpretation of $s_t$ depends on the training strategy, but a higher score usually signals failure and a lower score success. The threshold $h^\star$ may be fixed or vary with time, and it is often calibrated with probabilistic guarantees on a set of calibration rollouts $\textcolor{black}{\Dcal}$. If $s_t < h^\star$ for all $t$, the detection time $t_{\mathrm{det}}$ is undefined, and the rollout raises no alarm. The failure time definition follows \cite{gu_safe_2025}.

\subsection{Problem Formulation}

Consider a VLA policy $\mathbf{f}_{\text{VLA}}(\cdot)$ and a rollout $\tau_i$ of $T$ policy steps. At each policy step $t \in \{1,\dots,T\}$, $\mathbf{f}_{\text{VLA}}(\cdot)$ takes as input RGB observations $\mathbf{o}_t$ (agent and wrist cameras) and a natural language prompt $\mathbf{p}$, and produces a chunk of $m$ control actions $\mathbf{a}_{t;0:m-1}$. We denote the VLA's internal embedding at step $t$ as $\mathbf{e}_t$, so that $\tau_i = \{(\mathbf{p}, \mathbf{o}_t, \mathbf{e}_t, \mathbf{a}_{t;0:m-1})\}_{t=1}^{T}$. Each rollout has an episode-level label $Y_i \in \{0,1\}$, where $Y_i = 1$ marks a failure and $Y_i = 0$ a success. If $Y_i = 1$, the rollout also has a failure time $\textcolor{black}{\tfail}$ and a failure mode $\textcolor{black}{\fmode} \in \mathcal{F}$, where $\mathcal{F}$ is a fixed taxonomy of failure modes.

\textbf{Problem.} Given a training set and a calibration set $\textcolor{black}{\Dcal} = \{\tau_i\}$, and access to the internal embeddings $\mathbf{e}_t$ of $\mathbf{f}_{\text{VLA}}(\cdot)$, predict an imminent task failure \emph{before it happens}, i.e. $\textcolor{black}{\tdet}\leq \textcolor{black}{\tfail}$ where $\textcolor{black}{\tdet}$ is the detection time, and predict the failure mode $\textcolor{black}{\fmode}$ in a manner that holds up empirically on tasks unseen during calibration.

We propose \ours{}, a probe that is trained and calibrated to detect failures before they happen. It is:

{\footnotesize (T.)}~\emph{Timely.} We propose a training and evaluation strategy that accounts for the failure time and improves early failure detection.

{\footnotesize (C.)}~\emph{Calibrated.} We introduce a risk formulation that jointly bounds the false-positive rate and the early miss rate \textcolor{black}{$1-\tprearly$}, and examine how well these targets hold on unseen tasks.

{\footnotesize (I.)}~\emph{Informative.} We design \ours{} to predict the failure mode, giving the semantic information needed to choose a corrective action.

{\footnotesize (E.)}~\emph{Efficient.} \ours{} adds minimal latency and parameters to the VLA, and is cheap to train and calibrate.

\section{Methodology}
\label{sec:methodology}

In this section we introduce \ours{}. In \autoref{sec:latent_probing} we present the probe and how we train it, and in \autoref{sec:guarantees} we show how we calibrate its alarm threshold and what guarantees this gives.

\subsection{Failure Detection Through Latent Space Probing}
\label{sec:latent_probing}

Building on existing works \cite{gu_safe_2025}, we instantiate the probe as a combination of LSTMs and MLPs, the goal being to capture a non-linear pattern in the backbone embeddings that generalizes beyond the seen episodes. We briefly introduce the model inputs, followed by a discussion of the model architecture and finally the training strategy.

\paragraph{We use VLA backbone embeddings as the inputs.} We explore two distinct layers: backbone layer 18 (after normalization) in $\pi_{0.5}$ and layer 16 in GR00T-N1.7, focusing on the latter in particular. At every policy step $t$ we average the backbone tokens of each modality (agent camera, wrist camera, instructions) and concatenate the three $2048$-dimensional vectors into a single embedding $e_t \in \mathbb{R}^{6144}$.

\paragraph{We leverage a combination of LSTMs and MLPs.} A single-layer LSTM with $256$ hidden units reads the sequence of embeddings and produces a latent state $u_t = \text{LSTM}(e_{1:t})$ at every policy step. Three MLP heads read this shared latent state (see \autoref{fig:overview}). The failure score head predicts the failure score, $s_t = \sigma(g_s(u_t))$. To make the probe informative, the mode head predicts a distribution over the $10$ failure modes $\textcolor{black}{\fmode}$, $m_t = \text{softmax}(g_m(u_t))$. Following \cite{xuVLMADEndtoEndAutonomous2024}, an auxiliary head, $c_t = g_c(u_t)$, aligns the latent state with a CLIP embedding of the description of the current action. Each head $g$ is an MLP with one hidden layer of $64$ ReLU units. The probe has $6.6$M parameters.

\paragraph{We train the probe to predict features that precede a failure.} We hypothesize that training on post-failure embeddings teaches the probe what already failed trajectories look like, not what trajectories \emph{about to} fail look like. If so, to \emph{predict} failures the probe must be trained on the embeddings $e_t$ before the failure time $t_{\mathrm{fail}}$ (\emph{pre-onset} supervision), rather than after it \cite{ma_failurespot_2026} or at all times of failed rollouts \cite{gu_safe_2025}. To test this, we compare four ways of labeling the embeddings of failed rollouts (\autoref{fig:supervision}; the mathematical formulation is in App. \ref{app:supervision}). Successful rollouts are labeled $0$ in all four. The window $w$ in \emph{onset} and \emph{post-onset} is the time span before $t_{\mathrm{fail}}$ that is labeled as failure. Pre-onset and onset truncate the rollout at the failure time, while post-onset and trajectory keep all steps. We exclude the steps after $t_{\mathrm{fail}}$ in the first two strategies instead of labeling them $0$: these embeddings already show a failed rollout, so labeling them as successes would give the probe contradicting training signals.

\newcommand{\supbar}[3]{%
  \fill[#1] (0,-0.12) rectangle (1.45,0.12);
  \fill[#2] (1.45,-0.12) rectangle (2.3,0.12);
  \fill[#3] (2.3,-0.12) rectangle (3.7,0.12);
  \draw[black!35, line width=0.3pt]
    (0,-0.12) rectangle (3.7,0.12);
  \draw[black!35, line width=0.3pt]
    (1.45,-0.12) -- (1.45,0.12)
    (2.3,-0.12) -- (2.3,0.12);
}

\begin{figure}[h]
\centering
\begin{tikzpicture}[font=\footnotesize]
  % Strategy labels
  \node[anchor=east] at (1.85,0.9)   {\emph{Pre-onset}};
  \node[anchor=east] at (1.85,0.35)  {\emph{Onset}};
  \node[anchor=east] at (1.85,-0.2)  {\emph{Post-onset}};
  \node[anchor=east] at (1.85,-0.75) {\emph{Trajectory}};

  % Supervision bars
  \begin{scope}[shift={(2.0,0.9)}]
    \supbar{red!35}{red!35}{gray!18}
  \end{scope}
  \begin{scope}[shift={(2.0,0.35)}]
    \supbar{blue!22}{red!35}{gray!18}
  \end{scope}
  \begin{scope}[shift={(2.0,-0.2)}]
    \supbar{blue!22}{red!35}{red!35}
  \end{scope}
  \begin{scope}[shift={(2.0,-0.75)}]
    \supbar{red!35}{red!35}{red!35}
  \end{scope}

  % Time axis and markers aligned with the bar boundaries
  \draw[->, black!45]
    (2.0,-1.3) -- (5.7,-1.3)
    node[anchor=north east, font=\scriptsize] {time};

  \draw[black!60]
    (3.45,-1.24) -- (3.45,-1.36)
    node[below, font=\scriptsize, text=black]
    {$t_{\mathrm{fail}}-w$};

  \draw[black!60]
    (4.3,-1.24) -- (4.3,-1.36)
    node[below, font=\scriptsize, text=black]
    {$t_{\mathrm{fail}}$};

  \foreach \x/\c/\t in {
    0.0/red!35/{Failure ($y_t{=}1$)},
    2.1/blue!22/{No failure ($y_t{=}0$)},
    4.6/gray!18/{Excluded from the loss}
  } {
    \fill[\c]
      (\x,-2.15) rectangle (\x+0.22,-1.93);
    \draw[black!35, line width=0.3pt]
      (\x,-2.15) rectangle (\x+0.22,-1.93);
    \node[anchor=west, font=\scriptsize]
      at (\x+0.28,-2.04) {\t};
  }
\end{tikzpicture}
\caption{How the embeddings $e_t$ of a failed rollout are
labelled for the four supervision strategies.}
\label{fig:supervision}
\end{figure}
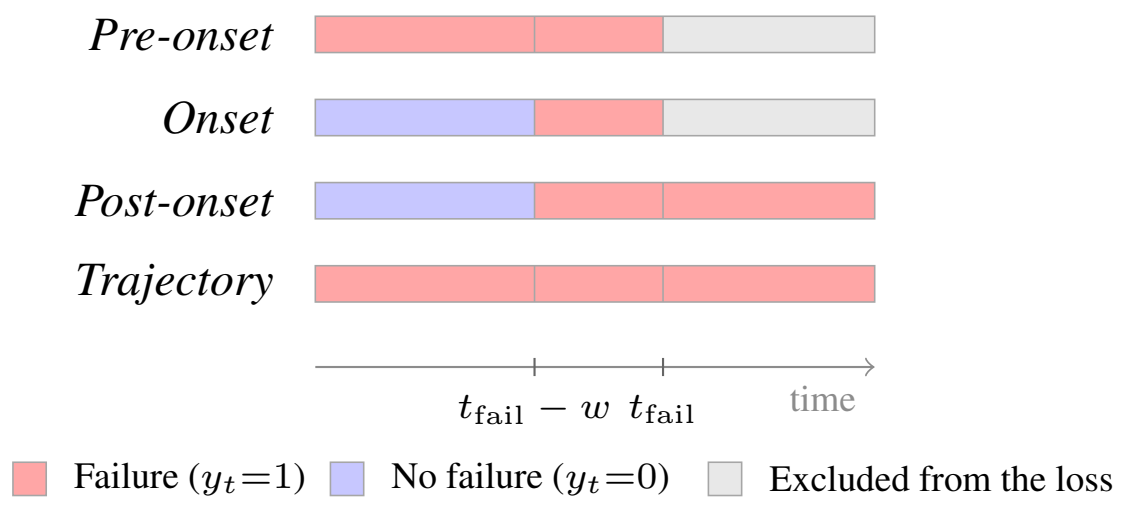

\paragraph{We optimize three objectives.} With the labels above, the probe is trained with
\begin{equation}
    \mathcal{L} = \mathcal{L}_{\text{score}} + \lambda_m \mathcal{L}_{\text{mode}} + \lambda_c \mathcal{L}_{\text{CLIP}},
\end{equation}
with $\lambda_m = \lambda_c = 1$. $\mathcal{L}_{\text{score}}$ is a binary cross-entropy over all steps that are not excluded from the loss, weighted by the ratio of negative to positive steps in the training set. $\mathcal{L}_{\text{mode}}$ is a cross-entropy against the true failure mode, applied to the positive steps of failed rollouts and weighted to balance the modes. $\mathcal{L}_{\text{CLIP}}$ is an auxiliary contrastive loss that aligns $c_t$ with the CLIP embedding of the step's description. We train with Adam (learning rate $3\times10^{-4}$, batch size $32$ rollouts, at most $100$ epochs) on standardized inputs, with early stopping on the validation $\mathcal{L}_{\text{score}}$ (patience $15$).

\subsection{Uncertainty Quantification and Guarantees}
\label{sec:guarantees}

Where previous works only provided guarantees of false-positive rates (i.e. the rate of successful rollouts that were flagged as failures) \citep{gu_safe_2025, rajaprakash_safecast_2026}, they did not provide guarantees on the false-negative rate (i.e. the rate of unsuccessful trajectories that were not flagged). We will first formulate the risks and show how to calibrate the detection threshold $h^\star$. In an effort to make these bounds interpretable to the practitioner, we compute the probability of failure given that a rollout was flagged.

\paragraph{\ours{} guarantees both FPR \textit{and} \textcolor{black}{$\tprearly$}.} In general, the \textcolor{black}{FPR} is an operational metric that we wish to minimize to prevent unnecessary steering, the \textcolor{black}{TPR} is a safety metric that we wish to maximize to prevent undetected failures. We can control these risks using the Learn-then-Test framework \cite{angelopoulosLearnThenTest2022}. Let $h^\star$ be a constant detection threshold, and $s_t$ the failure score produced by our probe at each timestep $t$ for each rollout $\tau$. Furthermore, let the detection time $\textcolor{black}{\tdet}$ be the first time the failure score $s_t$ exceeds the threshold $h^\star$. Finally, the ground truth failure time $\textcolor{black}{\tfail}$ remains the first time $t$ when failure becomes apparent or a human operator would choose to take over. We propose to formulate the above mentioned risks as follows:

{\scriptsize
\begin{align}
\mathrm{FPR}(h) &= \mathbb{E}_{\tau \sim \textcolor{black}{\Dcal^{\mathrm{succ}}}}[\mathbf{1}\{\exists\, \textcolor{black}{\tdet} : s_t(\tau) \textcolor{black}{\geq h}\}], \\
\textcolor{black}{\tprearly}(h) &= \mathbb{E}_{\tau \sim \textcolor{black}{\Dcal^{\mathrm{fail}}}}[\mathbf{1}\{\exists\, \textcolor{black}{\tdet} \leq \textcolor{black}{\tfail}: s_t(\tau) \textcolor{black}{\geq h}\}], 
\end{align}}

where $\textcolor{black}{\Dcal^{\mathrm{succ}}}$ and $\textcolor{black}{\Dcal^{\mathrm{fail}}}$ are the sets of successful and failed rollouts $\tau$, respectively, in the calibration set $\textcolor{black}{\Dcal}$. We are interested in finding the set $\hat\Lambda$ of values $h^\star$ that satisfy:
\begin{equation}
\label{eq:ltt_full}
\small
P\Big(\mathrm{FPR}(h) \leq \alpha, \ \textcolor{black}{\tprearly}(h) \geq \beta \ \ \forall\, h \in \hat\Lambda\Big) \geq 1-\delta,
\end{equation}

where $\alpha\in[0,1]$ and $\beta\in[0,1]$ are user-selected rates. This guarantee can be understood as saying that any $h\in \hat \Lambda$ guarantees that the false-positive rate (FPR) is at most $\alpha$ and the early true-positive rate (\textcolor{black}{$\tprearly$}) is at least $\beta$, with high-probability of no less than $1 - \delta$. We will see, in the experimental section, that this translates to guarantees that empirically behave better under distribution shift. Following \citep{angelopoulosLearnThenTest2022} we select a set of candidate values $\hat\Lambda$; from this set $\hat \Lambda$ of values $h$ we select the value $h^\star$ that maximizes $\textcolor{black}{\tprearly}$. 

\paragraph{Probability of failure given that an alarm is raised.}
Since our goal is a probe that can be deployed in real-world systems, we complement the formal $\mathrm{TPR}_{\mathrm{early}}$ and $\mathrm{FPR}$ guarantees with an actionable interpretation: a bound on the positive predictive value (PPV). The PPV is the probability that failure occurs given score $s_t$ exceeds the threshold $h^\star$. Existing calibration methods \cite{gu_safe_2025} cannot give such a bound, since they lack guarantees on the $\mathrm{TPR}$.

Let $\textsc{Failure}$ denote rollout-level failure, and let $\textsc{Exceed}$ denote the event that the score $s_t$ exceeds the threshold $h^\star$ at some time $t$. The PPV is $P(\textsc{Failure}\mid\textsc{Exceed})$. Define the failure prevalence $\pi = P(\textsc{Failure})$, the true positive rate $\mathrm{TPR} = P(\textsc{Exceed}\mid\textsc{Failure})$, and the false positive rate $\mathrm{FPR} = P(\textsc{Exceed}\mid\textsc{not Failure})$. By Bayes' rule,
\begin{equation}
\small
P(\textsc{Failure}\mid\textsc{Exceed}) = \frac{\pi \cdot \mathrm{TPR}}{\pi \cdot \mathrm{TPR} + (1-\pi) \cdot \mathrm{FPR}}.
\end{equation}
Detections that occur before the failure onset are a subset of all detections, so $\mathrm{TPR} \geq \mathrm{TPR}_{\mathrm{early}}$. Our guarantees give $\mathrm{TPR}_{\mathrm{early}} \geq \beta$ and $\mathrm{FPR} \leq \alpha$. Since the PPV is non-decreasing in $\mathrm{TPR}$ and non-increasing in $\mathrm{FPR}$, substituting the worst-case values yields the conservative bound:
\begin{equation}
\small
P(\textsc{Failure}\mid\textsc{Exceed}) \;\geq\; \frac{\pi \cdot \beta}{\pi \cdot \beta + (1-\pi) \cdot \alpha}.
\label{eq:ppv-bound}
\end{equation}
The bound increases with $\pi$, so a lower bound on the prevalence suffices. Given an estimate of $\pi$, this gives practitioners an intuitive reading of the guarantees.

\vspace{0.5 cm}

\section{Dataset}
\label{subsec:dataset}

To evaluate \ours{}, we create a labeled dataset of rollouts of two VLA policies, $\pi_{0.5}$ \citep{intelligence__05_2025} and GR00T-N1.7 \citep{nvidia_gr00t_2025}, with failure labels and semantic descriptions of the robot's actions. We first describe the benchmarks and the splits, and then the labeling pipeline.

\begin{figure}[h]
    \centering
    \includegraphics[width=\linewidth]{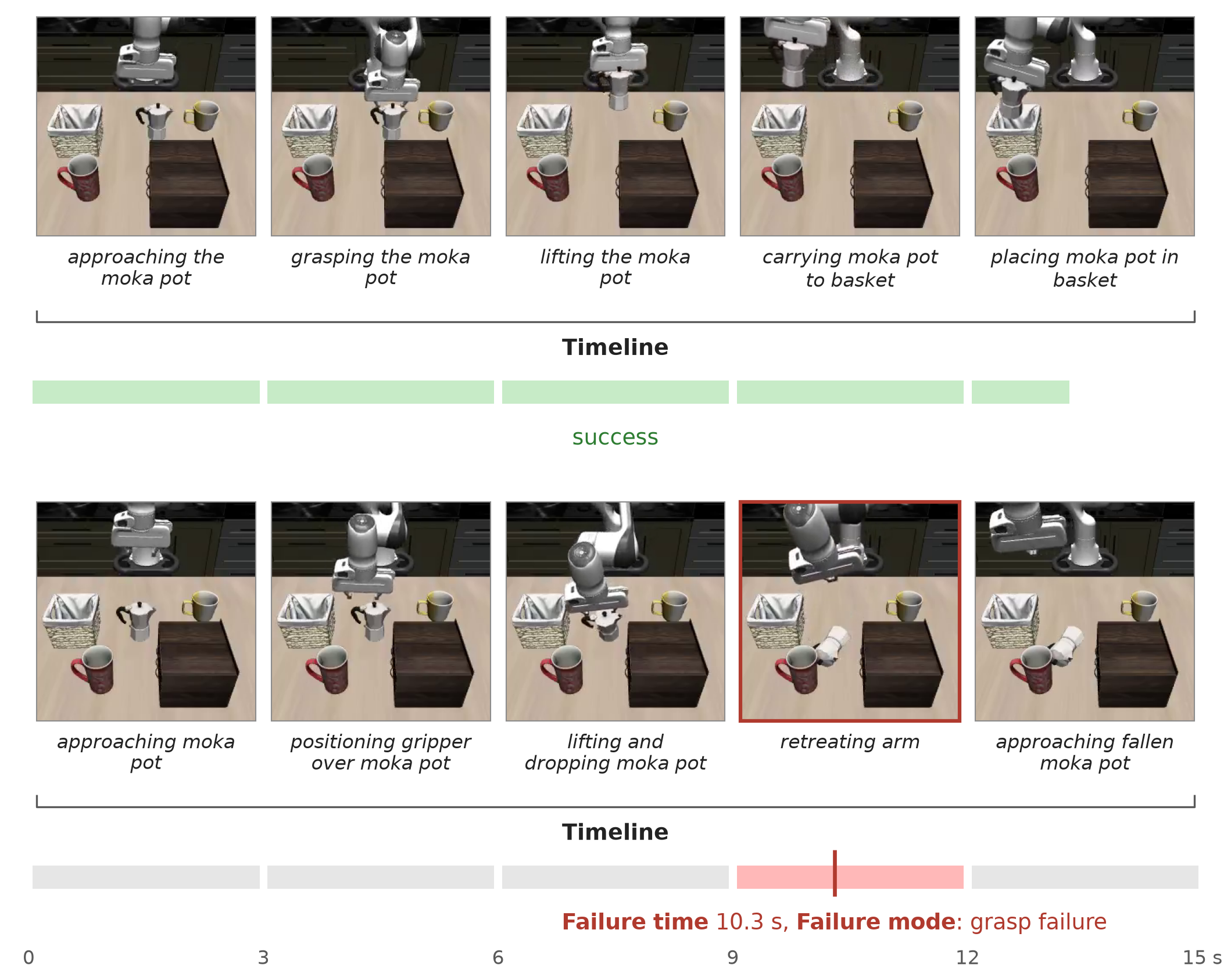}
    \caption{A successful (top) and a failed (bottom) rollout with their \emph{timeline}, \emph{failure time}, and \emph{failure mode} labels.}
    \label{fig:label_example}
\end{figure}

\paragraph{Benchmarks and splits.} \label{par:simulations} We use LIBERO \citep{liu_libero_2023}, LIBERO-Occ \citep{li_libero-occ_2026}, and LIBERO-X \citep{wang_libero-x_2026}, as well as real-robot tasks. Both policies succeed on most LIBERO episodes, so we combine it with LIBERO-Occ to obtain enough failures. We propose splits of increasing distribution shift to evaluate generalization in a principled way; \autoref{tab:libero_split_counts} gives how the tasks are held out and their counts. In LIBERO-X, each task has a \emph{skill}, the manipulation behavior (such as pushing a button or stacking objects), and a \emph{signature}, the combination of its skill, target class, relation, landmark, and horizon. The splits are:

\textbf{\newtask{}} holds out individual tasks; their skill and signature can still occur in the training tasks, the shift is mild.

\textbf{\newsig{}} holds out all tasks with a given signature: the components are familiar, but their combination is new.

\textbf{\newskill{}} holds out a complete skill, so the tested behavior is absent from training: this is the largest shift of the three.

\begin{table}[htb]
\centering
\caption{Task count by dataset and split.}
\label{tab:libero_split_counts}
\scriptsize
\setlength{\tabcolsep}{4pt}
\begin{tabular}{@{}ll cc@{}}
\toprule
Dataset & Split & Seen tasks & Unseen tasks \\
\midrule
LIBERO & leave-three-out & 7 & 3 \\
\midrule
\multirow{3}{*}{LIBERO-X} & \newtask{} & 175 & 43 \\
 & \newsig{} & 162 & 56 \\
 & \newskill{}$^\star$ & 159--209 & 9--59 \\
\midrule
Real-world & leave-one-out & 7 & 1 \\
\bottomrule
\end{tabular}

{\scriptsize
$^\star$ There are 8 skills, each with a different seen/unseen split. \par}

\end{table}

\paragraph{Efficient labeling pipeline.} We label each rollout from its video with Gemini (gemini-3.5-flash), as in \autoref{fig:label_example}. Every rollout receives a \emph{timeline} label, and failed rollouts also receive a \emph{failure mode} label and a \emph{failure time} label. The timeline is a short keyword phrase per 3-second window that we use for CLIP alignment. The failure mode $\textcolor{black}{\fmode}$ comes from a 10-way taxonomy $\mathcal{F}$. The failure time $\textcolor{black}{\tfail}$ is the earliest point at which the mistake responsible for the failure becomes apparent or, following \cite{gu_safe_2025}, the time a human operator would take over if none is apparent.

We open-source this dataset including over $288{,}000$ embeddings from $3{,}180$ labelled rollouts at \href{https://vla-tug.github.io}{vla-tug.github.io}.

% python -m cli.figures: areas and costs read from results/, see docs/FIGURES.md; needs booktabs, graphicx and xcolor with the table option
\begin{table*}[t]
\centering
\caption{Early failure detection on seen and unseen tasks: area under the TPR early vs.\ FPR curve, averaged over splits. Seen: held-out episodes of training tasks; unseen: held-out tasks. Cal.: calibrator (MC max-conformal, FC functional conformal). Cost on LIBERO-X (lower is better): fit time, memory to score a new episode, and per-step latency on a GPU and a CPU. Dashes: not available. Best in bold, second best in italics.}
\label{tab:main}
\setlength{\tabcolsep}{4pt}
\resizebox{\textwidth}{!}{%
\begin{tabular}{llrrrrrrrrrrrrrrrr}
\toprule
 & & \multicolumn{2}{c}{LIBERO-10 $\pi_{0.5}$} & \multicolumn{2}{c}{LIBERO-10 GR00T} & \multicolumn{2}{c}{LIBERO-X task} & \multicolumn{2}{c}{LIBERO-X signature} & \multicolumn{2}{c}{LIBERO-X skill} & \multicolumn{2}{c}{Real robot} & \multicolumn{4}{c}{Cost (LIBERO-X)} \\
\cmidrule(lr){3-4}\cmidrule(lr){5-6}\cmidrule(lr){7-8}\cmidrule(lr){9-10}\cmidrule(lr){11-12}\cmidrule(lr){13-14}\cmidrule(lr){15-18}
Method & Cal. & Seen & Unseen & Seen & Unseen & Seen & Unseen & Seen & Unseen & Seen & Unseen & Seen & Unseen & Fit (s) & State (MB) & GPU (ms) & CPU (ms) \\
\midrule
SAFE-LSTM & FC & \textit{75.5} & 34.9 & \textbf{74.1} & \textbf{51.2} & 64.6 & \textit{40.4} & 64.7 & 41.2 & 62.5 & \textit{30.1} & 44.5 & \textit{25.1} & 43.5 & 26.2 & 0.066 & 0.86 \\
SAFE-MLP & FC & 74.1 & 39.9 & 62.9 & 39.2 & 59.3 & \textbf{42.3} & 53.7 & \textit{46.2} & 56.8 & 23.5 & 40.1 & 20.2 & 114.5 & 6.3 & \textit{0.034} & \textit{0.032} \\
STAC & MC & 33.8 & 36.4 & 24.2 & 17.5 & 21.0 & 28.0 & 25.6 & 28.7 & 24.6 & 21.5 & -- & -- & \textbf{0.0} & \textbf{0.0} & 21.30 & -- \\
ActProbe & MC & 39.3 & 26.7 & 32.4 & 30.1 & 12.6 & 22.7 & 11.9 & 13.9 & 20.3 & 13.6 & -- & -- & 7.1 & \textit{0.09} & 0.15 & 0.11 \\
FailureSpot & FC & 61.9 & 36.2 & 36.9 & 44.7 & 26.3 & 20.4 & 28.0 & 26.2 & 30.4 & 19.6 & -- & -- & 9.2 & 6.4 & \textbf{0.033} & \textbf{0.029} \\
Mahalanobis & FC & \textbf{85.7} & \textit{40.2} & \textit{70.7} & 29.2 & \textit{69.6} & 30.6 & 64.4 & 22.3 & \textbf{71.2} & 15.9 & 43.0 & 19.6 & 15.2 & 302.0 & 0.41 & 5.37 \\
Euclidean & FC & 73.1 & 24.2 & 56.0 & 23.1 & \textbf{70.5} & 28.5 & \textit{65.4} & 19.7 & \textit{68.5} & 19.0 & \textit{45.1} & 8.3 & \textit{0.4} & 3056 & 13.19 & 368.83 \\
PCA-KMeans & FC & 35.7 & 12.9 & 44.5 & 20.0 & 31.7 & 25.0 & 35.3 & 32.5 & 37.8 & 20.7 & 24.0 & 19.4 & 39.9 & 1.6 & 0.12 & 0.055 \\
\midrule
\rowcolor{gray!15} \ours (Ours) & MC & 75.1 & \textbf{47.9} & 70.5 & \textit{47.4} & 67.9 & 40.2 & \textbf{67.0} & \textbf{47.0} & 67.8 & \textbf{39.0} & \textbf{47.6} & \textbf{25.8} & 6.8 & 26.5 & 0.12 & 0.94 \\
\bottomrule
\end{tabular}%
}
\end{table*}

\section{Experimental Results}

In this section, we answer five questions based on our problem formulation (\autoref{sec:problem_statement}):
(Q1. \emph{Timeliness}) Does training on features that precede the failure onset improve early failure detection on unseen tasks (\autoref{subsec:failure-detection})?
(Q2. \emph{Calibration}) Can \ours{} jointly bound the $\mathrm{FPR}$ and $\textcolor{black}{\tprearly}$, with bounds that hold on seen tasks and degrade gracefully on unseen tasks (\autoref{subsec:uncertainty-guarantees})?
(Q3. \emph{Informativeness}) Can \ours{} predict the failure mode based on VLA backbone embeddings (\autoref{subsec:informativeness})?
(Q4. \emph{Efficiency}) Are the cost and time of labeling, training, and inference appropriate for deployment in real-world systems (\autoref{subsec:efficiency})?
(Q5. \emph{Steering}) Can \ours{}, as a timely, informative, calibrated, and efficient probe, support VLA steering (\autoref{subsec:steering})?
We ablate our design choices in App. \ref{app:ablation}.

\subsection{Failure Detection and Timeliness Evaluation}
\label{subsec:failure-detection}

We evaluate \ours{} against a series of baselines using metrics that depend on the ground-truth failure time. Firstly, we describe how we split the data. Secondly, we review the baselines. Thirdly, we formally define our evaluation metrics. Lastly, we discuss the results.

\begin{figure}[h]
\centering
\begin{tikzpicture}[font=\scriptsize]
  \def\W{7.6}
  \def\H{0.72}
  \pgfmathsetmacro{\wseen}{0.76*\W}
  \pgfmathsetmacro{\xval}{0.65*\W}
  \pgfmathsetmacro{\xcal}{0.80*\W}
  \pgfmathsetmacro{\xheld}{0.90*\W}

  \tikzset{
    border/.style={draw=black!35, line width=0.3pt},
    splitlabel/.style={align=center, inner sep=1pt},
    guide/.style={draw=black!40, dashed, line width=0.3pt}
  }

  % Task-level split (schematic proportions)
  \fill[blue!22] (0,1.55) rectangle (\wseen,2.12);
  \fill[gray!18] (\wseen,1.55) rectangle (\W,2.12);
  \draw[border] (0,1.55) rectangle (\W,2.12);
  \draw[border] (\wseen,1.55) -- (\wseen,2.12);

  \node[splitlabel] at ({0.5*\wseen},1.835)
    {Seen tasks};
  \node[splitlabel] at ({0.5*(\wseen+\W)},1.835)
    {Unseen tasks};

  % Expand the seen-task subset into its episode-level split
  \draw[guide] (0,1.55) -- (0,\H);
  \draw[guide] (\wseen,1.55) -- (\W,\H);
  \node[text=black!65] at ({0.5*\W},1.14)
    {\emph{Seen tasks: split by episode}};

  % Episode-level split
  \fill[blue!22] (0,0) rectangle (\xval,\H);
  \fill[blue!10] (\xval,0) rectangle (\xcal,\H);
  \fill[gray!10] (\xcal,0) rectangle (\xheld,\H);
  \fill[gray!18] (\xheld,0) rectangle (\W,\H);

  \draw[border] (0,0) rectangle (\W,\H);
  \foreach \x in {\xval,\xcal,\xheld}
    \draw[border] (\x,0) -- (\x,\H);

  \node[splitlabel] at ({0.5*\xval},{0.5*\H})
    {Train\\65\%};
  \node[splitlabel] at ({0.5*(\xval+\xcal)},{0.5*\H})
    {Val.\\15\%};

  % Percentages inside narrow segments; labels below
  \node at ({0.5*(\xcal+\xheld)},{0.5*\H}) {10\%};
  \node at ({0.5*(\xheld+\W)},{0.5*\H}) {10\%};

  \draw[border]
    ({0.5*(\xcal+\xheld)},0) -- ++(0,-0.16);
  \node[anchor=north east, inner sep=1pt]
    at ({0.5*(\xcal+\xheld)+0.08},-0.20)
    {Calibration};

  \draw[border]
    ({0.5*(\xheld+\W)},0) -- ++(0,-0.16);
  \node[splitlabel, anchor=north]
    at ({0.5*(\xheld+\W)},-0.20)
    {Held-out\\seen};
\end{tikzpicture}
\caption{Task- and episode-level data splits. Episodes from
seen tasks are divided into training (65\%), validation (15\%),
calibration (10\%), and held-out seen (10\%) sets. Unseen tasks
form a separate evaluation set. Task-level proportions are schematic.}
\label{fig:data_split}
\end{figure}
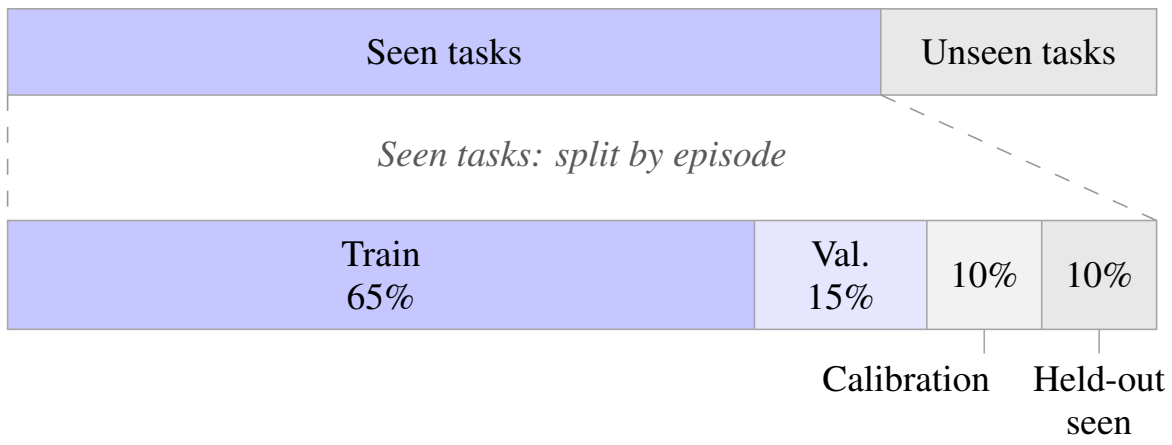

\paragraph{Data split.} For every run, we split the episodes as in \autoref{fig:data_split}. The training set fits the probe, the validation set selects the checkpoint, and the threshold is chosen on the calibration set. The calibration and held-out seen sets are exchangeable, and our guarantees apply to the held-out seen set. The unseen set, which the probe and the threshold never see, measures robustness to distribution shift. On LIBERO-X, the held-out tasks of the \newtask{} and \newsig{} splits are the same for every seed, because they are defined by the semantic axis described in \autoref{subsec:dataset}; the seed changes only the shuffle of the seen episodes and the initialization. The \newskill{} folds and the LIBERO-10 seeds hold out different tasks.

\paragraph{Baselines.} We consider the following baselines:

\textit{Embedding distance} (Mahalanobis, Euclidean, PCA-KMeans). A step is scored by how much closer its backbone embedding is to failed than to successful training steps.

\textit{Action consistency} (STAC, ActProbe, FailureSpot). Steps are scored based on the discrepancy between predicted and executed steps.

\textit{Latent} (SAFE). A step is scored by a learned projection of its backbone embedding.

\paragraph{Timeliness Metrics.} 
We evaluate \ours{} against three key metrics: TPR, $\textcolor{black}{\tprearly}$, and FPR. Existing works often evaluate their methods against AUROC \citep{gu_safe_2025, huang_actprobe_2026, ma_failurespot_2026} without accounting for the true failure times, making it hard to distinguish timely detections. The $\mathrm{TPR}$ and $\mathrm{FPR}$ follow their standard definitions. The $\textcolor{black}{\tprearly}$ is the rate of failures detected before the failure time, i.e. $\textcolor{black}{\tdet} \leq \textcolor{black}{\tfail}$. We additionally define $\mathrm{AUROC}_{\mathrm{early}}$ as the area under the $\mathrm{TPR}_{\mathrm{early}}$-$\mathrm{FPR}$ curve. We compute it on the support $\mathrm{FPR} \in [0, 0.4]$, normalized by its width \textcolor{black}{$0.4$} (reported in \%), since a detector that raises false alarms on more than $40\%$ of successful rollouts is impractical to deploy. We formally define these metrics in App. \ref{app:failure-detection}.

\paragraph{Results and Discussion.}

\begin{figure*}[t]
    \centering
    \includegraphics[width=\linewidth]{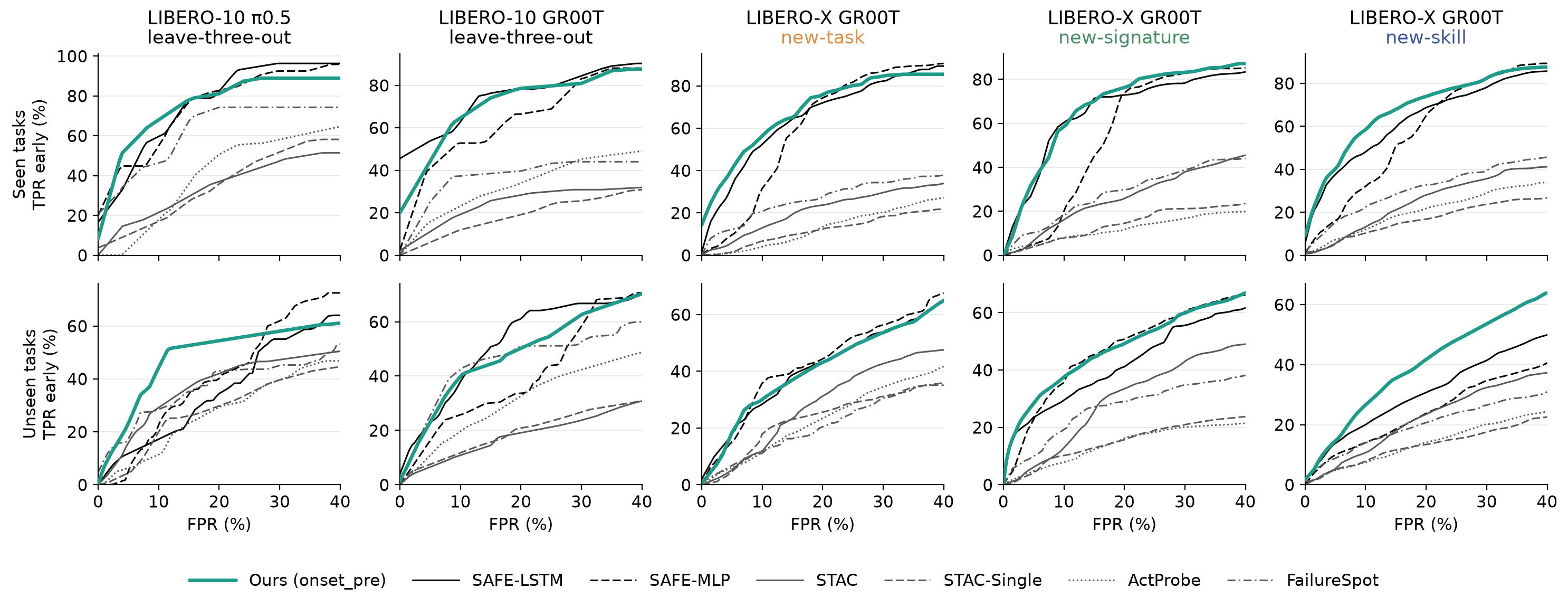}
    \caption{Early detection on seen (top) and unseen (bottom) tasks: the share of failures detected at or before the failure time (\textcolor{black}{$\tprearly$}) against the false-positive rate, for \ours{} and the learned and action-based baselines. Each curve is the mean over the splits of a setting, and the area under it up to $40\%$ FPR is the metric of \autoref{tab:main}. Panel titles give the benchmark, the policy, and the held-out split.}
    \label{fig:compact_plot_auroc_early}
\end{figure*}

We ask whether training on features that precede the failure onset improves early failure detection on unseen tasks. Our hypothesis is that it does: post-failure features only describe failures the probe has already seen, so training on them keeps it from learning a general notion of what leads to a failure, while features before the onset carry the cues that come before one. We compare \ours{} against the baselines in \autoref{tab:main} and \autoref{fig:compact_plot_auroc_early}. On unseen tasks, our domain of interest, \ours{} is the best method in four of six settings: LIBERO-10 with $\pi_{0.5}$, \newsig{}, \newskill{}, and the real robot. Against SAFE-LSTM, the strongest learned baseline, its margin \textbf{grows from $\mathbf{-1\%}$ on \newtask{} to $\mathbf{+14\%}$ on \newsig{} and $\mathbf{+30\%}$ on \newskill{}}. On the real robot, the lead is small, and \ours{} does not lead on LIBERO-10 with GR00T. On seen tasks, the results are mixed. We ablate the label in App. \ref{app:ablation}: replacing the pre-onset label with the other variants lowers early detection on LIBERO-X.

\paragraph{Answer to Q1. Timeliness.} Yes: for most unseen settings. Training on features that precede the failure onset lets \ours{} raise the alarm at or before the failure onset more often than the baselines, and on LIBERO-X the gap widens as the test tasks move away from the training tasks.

\subsection{Calibration Evaluation}
\label{subsec:uncertainty-guarantees}
We validate the theoretical results from \autoref{sec:methodology} on seen tasks and examine how they hold under distribution shift on unseen tasks. We calibrate our probe with Learn-then-Test (LTT, $\delta = 0.1$) at \textcolor{black}{seven} false-positive-rate (FPR) targets (\textcolor{black}{$10$ to $50\%$}), each certified together with a \textcolor{black}{TPR early} target, on the 14 LIBERO-X runs ($3$ \newtask{} seeds, $3$ \newsig{} seeds, and $8$ \newskill{} folds). We compare against max conformal (MC)~\cite{agia_unpacking_2024} and functional conformal (FC)~\cite{gu_safe_2025} calibration (the complete table is in App.~\ref{app:guarantees}). We draw three conclusions:

\begin{itemize}
    \item 
    % \textbf{The rates we calibrate for are the rates we get.} 
    LTT bounds the \textcolor{black}{$\tprearly$} together with the FPR, which \textbf{hold in all certified runs on seen tasks and in \textcolor{black}{$40$--$86\%$} (FPR) and \textcolor{black}{$57$--$90\%$} (TPR early) of runs on unseen tasks}; baselines give no guarantee on the TPR. 
    % A practitioner can set the detection and false-alarm rates an application needs and expect to get them.

    \item 
    % \textbf{Abstention is a feature.} 
   LTT states upfront whether the system can support the required guarantee whereas existing methods can give a false sense of safety. LTT abstains on $4$ to $7$ of the $14$ runs, depending on the target (\textcolor{black}{FPR targets of $15$--$50\%$}), which tells us before deployment that the model cannot support the requested rates. MC always returns a threshold, but it meets the FPR target in only \textcolor{black}{$57$--$71\%$} of runs on seen tasks and in at most $21\%$ on unseen tasks. 

    \item 
    % \textbf{\ours{} provides a PPV bound.}
    Our LTT targets give a lower bound on the positive predictive value (PPV), the probability that an alarm is a true failure. It is $58$--$71\%$ across the FPR targets of the full table and holds in every certified run on seen tasks and in $70$--$100\%$ of runs on unseen tasks. Methods without a TPR guarantee cannot give it, and it tells the practitioner what an alarm means.
\end{itemize}

On unseen tasks, the guarantees hold only empirically, since new samples are not exchangeable with the calibration set, which applies to all three methods. The shares above are trends over 14 runs, not precise rates.

\paragraph{Answer to Q2. Calibration.} Yes: the rates we calibrate for are the rates we get on seen tasks. Under distribution shift, they hold in far more runs than with MC and FC, and when a target cannot be supported, LTT abstains instead of silently missing it. Together with the PPV bound, this makes \ours{} easier to integrate into real-world systems.

\subsection{Evaluation of Informativeness}
\label{subsec:informativeness}
Here we focus on failure mode prediction, a crucial element to enable failure prevention through steering. Failure modes are strongly imbalanced, so we report balanced accuracy, the mean over classes of the share of their failures predicted correctly. Since existing baselines do not predict failure modes, we compare against a majority baseline and focus on LIBERO-X. 

On LIBERO-X, the \textbf{probe improves top-1 accuracy over the majority baseline by up to $149\%$ on seen and $79\%$ on unseen tasks,} and top-3 accuracy by a smaller but consistent margin on both. On the real robot, it improves top-1 by $171\%$ and $8\%$ on seen and unseen tasks, respectively, and stays well above the baseline in top-3 on both. 

The practical consequence is that the probe narrows a failure down to three candidate classes, which are more likely than the majority baseline to include the true one. Different failure modes carry different risks, so a practitioner can use these candidates to select the level of response, for example responding to the riskiest of the three predicted classes (a stop, a retry, or a handover to a human) instead of applying the same intervention to every failure.

\paragraph{Answer to Q3. Informativeness.}  Yes: backbone embeddings carry enough information to predict the failure class above a majority baseline, which lets a practitioner match the response to the predicted risk.

\begin{figure*}[t]
    \centering
    \begin{subfigure}[b]{0.83\linewidth}
        \centering
        \includegraphics[width=\linewidth]{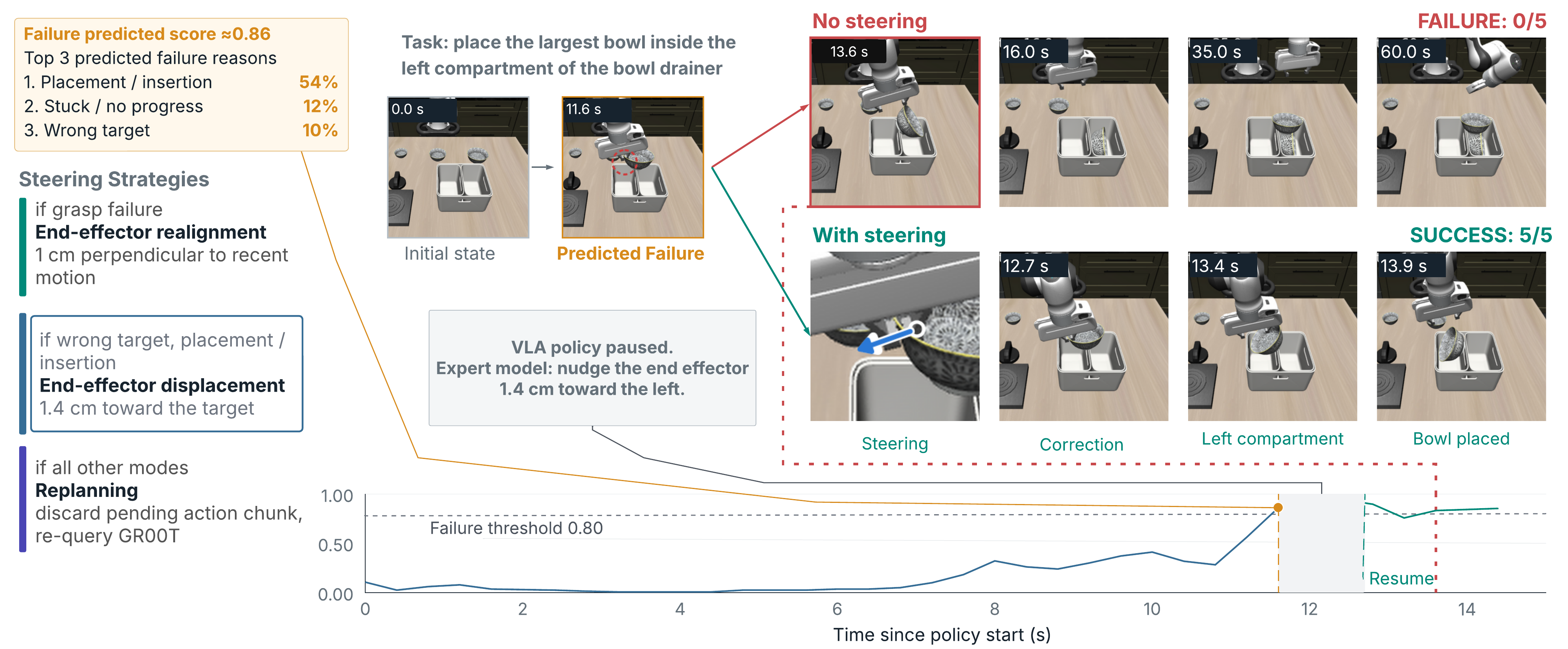}
        \caption{Example steering for the simulated task ``place the largest bowl inside the left compartment of the bowl drainer.'' The gripper drops the bowl into the right compartment with no steering and successfully drops it in the left compartment when steered towards the target location.}
        \label{fig:sim_bowl_steering}
    \end{subfigure}
    \hfill
    % Second subfigure (Figure 4)
    \begin{subfigure}[b]{0.83\linewidth}
        \centering
        \includegraphics[width=\linewidth]{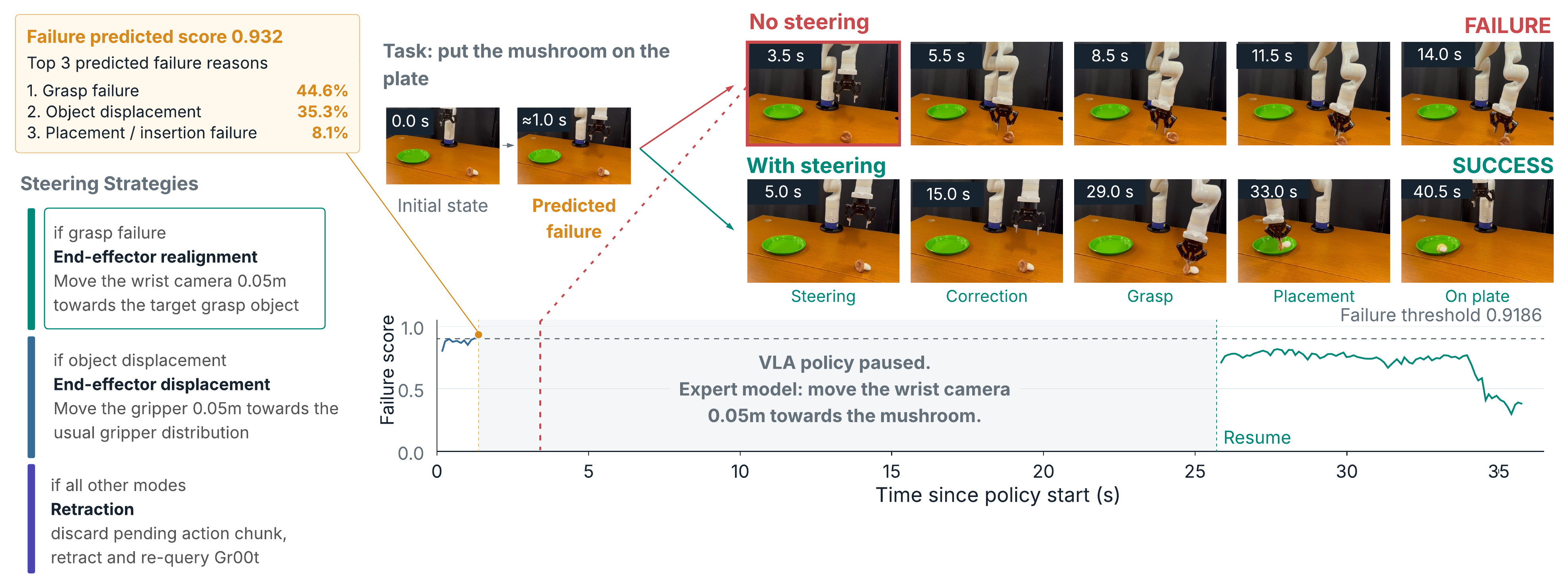}
        \caption{Example steering for the real-world task ``put the mushroom in the plate.'' The gripper misses the mushroom without steering and successfully grasps it when steered towards the mushroom.}
        \label{fig:mushroom_steering}
    \end{subfigure}
    
    \caption{Steering demonstrations showing failure scores and corrections for simulated and real-world tasks. When the probe's predicted failure score rises above the threshold, the VLA policy is automatically paused. For the simulated bowl task, we find that nudging the end-effector toward the drop-off location can help lead to success. For the real-world mushroom task, we find that nudging the wrist camera towards the target grasp object can help. Future work could systematically explore steering strategies. Refer to Figure \ref{fig:steering_strat} for more steering strategies and Figure \ref{fig:basket_steering} for another real world steering example. }
    \label{fig:combined_steering}
\end{figure*}

\subsection{Practical Considerations}
\label{subsec:efficiency}
Any real-world system must balance performance ({\footnotesize (T.)} Timeliness, {\footnotesize (C.)} Calibration, {\footnotesize (I.)} Informativeness) with the cost of building and running it ({\footnotesize (E.)} Efficiency). We first evaluate the cost of labeling the dataset, then of training the probe, calibrating it, and running it at inference time.

\paragraph{Dataset.} Labeling costs about one cent per failed episode (under \$50 for all annotations, including queries not used in this paper). Labels are therefore cheap, and practitioners will often already have unlabeled rollouts from system deployment.

\paragraph{Training.} \ours{} trains in $6.8$\,s on one RTX 4090, against $43.5$\,s for SAFE-LSTM, and holds $6.6$M parameters ($26.5$\,MB), about $0.2\%$ of the VLA. It is therefore cheap enough to retrain whenever the policy or the task set changes, or to train online as new rollouts are produced.

\paragraph{Calibration.} Calibrating takes under $2$\,ms per target pair on a CPU, so \ours{} can be re-calibrated if system requirements change, or online along with the probe.

\paragraph{Inference.} The VLA outputs a chunk of $m$ actions per call ($8$ at $20$\,Hz on LIBERO-X), so a policy step lasts $400$\,ms, and the probe reads the embedding of each call. It adds $0.12$\,ms per step on the GPU ($0.2\%$ of the $54$\,ms GR00T policy call) and $0.9$\,ms on four CPU threads, so its score is available before the chunk is executed and its latency is not observable: the step period is about $7\times$ the combined VLA and probe time in simulation and about $2\times$ on the real robot on our GPU. 

Faster baselines offer diminishing benefit, since the policy frequency bounds how fast a detector needs to be, while slower ones such as k-NN and STAC add noticeable latency to every step. \ours{} costs about the same as the cheapest learned baselines and far less than k-NN and STAC, while providing improved \emph{informativeness} and \emph{timeliness} on most unseen settings, which makes it a practical option for real-world deployment.

\paragraph{Answer to Q4. Efficiency.} Yes: labeling, training, calibrating, and running the probe introduce minimal cost, and on a GPU the probe adds under $1\%$ to the policy call and fits within the step period.

\vspace{0.4 cm}

\subsection{Steering Demonstration}
\label{subsec:steering}

We demonstrate how the four properties of \ours{}: (T.) timeliness, (C.) calibration, (I.) informativeness, and (E.) efficiency, can jointly support failure prevention through VLA steering. Our goal is to illustrate how an effective failure predictor can provide the signals needed to guide interventions rather than develop a steering method itself. We consider a steering architecture in which an expert model (e.g. a VLM) can invoke failure-recovery tools in response to the probe's predictions. For our experiments, we implement these tools using failure-mode-specific heuristics in simulation and demonstrate corresponding interventions on a real robot. 

For example, in the simulated rollout shown in Figure \ref{fig:sim_bowl_steering}, the task is ``place the largest bowl inside the left compartment of the bowl drainer." The probe raises an alarm before the bowl is mistakenly placed in the compartment on the right. After steering the end-effector towards the left side, the policy is resumed and the rollout succeeds. Without intervention, the bowl would be placed on the right side and the rollout fails. 

Our real-world experiments similarly demonstrate that the joint properties of \ours{} help in steering a VLA away from failure. Figure \ref{fig:mushroom_steering} illustrates this with the ``put the mushroom on the plate’' task. Without \ours{}, the gripper misses the mushroom and hits the table. With it turned on, however, the probe predicts the imminent grasp failure (which would likely happen), and corrective action is taken. In the example, the corrective action is to pause the VLA, let an expert model shift trajectory to a successful one, and resume the VLA.

Such a strategy is possible with a failure detector like \ours{}, which \textit{anticipates} the failure before it happens, rather than after it is too late. Our approach is agnostic towards the steering architecture. We leave exploration of robust steering strategies for future work. Nevertheless, these examples demonstrate the applicability of \ours to be utilized in safe and principled robotic systems. 

\textbf{Answer to Q5. Steering.} Yes: Our proof-of-concept demonstrations show how \ours{} can provide the timing and failure mode information needed for targeted interventions during VLA execution. Timeliness creates an opportunity to intervene, calibration controls when interventions are triggered, informativeness guides which intervention is applied, and efficiency allows the probe to operate alongside the VLA. 

\section{Conclusion}
We introduced \ours{}, a lightweight probe on the latent layers of a VLA that is \emph{timely}, \emph{calibrated}, \emph{informative}, and \emph{efficient}, making it well suited for real-world deployment. It is \emph{timely}: supervising on the pre-failure part of the trajectory drives early detection, and the probe's performance compared to the strongest learned baseline increases as the shift from the training tasks increases. It is \emph{calibrated}: calibrating the FPR and the $\tprearly$ jointly provides guarantees that hold in far more runs than the baselines under shift. It is \emph{informative}: beyond a failure score, it names the likely failure mode. It is \emph{efficient}: the probe adds under 1\% to the cost of a policy call. Our steering demonstrations, in simulation and on a real robot, show that \ours{} can inform targeted interventions, and we release 3,180 annotated episodes to support further work on latent failure detection.

\paragraph{Limitations.} First, our calibration guarantees rely on exchangeability between the calibration and test episodes, which weakens as the distribution shift grows. Empirically, we observe that the degradation is gradual. Second, we do not analyze the choice of source backbone embedding systematically. Third, the steering component, in which an expert model intervenes on a predicted failure, is heuristic and not evaluated in the scope of this work.

\paragraph{Future Work.} Looking forward, first, we will adapt the probe and its calibration guarantees online, which our preliminary experiments suggest is computationally feasible. Second, we will maintain and extend our embedding dataset with additional backbones and action-expert layers to determine which layers contain the most generalizable failure prediction signal. Third, we will explore agentic frameworks as the expert model in our setup, combining fast and local inference of a VLA with general-purpose intervention and steering from a cloud-based LLM, triggered by a principled detector such as \ours{}.

% \clearpage
\bibliography{references-2}
\bibliographystyle{mlsys2025}

% \clearpage
\appendix

% Appendix (new draft). Written from the code (vla-probe) with the tables and figures of the previous appendix; the tables keep their numbers and have shorter captions.
% black: numbers that are new or changed relative to the previous appendix. Red: still to fill in. Sources of the numbers are in the % comments.

\section{Method Details}
\label{app:method}

This section gives the details behind \autoref{sec:methodology}.

\subsection{Probe and Training}
\label{app:probe}

\paragraph{Inputs.} At every policy step we average the backbone tokens of the agent camera, of the wrist camera, and of the (unpadded) language prompt, and concatenate the three means into $e_t \in \mathbb{R}^{6144}$. No other feature of the backbone is used. All models in this paper read this vector, the baselines included.

\paragraph{Probe.} A single-layer LSTM with 256 hidden units reads $e_{1:t}$. Three heads read its state, each an MLP with one hidden layer of 64 ReLU units: the failure score head (one sigmoid output), the failure mode head (logits over the 10 modes), and the semantic head (a 64-dimensional vector).

\paragraph{Semantic (CLIP) head.} The target space of the semantic head is a PCA of the CLIP text embeddings of the timeline phrases (\texttt{clip-vit-base-patch32}), reduced to 64 dimensions with a PCA fitted on the training split only. The loss is InfoNCE with temperature 0.07: the normalized output of a step must pick the target of the step's phrase among the targets of all phrases of the training split. A step whose phrase is not among them is left out of this term. At inference the head is only a 64-dimensional MLP output; no CLIP encoder runs.
% fp/train/semantic.py, fp/train/losses.py

\paragraph{Loss and training.} $\mathcal{L}_{\text{score}}$ is a binary cross-entropy over the steps that are not excluded, weighted by the ratio of negative to positive steps of the training set. $\mathcal{L}_{\text{mode}}$ is a cross-entropy with class weights that balance the modes, on the positive steps of failed rollouts. Both auxiliary terms have weight 1. Early stopping follows $\mathcal{L}_{\text{score}}$ on the validation episodes only, so the auxiliary heads never choose the checkpoint.

\paragraph{Labeling window.} The window $w$ of the \emph{onset} and \emph{post-onset} labels, and the window of the mode read-out (\autoref{app:metrics}), is 2.4\,s: 6 policy steps for GR00T on LIBERO, 20 on the real robot, and 10 steps (2.5\,s) for $\pi_{0.5}$.
% fp/config.py window_steps

\subsection{Supervision Strategies}
\label{app:supervision}

Let $e_t$ be the backbone embedding at step $t$ and $y_t \in \{0, 1, \varnothing\}$ its target, where $y_t = \varnothing$ excludes the step from the loss. Let $Y_i \in \{0, 1\}$ be the episode label (success, failure) and $w$ the width of the window around $t_{\mathrm{fail}}$. For a failed rollout ($Y_i = 1$) the four strategies of \autoref{fig:supervision} are

{\small
\begin{align*}
    \text{\emph{Pre-onset.}}\quad y_t &= \begin{cases} 1 & t \leq t_{\mathrm{fail}}, \\ \varnothing & t > t_{\mathrm{fail}}, \end{cases} \\[1em]
    \text{\emph{Onset.}}\quad y_t &= \begin{cases} 1 & t \in [t_{\mathrm{fail}} - w,\, t_{\mathrm{fail}}], \\ 0 & t < t_{\mathrm{fail}} - w, \\ \varnothing & t > t_{\mathrm{fail}}, \end{cases} \\[1em]
    \text{\emph{Post-onset.}}\quad y_t &= \begin{cases} 1 & t \geq t_{\mathrm{fail}} - w, \\ 0 & t < t_{\mathrm{fail}} - w, \end{cases} \\[1em]
    \text{\emph{Trajectory.}}\quad y_t &= 1 \quad \text{for all } t.
\end{align*}}

In all four, $y_t = 0$ for every step of a successful rollout. Excluding a step ($y_t = \varnothing$) is the same as giving it zero loss. \emph{Trajectory} is the labeling of SAFE.

\subsection{Calibration and Guarantees}
\label{app:calibration}

\paragraph{Alarm rule and risks.} For a constant threshold $h$ the alarm time of an episode is $t_{\mathrm{det}}(h) = \min\{t : s_t \geq h\}$. The $\mathrm{FPR}(h)$ is the probability that a successful episode has an alarm, and $\mathrm{TPR}_{\mathrm{early}}(h)$ the probability that a failed episode has its first alarm at or before $t_{\mathrm{fail}}$.

\paragraph{Learn-then-Test.} We follow LTT~\citep{angelopoulosLearnThenTest2022} with two risks, the FPR and the early miss rate $1-\mathrm{TPR}_{\mathrm{early}}$ (Section~2.4 there). A threshold $h$ is tested against the null that it violates either target, and its p-value is the larger of the two p-values (Proposition~6). Both losses are binary, so each p-value is an exact binomial tail (Section~3.2): the number of calibration failures without an alarm at or before $t_{\mathrm{fail}}$ is compared with $\mathrm{Bin}(n_{\mathrm f}, 1-\beta)$, and the number of calibration successes with any alarm with $\mathrm{Bin}(n_{\mathrm s}, \alpha)$. We combine the p-values by multi-start fixed-sequence testing (Section~2.3.1, Algorithm~1, Proposition~3) over a grid of 181 thresholds $\mathrm{sigmoid}(z)$, $z \in [-9, 9]$, ordered from low to high, with $S = 40$ starts and $\delta = 0.1$, i.e.\ a level of 0.0025 per walk. By Theorem~1 of~\citet{angelopoulosLearnThenTest2022}, the set $\hat\Lambda$ of accepted thresholds satisfies $\mathbb{P}\big(\sup_{h \in \hat\Lambda}\{R(h)\} \le \alpha\big) \ge 1-\delta$, where $R$ stands for each of our two risks with its target; that is, with probability at least $1-\delta$ every accepted threshold meets both targets, for episodes exchangeable with the calibration episodes. We select the accepted threshold with the highest calibration $\mathrm{TPR}_{\mathrm{early}}$ (ties: lowest calibration FPR, then lowest threshold), and abstain if none is accepted.

\paragraph{Max-conformal (MC).} For STAC (whose accumulated scores are non-negative, so the maximum over steps is the final score), for ActProbe (as in its paper), and for our probe in the main table, we use max-conformal prediction. It takes the maximum score $M_i$ of each of the $n$ calibration successes and sets the constant threshold to the $k$-th smallest of them, with $k = \lceil (n+1)(1-\alpha) \rceil$, and to $+\infty$ if $k > n$. For a new success exchangeable with the calibration successes, its maximum has a uniformly distributed rank, so the false-alarm probability is at most $\alpha$, and conservative by less than $1/(n+1)$. It says nothing about the $\mathrm{TPR}$. 
% fp/calibration/max_conformal.py

\paragraph{Functional conformal (FC).} For SAFE and the methods that follow its calibration (FailureSpot and the embedding-distance baselines), we use the one-sided functional conformal band of SAFE~\citep{gu_safe_2025}. The band is a time-varying threshold built from the score trajectories of the calibration successes: the mean trajectory gives its shape, the local spread its width over time, and a conformal quantile of the successes' deviations sets its level so that a success stays below it with probability about $1-\alpha$.
% fp/calibration/safe_band.py, docs/CALIBRATION.md

\paragraph{Which method gets which calibrator.} SAFE-MLP, SAFE-LSTM, FailureSpot, and the three embedding-distance baselines use the SAFE band, as in their own work. STAC, ActProbe, and our probe use max-conformal.

\section{Datasets and Labeling}
\label{app:dataset}

\subsection{Benchmarks and Splits}
\label{app:benchmarks}

\paragraph{LIBERO-10 with occlusions.} We use the 10-suite of LIBERO-Occ~\citep{li_libero-occ_2026}, an extension of LIBERO~\citep{liu_libero_2023} that occludes the agent-view camera. Both policies fail rarely on the original benchmark, so we mix occluded and unoccluded episodes in equal parts to obtain enough failures (250 of each per policy).

\paragraph{LIBERO-X.} LIBERO-X~\citep{wang_libero-x_2026} introduces five levels of distribution shift. We use levels L1--L3 and select 218 tasks that cover diverse skills. \autoref{tab:libero_split_counts} gives the task counts of the splits.

\paragraph{Real robot.} We collected demonstrations for 8 tabletop manipulation tasks on a Kinova Gen3 arm (\autoref{tab:eight_tasks}) and fine-tuned GR00T on them. The evaluation data are 399 rollouts of the fine-tuned policy (49 to 50 per task), \textcolor{black}{254} of them failures; \autoref{tab:failure_distributions} gives their failure modes.

\begin{table}[H]
  \centering
  \caption{The eight real-world tasks and the number of demonstrations used to fine-tune the policy (\textcolor{black}{510} in total).}
  \label{tab:eight_tasks}
  \begingroup
  \setlength{\tabcolsep}{4pt}
  \renewcommand{\arraystretch}{0.9}
  \begin{tabular*}{\linewidth}{@{\extracolsep{\fill}}>{\raggedright\arraybackslash}m{0.45\linewidth}
                      >{\centering\arraybackslash}m{0.3\linewidth}
                      >{\centering\arraybackslash}m{0.15\linewidth}@{}}
    \toprule
    \textbf{Task instruction} & \textbf{Third-person view} & \textbf{Demos} \\
    \midrule
    Close the drawer &
      \includegraphics[width=1.3cm]{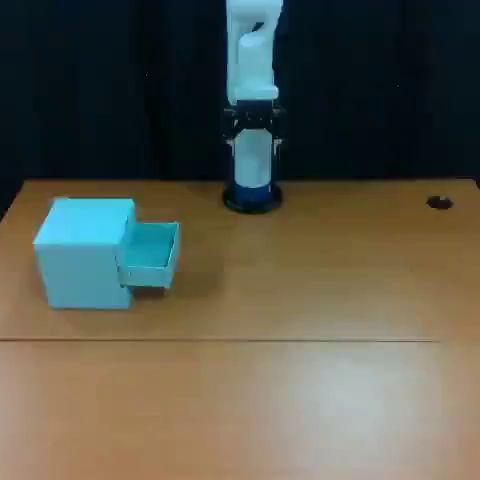} & 50 \\
    \addlinespace[1pt]
    Flip the basket upright &
      \includegraphics[width=1.3cm]{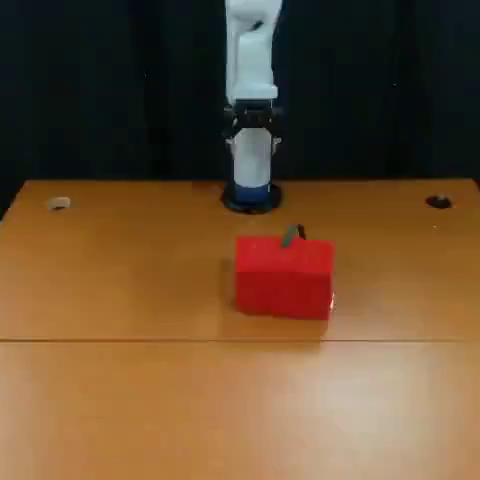} & 50 \\
    \addlinespace[1pt]
    Push the pineapple off the stove &
      \includegraphics[width=1.3cm]{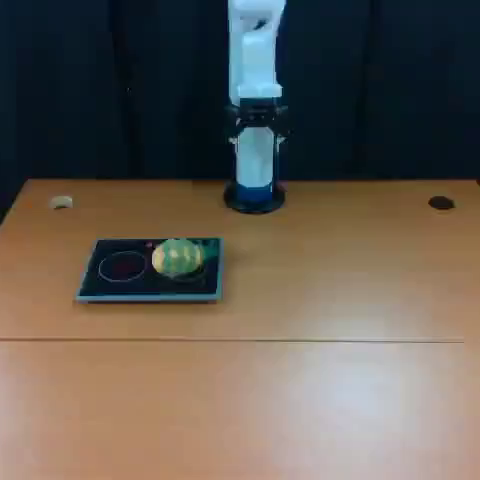} & 100 \\
    \addlinespace[1pt]
    Put the mushroom on the plate &
      \includegraphics[width=1.3cm]{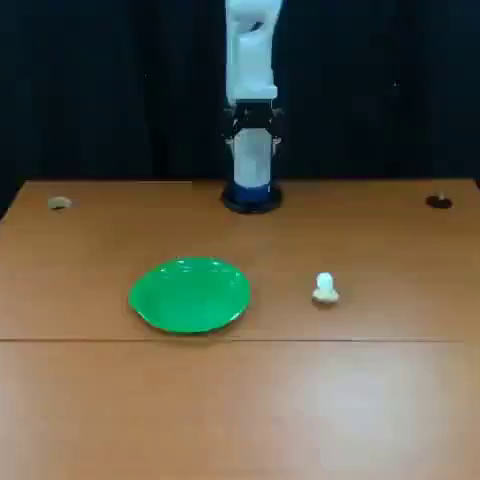} & 50 \\
    \addlinespace[1pt]
    Put the tomato in the bin &
      \includegraphics[width=1.3cm]{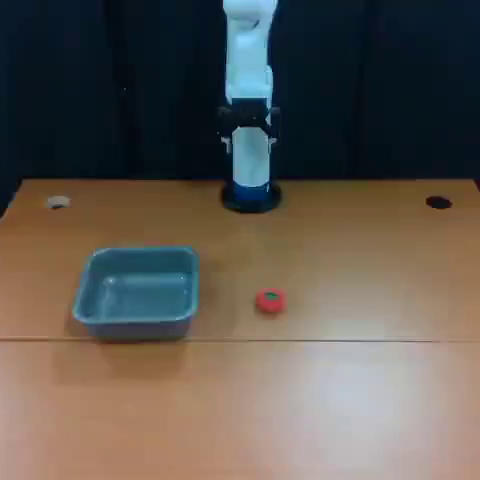} & 100 \\
    \addlinespace[1pt]
    Stack the yellow block on the wooden block &
      \includegraphics[width=1.3cm]{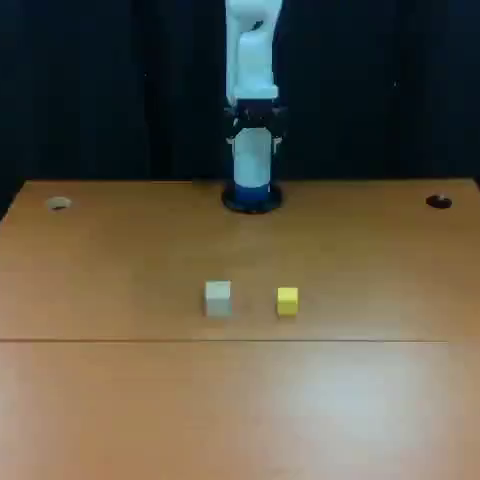} & 60 \\
    \addlinespace[1pt]
    Take the eggplant off the plate &
      \includegraphics[width=1.3cm]{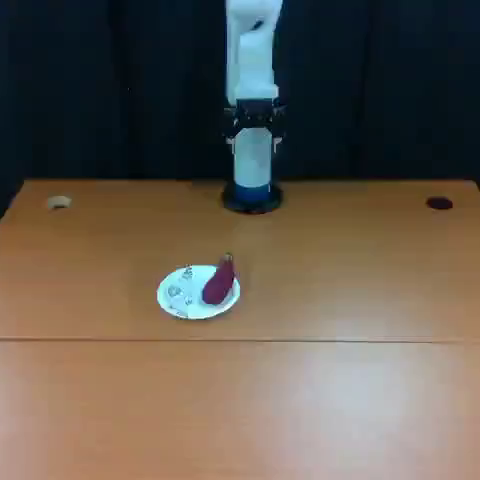} & 50 \\
    \addlinespace[1pt]
    Take the pineapple out of the drawer &
      \includegraphics[width=1.3cm]{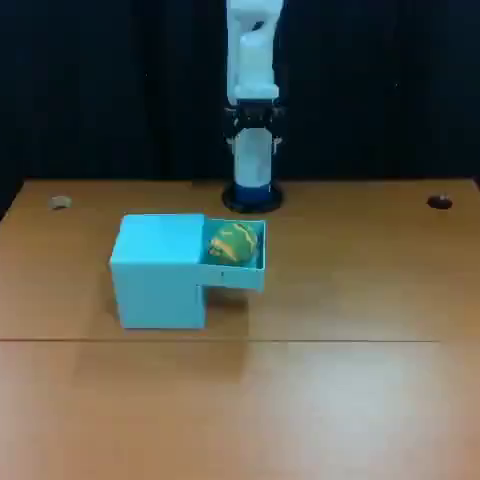} & 50 \\
    \bottomrule
  \end{tabular*}
  \endgroup
\end{table}

\paragraph{Construction of the splits.} The episodes of the held-out tasks form the unseen set. The episodes of the seen tasks are split 65/15/20\% into train, validation, and test. The test part is halved, into the calibration set and the held-out seen set. The calibration sets of the LIBERO-X runs hold 158 to 209 episodes (66 to 90 successes and 92 to 120 failures). On the real robot each run holds one task out, so the calibration part has only 12 to 14 successes, a max-conformal threshold exists only for $\alpha$ of about \textcolor{black}{0.07} and above, and the curves of all methods start at an FPR of about \textcolor{black}{8\%}. The area up to 40\% FPR is still defined and treats every method the same, but it is noisy. The seed changes the shuffle of the seen episodes and the initialization of the learned models; for the new-task and new-signature splits the unseen episodes are identical in the three seeds, so their spread is not a spread over different unseen tasks.
% docs/SPLITS.md, fp/data/splits.py

\subsection{Labeling Pipeline}
\label{app:labeling}

We label each rollout with Gemini (\texttt{gemini-3.5-flash}) from its video, in three passes. A coarse pass over the whole video of a failed rollout returns the failure mode and a failure time; a second pass on a slowed clip around that time refines the failure time to a single frame; and a third pass on the same video session gives the \emph{timeline}: a short keyword phrase (2--6 words, e.g.\ ``reaching for mug'') for each 3-second window, which provides the targets of the semantic head. The failure time $t_{\mathrm{fail}}$ is the earliest point at which the mistake responsible for the failure becomes apparent. Following~\cite{gu_safe_2025}, if there is no single mistake but a human operator would take over (for example when the robot is stuck), we use that time, and if the robot behaves plausibly but is too slow, the failure time is the end of the rollout.

The failure mode is one of 10 classes:
\emph{wrong\_object} (acts on an object or control other than the one the task specifies);
\emph{wrong\_target} (right object, wrong destination);
\emph{grasp\_failure} (never acquires the correct object securely);
\emph{object\_displacement} (knocks over, pushes away, or drops an object);
\emph{placement\_or\_insertion\_failure} (right object and target, wrong final pose);
\emph{open\_close\_failure} (right container, but not the required open or closed state);
\emph{press\_failure} (right button or knob, but no effect);
\emph{stuck\_or\_no\_progress} (looping, oscillating, hovering, or freezing);
\emph{timeout\_or\_insufficient\_progress} (purposeful but too slow, no identifiable mistake);
and \emph{other}. The full definitions and the prompts are on the project page, \href{https://vla-tug.github.io}{vla-tug.github.io}, and on the card of each dataset.

\subsection{Dataset Statistics and Release}
\label{app:statistics}

\autoref{tab:failure_distributions} gives, for each policy and dataset, the share of each failure mode among the failures and the mean and standard deviation of its failure time.

\begin{table*}[htb]
\centering
\caption{Failure-mode distribution and mean $\pm$ std failure time (s) per mode, by policy and dataset.}
\label{tab:failure_distributions}
\scriptsize
\setlength{\tabcolsep}{4pt}
\begin{tabular}{l cc cc cc cc}
\toprule
 & \multicolumn{2}{c}{$\pi_{0.5}$ LIBERO-10-Occ} & \multicolumn{2}{c}{$\texttt{GR00T}$ LIBERO-10-Occ} & \multicolumn{2}{c}{$\texttt{GR00T}$ LIBERO-X} & \multicolumn{2}{c}{$\texttt{GR00T}$ real robot} \\
\cmidrule(lr){2-3} \cmidrule(lr){4-5} \cmidrule(lr){6-7} \cmidrule(lr){8-9}
Mode & \% of failures & Time (s) & \% of failures & Time (s) & \% of failures & Time (s) & \% of failures & Time (s) \\
\midrule
stuck\_or\_no\_progress    & 24.5 & $10.3\pm3.5$ & 30.2 & $8.6\pm3.7$ & 14.1 & $15.5\pm10.7$ & \textcolor{black}{12.2} & \textcolor{black}{$3.8\pm3.0$} \\
object\_displacement       & 18.2 & $11.9\pm5.3$ & 17.8 & $11.8\pm6.7$ & 10.8 & $15.1\pm10.9$ & \textcolor{black}{9.4} & \textcolor{black}{$4.9\pm2.3$} \\
placement\_or\_insertion   & 13.6 & $15.2\pm4.2$ & 16.0 & $12.9\pm7.0$ & 9.9 & $21.9\pm10.2$ & \textcolor{black}{11.4} & \textcolor{black}{$8.3\pm4.5$} \\
grasp\_failure             & 17.3 & $9.6\pm5.4$ & 13.0 & $8.6\pm4.3$ & 9.0 & $12.0\pm9.7$ & \textcolor{black}{56.7} & \textcolor{black}{$4.7\pm2.3$} \\
open\_close\_failure       & 7.3 & $12.6\pm4.8$ & 9.5 & $13.9\pm3.1$ & 4.7 & $18.7\pm11.5$ & \textcolor{black}{7.5} & \textcolor{black}{$3.8\pm1.7$} \\
wrong\_target              & 13.6 & $6.8\pm4.9$ & 8.9 & $4.9\pm4.4$ & 12.3 & $14.3\pm9.7$ & \textcolor{black}{0.0} & \textcolor{black}{--} \\
wrong\_object              & 0.0 & -- & 1.8 & $5.3\pm3.1$ & 37.0 & $5.8\pm7.2$ & \textcolor{black}{1.2} & \textcolor{black}{$2.0\pm0.8$} \\
press\_failure             & 2.7 & $4.4\pm1.4$ & 1.8 & $10.7\pm4.8$ & 1.5 & $26.8\pm18.5$ & \textcolor{black}{0.0} & \textcolor{black}{--} \\
timeout                    & 2.7 & $25.9\pm0.0$ & 1.2 & $25.9\pm0.0$ & 0.2 & $60.0\pm0.0$ & \textcolor{black}{1.6} & \textcolor{black}{$6.1\pm1.1$} \\
other                      & 0.0 & -- & 0.0 & -- & 0.4 & $8.6\pm7.4$ & \textcolor{black}{0.0} & \textcolor{black}{--} \\
\midrule
\textit{n failed / total} & \multicolumn{2}{c}{110/500} & \multicolumn{2}{c}{169/500} & \multicolumn{2}{c}{1258/\textcolor{black}{2180}} & \multicolumn{2}{c}{\textcolor{black}{254/399}} \\
\bottomrule
\end{tabular}
\end{table*}
% numbers: failure modes and onset frames in data/<dataset>/manifest.csv and targets.npz (failure time = onset frame / control rate)

We release the embeddings, the labels, the prompts, and the splits of the simulated datasets, $3{,}180$ episodes, at \href{https://vla-tug.github.io}{vla-tug.github.io}.

\section{Experimental Setup}
\label{app:setup}

\subsection{Baselines}
\label{app:failure-detection}

We re-implement all baselines on the same inputs and splits as our probe. Where a baseline has a configuration grid, we run the whole grid on every split and select, for every setting, the configuration with the highest mean area on the \emph{validation} episodes, never looking at the unseen tasks; ties go to the earlier configuration of the grid. LIBERO-10 and the real robot pool their runs into one choice per method. On LIBERO-X, SAFE uses one configuration for the three kinds of splits, as in its own protocol, and every other method chooses per kind of held-out set. Our probe uses one architecture in every setting.

\paragraph{Embedding distance.} Mahalanobis, Euclidean, and PCA-KMeans are SAFE's embedding baselines~\citep{gu_safe_2025}. The score of a step is $D_{\mathrm{succ}}(x) - D_{\mathrm{fail}}(x)$, where $D_c$ is the Mahalanobis distance to a Gaussian fitted to the steps of class $c$, the mean Euclidean distance to the $k$ nearest stored steps of that class (Euclidean, also called k-NN), or the distance to the nearest of a few $k$-means centroids in a PCA-reduced space (PCA-KMeans); it can be accumulated over the steps. We run a sweep of 26 configurations (Mahalanobis 2, Euclidean 6, PCA-KMeans 18).

\paragraph{Action consistency.} \emph{STAC}~\citep{agia_unpacking_2024} compares the unexecuted tail of the previous plan with the first actions of the new plan, using the $K=10$ plans the policy samples at every call and the maximum mean discrepancy with an RBF kernel; nothing is learned. We run \textcolor{black}{8} variants (MMD over 3, 6, or 7 action dimensions, with the step score or accumulated; and STAC-Single, SAFE's single-plan variant). \emph{ActProbe}~\citep{huang_actprobe_2026} is a small LSTM on two action signals and an embedding of the task instruction; we train it with the recipe of its repository and select its checkpoint with its own validation criterion. \emph{FailureSpot}~\citep{ma_failurespot_2026} has no public code, so we implemented it from the paper: a detector on the pooled features is trained on weak targets computed from action consistency and magnitude, and then fine-tuned on timestamp labels of the \textcolor{black}{15\%} most uncertain episodes. 

\paragraph{Latent.} SAFE-MLP and SAFE-LSTM~\citep{gu_safe_2025} are written from SAFE's code with its losses and its labels: every step of a failed rollout is positive, i.e.\ our \emph{trajectory} label. We run SAFE's learning-rate and regularization grid, \textcolor{black}{12} configurations per model. SAFE trains for a fixed number of epochs without validation; we stop early after \textcolor{black}{50} epochs without improvement of its validation loss (cap: \textcolor{black}{1{,}000} epochs) and keep the best epoch, so that all learned methods are treated alike.

\subsection{Metrics}
\label{app:metrics}

We define three events for a rollout: detection $\mathcal{A} = \{t_{\mathrm{det}} \leq t_{\max} \mid Y = 1\}$, early detection $\mathcal{A}_{\mathrm{early}} = \{t_{\mathrm{det}} \leq t_{\mathrm{fail}} \mid Y = 1\}$, and no alarm $\mathcal{N} = \{t_{\mathrm{det}} = \mathrm{null} \mid Y = 0\}$. We count $\mathcal{A}$ as a true positive (TP) and $\mathcal{A}_{\mathrm{early}}$ as an early true positive ($\mathrm{TP}_{\mathrm{early}}$); the complements are the false negatives. The $\mathrm{TPR}$ and $\mathrm{FPR}$ follow their standard definitions, and
\begin{equation}
    \mathrm{TPR}_{\mathrm{early}} = \frac{\mathrm{TP}_{\mathrm{early}}}{\mathrm{TP}_{\mathrm{early}} + \mathrm{FN}_{\mathrm{early}}},
    \label{eq:tpr-early}
\end{equation}
\begin{equation}
    \mathrm{AUROC}_{\mathrm{early}} = \textcolor{black}{\frac{1}{0.4}} \int_{0}^{0.4} \mathrm{TPR}_{\mathrm{early}}(\mathrm{FPR})\, d\mathrm{FPR},
    \label{eq:auroc-early}
\end{equation}
so that $\mathrm{AUROC}_{\mathrm{early}}$ is reported in \% and a detector with $\mathrm{TPR}_{\mathrm{early}} = 100\%$ everywhere scores 100.

\subsection{Full Results}
\label{app:full-results}

\begin{figure*}[t]
    \centering
    \includegraphics[width=\linewidth]{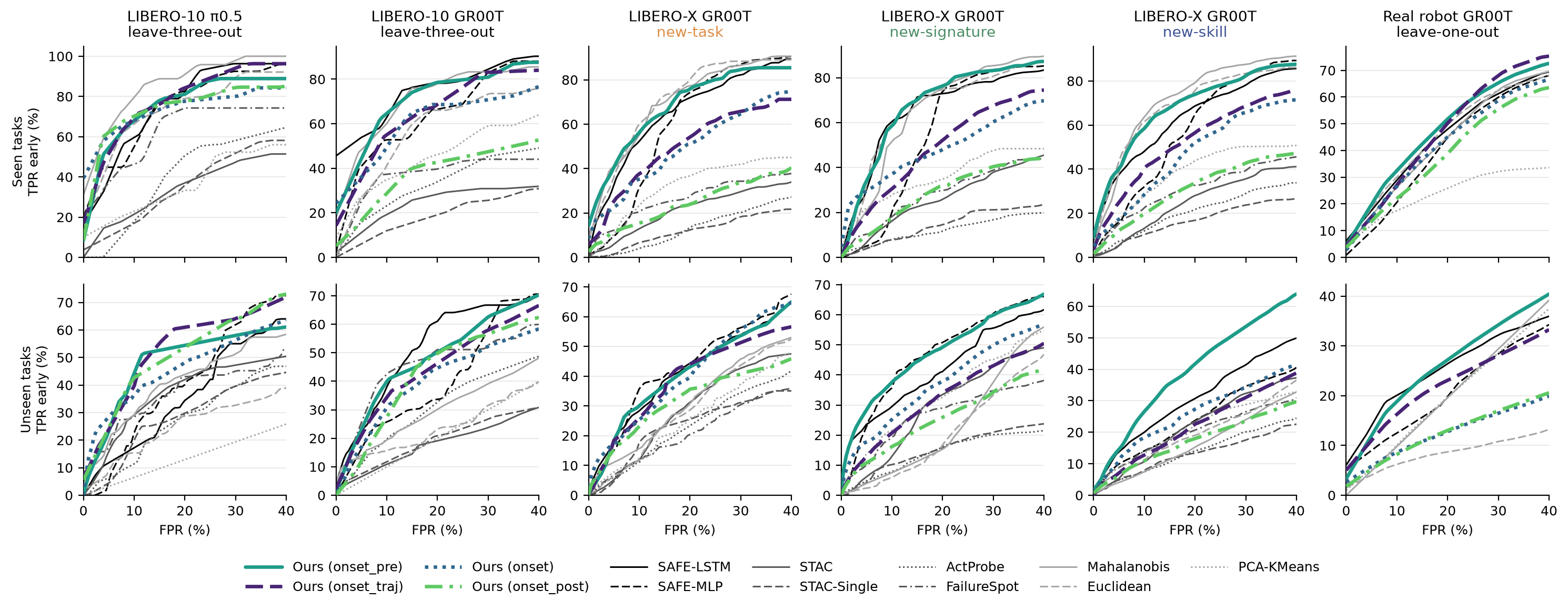}
    \caption{$\mathrm{TPR}_{\mathrm{early}}$ against FPR for every setting and method, on seen (top) and unseen (bottom) tasks; mean over the splits of a setting.}
    \label{fig:full_plot_auroc_early}
\end{figure*}

\autoref{tab:main_std} gives \autoref{tab:main} with the standard deviation over the splits, and \autoref{fig:full_plot_auroc_early} shows the curves of every setting and method. On the real robot the standard deviation is over the 40 splits (8 held-out tasks, 5 seeds).

\begin{table*}[t]
\centering
\caption{\autoref{tab:main} with the standard deviation over the splits: area under TPR early vs.\ FPR up to 40\% (in \%). Dashes: the action chunks the method needs were not saved. Best in bold, second best in italics.}
\label{tab:main_std}
\setlength{\tabcolsep}{4pt}
\resizebox{\textwidth}{!}{%
\begin{tabular}{llrrrrrrrrrrrrrrrr}
\toprule
 & & \multicolumn{2}{c}{LIBERO-10 $\pi_{0.5}$} & \multicolumn{2}{c}{LIBERO-10 GR00T} & \multicolumn{2}{c}{LIBERO-X task} & \multicolumn{2}{c}{LIBERO-X signature} & \multicolumn{2}{c}{LIBERO-X skill} & \multicolumn{2}{c}{Real robot} & \multicolumn{4}{c}{Cost (LIBERO-X)} \\
\cmidrule(lr){3-4}\cmidrule(lr){5-6}\cmidrule(lr){7-8}\cmidrule(lr){9-10}\cmidrule(lr){11-12}\cmidrule(lr){13-14}\cmidrule(lr){15-18}
Method & Cal. & Seen & Unseen & Seen & Unseen & Seen & Unseen & Seen & Unseen & Seen & Unseen & Seen & Unseen & Fit (s) & State (MB) & GPU (ms) & CPU (ms) \\
\midrule
SAFE-LSTM & FC & \textit{75.5{\scriptsize$\pm$13.4}} & 34.9{\scriptsize$\pm$3.5} & \textbf{74.1{\scriptsize$\pm$13.0}} & \textbf{51.2{\scriptsize$\pm$8.4}} & 64.6{\scriptsize$\pm$3.5} & \textit{40.4{\scriptsize$\pm$2.3}} & 64.7{\scriptsize$\pm$7.4} & 41.2{\scriptsize$\pm$7.1} & 62.5{\scriptsize$\pm$2.4} & \textit{30.1{\scriptsize$\pm$8.5}} & 44.5{\scriptsize$\pm$13.5} & \textit{25.1{\scriptsize$\pm$18.8}} & 43.5 & 26.2 & 0.066 & 0.86 \\
SAFE-MLP & FC & 74.1{\scriptsize$\pm$2.0} & 39.9{\scriptsize$\pm$15.1} & 62.9{\scriptsize$\pm$18.2} & 39.2{\scriptsize$\pm$9.4} & 59.3{\scriptsize$\pm$2.0} & \textbf{42.3{\scriptsize$\pm$2.9}} & 53.7{\scriptsize$\pm$2.4} & \textit{46.2{\scriptsize$\pm$2.3}} & 56.8{\scriptsize$\pm$6.0} & 23.5{\scriptsize$\pm$12.4} & 40.1{\scriptsize$\pm$15.9} & 20.2{\scriptsize$\pm$16.6} & 114.5 & 6.3 & \textit{0.034} & \textit{0.032} \\
STAC & MC & 33.8{\scriptsize$\pm$5.1} & 36.4{\scriptsize$\pm$5.3} & 24.2{\scriptsize$\pm$7.5} & 17.5{\scriptsize$\pm$9.8} & 21.0{\scriptsize$\pm$2.3} & 28.0{\scriptsize$\pm$0.1} & 25.6{\scriptsize$\pm$3.7} & 28.7{\scriptsize$\pm$0.3} & 24.6{\scriptsize$\pm$5.0} & 21.5{\scriptsize$\pm$8.0} & -- & -- & \textbf{0.0} & \textbf{0.0} & 21.30 & -- \\
ActProbe & MC & 39.3{\scriptsize$\pm$11.3} & 26.7{\scriptsize$\pm$2.6} & 32.4{\scriptsize$\pm$16.8} & 30.1{\scriptsize$\pm$9.3} & 12.6{\scriptsize$\pm$4.2} & 22.7{\scriptsize$\pm$4.6} & 11.9{\scriptsize$\pm$3.1} & 13.9{\scriptsize$\pm$1.5} & 20.3{\scriptsize$\pm$9.4} & 13.6{\scriptsize$\pm$7.1} & -- & -- & 7.1 & \textit{0.09} & 0.15 & 0.11 \\
FailureSpot & FC & 61.9{\scriptsize$\pm$3.0} & 36.2{\scriptsize$\pm$21.0} & 36.9{\scriptsize$\pm$19.8} & 44.7{\scriptsize$\pm$18.6} & 26.3{\scriptsize$\pm$2.1} & 20.4{\scriptsize$\pm$1.2} & 28.0{\scriptsize$\pm$2.5} & 26.2{\scriptsize$\pm$5.6} & 30.4{\scriptsize$\pm$3.4} & 19.6{\scriptsize$\pm$7.9} & -- & -- & 9.2 & 6.4 & \textbf{0.033} & \textbf{0.029} \\
Mahalanobis & FC & \textbf{85.7{\scriptsize$\pm$4.1}} & \textit{40.2{\scriptsize$\pm$15.5}} & \textit{70.7{\scriptsize$\pm$6.4}} & 29.2{\scriptsize$\pm$8.7} & \textit{69.6{\scriptsize$\pm$6.2}} & 30.6{\scriptsize$\pm$2.2} & 64.4{\scriptsize$\pm$3.0} & 22.3{\scriptsize$\pm$4.0} & \textbf{71.2{\scriptsize$\pm$3.1}} & 15.9{\scriptsize$\pm$9.6} & 43.0{\scriptsize$\pm$16.5} & 19.6{\scriptsize$\pm$2.5} & 15.2 & 302.0 & 0.41 & 5.37 \\
Euclidean & FC & 73.1{\scriptsize$\pm$4.4} & 24.2{\scriptsize$\pm$10.6} & 56.0{\scriptsize$\pm$6.8} & 23.1{\scriptsize$\pm$11.4} & \textbf{70.5{\scriptsize$\pm$4.2}} & 28.5{\scriptsize$\pm$3.0} & \textit{65.4{\scriptsize$\pm$2.1}} & 19.7{\scriptsize$\pm$0.8} & \textit{68.5{\scriptsize$\pm$4.8}} & 19.0{\scriptsize$\pm$6.5} & \textit{45.1{\scriptsize$\pm$14.2}} & 8.3{\scriptsize$\pm$12.8} & \textit{0.4} & 3056 & 13.19 & 368.83 \\
PCA-KMeans & FC & 35.7{\scriptsize$\pm$12.2} & 12.9{\scriptsize$\pm$5.1} & 44.5{\scriptsize$\pm$8.7} & 20.0{\scriptsize$\pm$0.0} & 31.7{\scriptsize$\pm$3.4} & 25.0{\scriptsize$\pm$1.2} & 35.3{\scriptsize$\pm$1.8} & 32.5{\scriptsize$\pm$1.7} & 37.8{\scriptsize$\pm$7.7} & 20.7{\scriptsize$\pm$7.1} & 24.0{\scriptsize$\pm$11.4} & 19.4{\scriptsize$\pm$7.7} & 39.9 & 1.6 & 0.12 & 0.055 \\
\midrule
\rowcolor{gray!15} \ours{} (Ours) & MC & 75.1{\scriptsize$\pm$12.5} & \textbf{47.9{\scriptsize$\pm$20.9}} & 70.5{\scriptsize$\pm$10.4} & \textit{47.4{\scriptsize$\pm$13.5}} & 67.9{\scriptsize$\pm$5.0} & 40.2{\scriptsize$\pm$1.2} & \textbf{67.0{\scriptsize$\pm$6.6}} & \textbf{47.0{\scriptsize$\pm$0.6}} & 67.8{\scriptsize$\pm$4.8} & \textbf{39.0{\scriptsize$\pm$7.0}} & \textbf{47.6{\scriptsize$\pm$10.9}} & \textbf{25.8{\scriptsize$\pm$11.6}} & 6.8 & 26.5 & 0.12 & 0.94 \\
\bottomrule
\end{tabular}%
}
\end{table*}

\section{Results in Detail}
\label{app:results}

\subsection{Calibration and Guarantees}
\label{app:guarantees}

This section gives the details behind \autoref{subsec:uncertainty-guarantees}.

\paragraph{Setup.} We run this analysis on LIBERO-X only. LIBERO-10 has only a few tens of failures per split, and the real robot has only 12 to 14 calibration successes per run, too few for LTT to certify any $\mathrm{TPR}_{\mathrm{early}}$ target together with an FPR target in at least half of the runs; its LTT rows would be empty. The 14 runs are the 3 \newtask{}, 3 \newsig{}, and 8 \newskill{} splits. We use the FPR targets 10, 15, 20, 25, 30, 40, and 50\%. For each of them, we report the highest $\mathrm{TPR}_{\mathrm{early}}$ target that LTT certifies together with it in at least half of the runs, searched on a grid from \textcolor{black}{10\% to 95\%} in steps of 5\%. LTT uses $\delta = 0.1$ and $S = 40$ starting points for its fixed-sequence walks (\autoref{app:calibration}). \autoref{tab:ltt} gives for each FPR target the shares of runs in which the targets hold on the held-out seen and unseen episodes, the realized rates, and the PPV bound, for LTT and for the two calibrators of the baselines, max-conformal (MC) and functional conformal (FC).

\begin{table*}[t]
\centering
\caption{Learn-then-Test (LTT), max-conformal (MC), and functional conformal (FC) at the same FPR target, over the 14 LIBERO-X runs ($\delta = 0.1$). LTT certifies the TPR-early target of the second column together with the FPR target, and abstains otherwise; its percentages are over the certified runs. PPV bound: $\beta\pi/(\beta\pi + \alpha(1-\pi))$.}
\label{tab:ltt}
\setlength{\tabcolsep}{4pt}
\resizebox{1.0\textwidth}{!}{%
\begin{tabular}{llrlrrrrrrrrrrrrrrr}
\toprule
 & & & & & \multicolumn{7}{c}{Seen tasks} & \multicolumn{7}{c}{Unseen tasks} \\
\cmidrule(lr){6-12}\cmidrule(lr){13-19}
FPR target & TPR-early target & PPV bound & Method & Runs & FPR holds & TPR holds & FPR & TPR early & PPV & \shortstack{PPV\\holds} & \shortstack{PPV holds\\(own $\pi$)} & FPR holds & TPR holds & FPR & TPR early & PPV & \shortstack{PPV\\holds} & \shortstack{PPV holds\\(own $\pi$)} \\
\midrule
\rowcolor{gray!15} 10\% & -- & -- & LTT & 0\% (0/14) & -- & -- & -- & -- & -- & -- & -- & -- & -- & -- & -- & -- & -- & -- \\
 & -- & -- & Max conformal (MC) & 100\% (14/14) & 57\% & -- & 9.3{\scriptsize$\pm$4.4} & 53.4{\scriptsize$\pm$12.7} & 90.1 & -- & -- & 50\% & -- & 15.6{\scriptsize$\pm$12.3} & 35.3{\scriptsize$\pm$20.5} & 80.7 & -- & -- \\
 & -- & -- & Functional conformal (FC) & 100\% (14/14) & 71\% & -- & 9.8{\scriptsize$\pm$5.8} & 33.9{\scriptsize$\pm$17.2} & 90.7 & -- & -- & 36\% & -- & 14.9{\scriptsize$\pm$9.4} & 26.3{\scriptsize$\pm$16.2} & 84.3 & -- & -- \\
\midrule
\rowcolor{gray!15} 15\% & 15\% & 58.1 & LTT & 50\% (7/14) & 100\% & 100\% & 5.3{\scriptsize$\pm$3.1} & 39.2{\scriptsize$\pm$8.0} & 92.5 & 100\% & 100\% & 86\% & 71\% & 9.8{\scriptsize$\pm$11.5} & 27.1{\scriptsize$\pm$14.7} & 85.8 & 100\% & 100\% \\
 & -- & -- & Max conformal (MC) & 100\% (14/14) & 57\% & -- & 14.2{\scriptsize$\pm$7.3} & 64.8{\scriptsize$\pm$9.0} & 88.1 & -- & -- & 14\% & -- & 25.8{\scriptsize$\pm$12.3} & 50.7{\scriptsize$\pm$15.7} & 77.5 & -- & -- \\
 & -- & -- & Functional conformal (FC) & 100\% (14/14) & 64\% & -- & 13.2{\scriptsize$\pm$6.1} & 40.0{\scriptsize$\pm$13.9} & 88.9 & -- & -- & 29\% & -- & 21.9{\scriptsize$\pm$11.9} & 32.8{\scriptsize$\pm$18.5} & 81.4 & -- & -- \\
\midrule
\rowcolor{gray!15} 20\% & 35\% & 70.7 & LTT & 50\% (7/14) & 100\% & 100\% & 10.4{\scriptsize$\pm$3.5} & 58.1{\scriptsize$\pm$5.4} & 89.7 & 100\% & 100\% & 71\% & 57\% & 15.5{\scriptsize$\pm$10.4} & 37.0{\scriptsize$\pm$16.4} & 81.1 & 100\% & 100\% \\
 & -- & -- & Max conformal (MC) & 100\% (14/14) & 57\% & -- & 19.2{\scriptsize$\pm$7.2} & 72.9{\scriptsize$\pm$5.8} & 85.5 & -- & -- & 21\% & -- & 37.2{\scriptsize$\pm$13.8} & 61.8{\scriptsize$\pm$13.9} & 73.6 & -- & -- \\
 & -- & -- & Functional conformal (FC) & 100\% (14/14) & 79\% & -- & 18.0{\scriptsize$\pm$6.1} & 54.4{\scriptsize$\pm$11.2} & 86.6 & -- & -- & 21\% & -- & 30.7{\scriptsize$\pm$11.1} & 47.0{\scriptsize$\pm$16.7} & 77.9 & -- & -- \\
\midrule
\rowcolor{gray!15} 25\% & 35\% & 65.7 & LTT & 57\% (8/14) & 100\% & 100\% & 12.4{\scriptsize$\pm$3.7} & 64.5{\scriptsize$\pm$5.2} & 88.8 & 100\% & 100\% & 75\% & 75\% & 22.7{\scriptsize$\pm$11.6} & 49.2{\scriptsize$\pm$18.9} & 79.9 & 100\% & 100\% \\
 & -- & -- & Max conformal (MC) & 100\% (14/14) & 57\% & -- & 25.1{\scriptsize$\pm$8.1} & 80.0{\scriptsize$\pm$5.0} & 83.1 & -- & -- & 7\% & -- & 48.7{\scriptsize$\pm$14.3} & 72.1{\scriptsize$\pm$11.8} & 70.2 & -- & -- \\
 & -- & -- & Functional conformal (FC) & 100\% (14/14) & 79\% & -- & 22.7{\scriptsize$\pm$7.0} & 63.8{\scriptsize$\pm$12.0} & 84.7 & -- & -- & 7\% & -- & 42.5{\scriptsize$\pm$12.1} & 57.5{\scriptsize$\pm$16.1} & 74.3 & -- & -- \\
\midrule
\rowcolor{gray!15} 30\% & 45\% & 67.4 & LTT & 50\% (7/14) & 100\% & 100\% & 15.2{\scriptsize$\pm$3.5} & 68.2{\scriptsize$\pm$4.6} & 87.4 & 100\% & 100\% & 86\% & 57\% & 24.5{\scriptsize$\pm$6.6} & 50.6{\scriptsize$\pm$12.6} & 77.9 & 100\% & 100\% \\
 & -- & -- & Max conformal (MC) & 100\% (14/14) & 57\% & -- & 28.6{\scriptsize$\pm$7.9} & 83.1{\scriptsize$\pm$4.5} & 81.6 & -- & -- & 0\% & -- & 56.9{\scriptsize$\pm$13.4} & 77.3{\scriptsize$\pm$10.2} & 67.9 & -- & -- \\
 & -- & -- & Functional conformal (FC) & 100\% (14/14) & 71\% & -- & 27.8{\scriptsize$\pm$7.4} & 71.2{\scriptsize$\pm$11.1} & 82.5 & -- & -- & 7\% & -- & 54.1{\scriptsize$\pm$15.3} & 67.3{\scriptsize$\pm$15.0} & 70.7 & -- & -- \\
\midrule
\rowcolor{gray!15} 40\% & 55\% & 65.8 & LTT & 71\% (10/14) & 100\% & 100\% & 24.1{\scriptsize$\pm$7.3} & 78.4{\scriptsize$\pm$5.7} & 83.5 & 100\% & 100\% & 60\% & 80\% & 43.3{\scriptsize$\pm$14.6} & 66.6{\scriptsize$\pm$13.8} & 70.5 & 80\% & 100\% \\
 & -- & -- & Max conformal (MC) & 100\% (14/14) & 64\% & -- & 37.6{\scriptsize$\pm$6.2} & 87.5{\scriptsize$\pm$3.1} & 77.7 & -- & -- & 0\% & -- & 69.8{\scriptsize$\pm$14.1} & 84.9{\scriptsize$\pm$7.9} & 65.0 & -- & -- \\
 & -- & -- & Functional conformal (FC) & 100\% (14/14) & 64\% & -- & 38.7{\scriptsize$\pm$7.3} & 82.0{\scriptsize$\pm$8.1} & 77.8 & -- & -- & 0\% & -- & 74.0{\scriptsize$\pm$15.0} & 80.2{\scriptsize$\pm$8.1} & 65.0 & -- & -- \\
\midrule
\rowcolor{gray!15} 50\% & 65\% & 64.3 & LTT & 71\% (10/14) & 100\% & 100\% & 31.1{\scriptsize$\pm$5.4} & 84.5{\scriptsize$\pm$3.3} & 80.2 & 100\% & 100\% & 40\% & 90\% & 59.2{\scriptsize$\pm$18.7} & 77.6{\scriptsize$\pm$10.7} & 67.7 & 70\% & 60\% \\
 & -- & -- & Max conformal (MC) & 100\% (14/14) & 71\% & -- & 46.1{\scriptsize$\pm$6.2} & 92.1{\scriptsize$\pm$2.2} & 74.6 & -- & -- & 0\% & -- & 79.2{\scriptsize$\pm$13.2} & 89.8{\scriptsize$\pm$6.8} & 63.0 & -- & -- \\
 & -- & -- & Functional conformal (FC) & 100\% (14/14) & 50\% & -- & 49.0{\scriptsize$\pm$7.2} & 90.6{\scriptsize$\pm$3.3} & 73.8 & -- & -- & 0\% & -- & 83.1{\scriptsize$\pm$12.6} & 87.2{\scriptsize$\pm$5.9} & 62.4 & -- & -- \\
\bottomrule
\end{tabular}%
}
\end{table*}

\paragraph{How to read \autoref{tab:ltt}.} \emph{Runs} is the share of runs for which LTT certifies the targets; the other columns of an LTT row are over these runs. \emph{Holds} is the share of runs in which the target is met on the evaluated episodes. The \emph{PPV bound} is $\beta\pi/(\beta\pi + \alpha(1-\pi))$ with $\pi$ the prevalence of the calibration set, and \emph{PPV holds} is the share of runs in which the realized PPV is at least the bound (computed with the calibration set's prevalence, or with that of the evaluated set).

\paragraph{PPV bound.} The bound follows from the two LTT targets, so for \ours{} it comes directly from the FPR and $\mathrm{TPR}_{\mathrm{early}}$ targets we calibrate for. It depends on the prevalence $\pi$, which we estimate on the calibration set. On seen tasks the estimate is stable. Under shift it can drift on individual \newskill{} folds, where the share of failures can differ a lot between the calibration set and the held-out tasks; this is what limits the bound on unseen tasks, and using the unseen set's own $\pi$ removes most of the misses. A moving average of $\pi$ over the observations at deployment time could deal with more severe shift.

\paragraph{Unseen tasks.} On the \newtask{} and \newsig{} splits, the three runs of each split share the same unseen tasks and differ only in the calibration data and the initialization. The held-out tasks define the shift each split is designed to test, so we cannot redraw them between runs. Only the eight \newskill{} folds hold out different tasks. The shares on unseen tasks therefore rest on fewer distinct unseen sets than runs, so they indicate trends and not precise rates.

\subsection{Informativeness}
\label{app:informativeness}

This section gives the details behind \autoref{subsec:informativeness}.

\paragraph{Classes.} The failure modes are strongly imbalanced, so we report a balanced accuracy: the mean over classes of the share of their failures predicted correctly. We merge the two smallest classes repeatedly, once per dataset, until every class holds at least \textcolor{black}{5\%} of the failures. The rule uses label frequencies only. The mode head is not retrained: a merged class gets the sum of the probabilities of its modes. The accuracy is computed on the predictions pooled over the splits of a setting, since a split has few failures of a rare class. We exclude LIBERO-10: it has too few failures for a per-class accuracy, since even after merging its smallest classes hold only \textcolor{black}{6 to 16} failures, so one failure changes a class's accuracy by several points.

\paragraph{Prediction.} A failed episode is predicted at an anchor step: the class with the highest mean probability over the 2.4\,s window before the anchor and the anchor itself (the top three classes for top-3). The anchor is either the probe's first alarm (max-conformal at $\alpha = 0.2$; episodes without an alarm are not scored) or the true failure time, which is an upper bound that a deployed system can reach. The baselines predict the most common training classes (always) or guess uniformly among the training classes.

\begin{table*}[t]
\centering
\caption{Balanced top-1 and top-3 failure-mode accuracy (\%), pooled over the splits of each kind. Shaded rows: our probe, by label variant; the oracle rows use the true failure time and are an upper bound. Best in bold, second best in italics, among the other rows.}
\label{tab:modes_balanced}
\setlength{\tabcolsep}{4pt}
\resizebox{0.78\textwidth}{!}{%
\begin{tabular}{lrrrrrrrrrrrrrrrr}
\toprule
 & \multicolumn{8}{c}{Top-1 accuracy (\%)} & \multicolumn{8}{c}{Top-3 accuracy (\%)} \\
\cmidrule(lr){2-9}\cmidrule(lr){10-17}
 & \multicolumn{2}{c}{task} & \multicolumn{2}{c}{signature} & \multicolumn{2}{c}{skill} & \multicolumn{2}{c}{Real robot} & \multicolumn{2}{c}{task} & \multicolumn{2}{c}{signature} & \multicolumn{2}{c}{skill} & \multicolumn{2}{c}{Real robot} \\
\cmidrule(lr){2-3}\cmidrule(lr){4-5}\cmidrule(lr){6-7}\cmidrule(lr){8-9}\cmidrule(lr){10-11}\cmidrule(lr){12-13}\cmidrule(lr){14-15}\cmidrule(lr){16-17}
Method & Seen & Unseen & Seen & Unseen & Seen & Unseen & Seen & Unseen & Seen & Unseen & Seen & Unseen & Seen & Unseen & Seen & Unseen \\
\midrule
Most common & 14.3 & 14.3 & 14.3 & 14.3 & 14.3 & 14.3 & 20.0 & \textit{20.0} & 42.9 & 42.9 & 42.9 & 42.9 & 43.9 & 39.0 & 60.4 & 32.6 \\
Random & 14.3 & 14.3 & 14.3 & 14.3 & 14.3 & 14.3 & 20.0 & \textit{20.0} & 42.9 & 42.9 & 42.9 & 42.9 & 42.9 & 42.9 & 60.0 & \textit{60.0} \\
\midrule
\multicolumn{17}{l}{\ours (Ours), own detection (first alarm)} \\
\rowcolor{gray!15} onset\_pre & \textit{34.2} & \textit{25.5} & \textit{35.6} & \textit{22.6} & \textit{33.6} & 20.3 & 54.3 & \textbf{21.7} & 65.1 & \textbf{59.9} & \textit{69.8} & 56.3 & \textit{69.4} & 56.3 & \textit{88.2} & \textbf{70.4} \\
\rowcolor{gray!15} onset\_traj & \textbf{39.3} & \textbf{25.9} & \textbf{37.1} & \textbf{29.9} & \textbf{39.7} & \textbf{27.0} & \textbf{57.2} & 15.8 & \textbf{71.8} & \textit{58.8} & \textbf{71.4} & \textbf{61.1} & \textbf{69.8} & \textbf{59.7} & 87.7 & 58.6 \\
\rowcolor{gray!15} onset & 28.0 & 19.3 & 32.8 & 19.5 & 25.0 & 17.2 & 54.9 & 13.2 & 63.8 & 54.6 & 59.3 & 50.4 & 56.0 & 48.1 & 88.0 & 57.8 \\
\rowcolor{gray!15} onset\_post & 31.1 & 19.8 & 32.1 & 22.1 & 32.5 & \textit{25.2} & \textit{56.6} & 18.8 & \textit{65.3} & 55.7 & 69.6 & \textit{58.0} & 65.2 & \textit{57.2} & \textbf{91.4} & 54.4 \\
\midrule
\multicolumn{17}{l}{\ours (Ours), oracle onset (upper bound)} \\
\rowcolor{gray!15} onset\_pre & 41.3 & 31.0 & 41.0 & 30.2 & 42.4 & 30.0 & 68.4 & 12.1 & 73.0 & 65.0 & 78.4 & 66.8 & 72.9 & 64.8 & 92.5 & 57.1 \\
\rowcolor{gray!15} onset\_traj & 36.1 & 29.3 & 39.8 & 29.9 & 39.1 & 27.4 & 66.1 & 12.4 & 69.5 & 59.3 & 73.3 & 57.1 & 71.5 & 59.4 & 93.8 & 53.1 \\
\rowcolor{gray!15} onset & 37.5 & 30.5 & 39.7 & 29.2 & 37.1 & 27.1 & 70.2 & 11.4 & 74.2 & 67.5 & 75.0 & 66.9 & 70.1 & 62.7 & 92.6 & 60.0 \\
\rowcolor{gray!15} onset\_post & 23.0 & 20.0 & 28.0 & 22.2 & 33.0 & 21.4 & 67.8 & 16.3 & 64.8 & 59.2 & 69.6 & 56.2 & 66.0 & 58.8 & 94.5 & 52.9 \\
\bottomrule
\end{tabular}%
}
\end{table*}

\subsection{Cost and Latency}
\label{app:efficiency}

This section gives the details behind \autoref{subsec:efficiency}.

\paragraph{How the VLA runs.} At each policy step the VLA receives the camera images and the prompt and produces a chunk of $m$ actions ($m=8$ on LIBERO-X, $5$ for $\pi_{0.5}$ on LIBERO-10, $6$ on the real robot), which the robot executes at its control rate ($20$\,Hz in simulation, $50$\,Hz on the real robot). A policy step therefore lasts $m$ divided by the control rate: $400$\,ms on LIBERO-X, $250$\,ms for $\pi_{0.5}$, and $120$\,ms on the real robot. The probe reads the embedding of each call and must return its score well within one step.

\paragraph{Measurement.} All times are measured on one RTX 4090 and 32 CPU threads. The fit time is the training time of the probe or the SAFE models, or the time to fit the statistics of the distance baselines. The state is what a fitted method must keep to score a new episode: the weights, the stored training steps of the k-NN, or the statistics of the distance baselines. The latency is the \emph{online} latency: the time to turn one new policy step into a score, one embedding at a time and with the recurrent state of the previous step, after a warm-up and with the device synchronized, as the median over up to 2{,}000 steps. The CPU is timed with four threads. Online latency is the one that matters for deployment.

\paragraph{Baselines.} The k-NN baseline has to keep every training step and compare each new step with all of them, so its cost grows with the training set; Mahalanobis keeps two $6144 \times 6144$ matrices. STAC needs the $K=10$ plans of the policy at every call, which no other method needs, so its latency is its scoring time plus the time to sample them.

\paragraph{Probe.} On the GPU the LSTM takes most of the time; the mode head and the CLIP head add little, and on the CPU they add nothing measurable. The probe is not the fastest online method (SAFE-LSTM and the MLP detectors are faster), but it is far below the duration of a policy step.

\paragraph{Calibration.} Per target pair, max-conformal takes $0.1$\,ms, functional conformal $0.7$\,ms, and LTT $1.3$\,ms on a CPU. The calibration sets hold $158$ to $209$ labeled episodes, of which $92$ to $120$ are failures.

\subsection{Ablations}
\label{app:ablation}

We ablate the supervision strategy, the auxiliary heads, and the failure time used for evaluation.

\paragraph{Supervision strategy.} \autoref{tab:ablations} compares our pre-onset label (every step before the failure time is positive) with the window before the onset only, every step of the failed rollout, and the window before the onset together with all steps after it. The pre-onset label is the best, or within the spread across splits, in nearly every setting. On LIBERO-X the alternatives lose $24\%$ to $50\%$ of the detection area on average, and are lower on every kind of held-out split, seen and unseen; labeling the steps after the onset as positive is worst. On LIBERO-10 the differences are smaller ($-1\%$ to $-14\%$ on average), within the spread across splits for $\pi_{0.5}$, and clear on seen tasks with GR00T.

\paragraph{Auxiliary heads.} We include the mode and CLIP heads because their labels already exist for every failure. We find no consistent benefit from either on detection (\autoref{tab:ablations}): removing the heads changes the area by less than the spread across splits in most settings. For mode prediction (\autoref{tab:mode_ablation}), the CLIP head changes top-1 accuracy by less than one point on average, while training the mode head jointly with the failure score adds three to four points over a mode-only probe. The CLIP alignment might help some subpopulations and hurt others, which we leave to future work.

\paragraph{Sensitivity to the failure time.} Moving the failure time later makes the metric easier, since more alarms then count as early, so we check how much our results depend on it. We score the saved scores again with the failure time of every failed episode moved $1$ and $2$\,s later, with the same thresholds and without retraining (\autoref{tab:onset_shift}). Every method gains, but late detectors gain more than \ours{}, which already alarms early: on unseen LIBERO-X \ours{} gains $2.1$ to $2.4$ points of area over $2$\,s and SAFE-LSTM $3.5$ to $4.0$. On LIBERO-X our rank among the methods does not change, and our margins hold, so these results do not depend on the exact failure time. On the real robot, where our lead is a single point, SAFE-LSTM overtakes \ours{} at $1$\,s for the same reason.

\begin{table*}[t]
\centering
\caption{Ablations of our probe in the simulated settings: area under TPR early vs.\ FPR up to 40\% (in \%), mean $\pm$ standard deviation over the splits, and in brackets the change relative to our probe (shaded). Best in bold, second best in italics, within each block.}
\label{tab:ablations}
\setlength{\tabcolsep}{4pt}
\resizebox{1.0\textwidth}{!}{%
\begin{tabular}{lrrrrrrrrrr}
\toprule
 & \multicolumn{2}{c}{LIBERO-10 $\pi_{0.5}$} & \multicolumn{2}{c}{LIBERO-10 GR00T} & \multicolumn{2}{c}{LIBERO-X task} & \multicolumn{2}{c}{LIBERO-X signature} & \multicolumn{2}{c}{LIBERO-X skill} \\
\cmidrule(lr){2-3}\cmidrule(lr){4-5}\cmidrule(lr){6-7}\cmidrule(lr){8-9}\cmidrule(lr){10-11}
 & Seen & Unseen & Seen & Unseen & Seen & Unseen & Seen & Unseen & Seen & Unseen \\
\midrule
\multicolumn{11}{l}{Heads (LSTM, pre-onset label)} \\
\rowcolor{gray!15} Modes + CLIP (ours) & 75.1{\scriptsize$\pm$12.5} & 47.9{\scriptsize$\pm$20.9} & 70.5{\scriptsize$\pm$10.4} & 47.4{\scriptsize$\pm$13.5} & \textit{67.9}{\scriptsize$\pm$5.0} & 40.2{\scriptsize$\pm$1.2} & 67.0{\scriptsize$\pm$6.6} & \textbf{47.0}{\scriptsize$\pm$0.6} & 67.8{\scriptsize$\pm$4.8} & 39.0{\scriptsize$\pm$7.0} \\
No head & \textbf{82.3}{\scriptsize$\pm$6.4} {\scriptsize(+10\%)} & \textit{51.1}{\scriptsize$\pm$4.0} {\scriptsize(+7\%)} & 70.1{\scriptsize$\pm$11.0} {\scriptsize($-$1\%)} & \textbf{55.1}{\scriptsize$\pm$5.5} {\scriptsize(+16\%)} & 67.1{\scriptsize$\pm$1.8} {\scriptsize($-$1\%)} & \textbf{42.7}{\scriptsize$\pm$1.5} {\scriptsize(+6\%)} & \textbf{68.3}{\scriptsize$\pm$5.7} {\scriptsize(+2\%)} & 45.0{\scriptsize$\pm$1.3} {\scriptsize($-$4\%)} & \textbf{69.2}{\scriptsize$\pm$4.4} {\scriptsize(+2\%)} & \textit{43.6}{\scriptsize$\pm$11.5} {\scriptsize(+12\%)} \\
Modes only & \textit{80.4}{\scriptsize$\pm$8.4} {\scriptsize(+7\%)} & 48.3{\scriptsize$\pm$8.8} {\scriptsize(+1\%)} & \textit{70.9}{\scriptsize$\pm$8.2} {\scriptsize(+0\%)} & \textit{52.9}{\scriptsize$\pm$8.4} {\scriptsize(+12\%)} & \textbf{69.4}{\scriptsize$\pm$1.0} {\scriptsize(+2\%)} & 39.9{\scriptsize$\pm$3.8} {\scriptsize($-$1\%)} & \textit{68.0}{\scriptsize$\pm$8.8} {\scriptsize(+2\%)} & \textit{46.6}{\scriptsize$\pm$3.5} {\scriptsize($-$1\%)} & \textit{68.3}{\scriptsize$\pm$3.5} {\scriptsize(+1\%)} & \textbf{43.7}{\scriptsize$\pm$10.2} {\scriptsize(+12\%)} \\
CLIP only & 78.2{\scriptsize$\pm$11.4} {\scriptsize(+4\%)} & \textbf{58.3}{\scriptsize$\pm$15.3} {\scriptsize(+22\%)} & \textbf{72.6}{\scriptsize$\pm$4.8} {\scriptsize(+3\%)} & 44.4{\scriptsize$\pm$12.7} {\scriptsize($-$6\%)} & 66.1{\scriptsize$\pm$3.1} {\scriptsize($-$3\%)} & \textit{41.4}{\scriptsize$\pm$2.4} {\scriptsize(+3\%)} & 67.3{\scriptsize$\pm$8.3} {\scriptsize(+1\%)} & 46.2{\scriptsize$\pm$2.8} {\scriptsize($-$2\%)} & 67.8{\scriptsize$\pm$5.5} {\scriptsize(+0\%)} & 38.5{\scriptsize$\pm$9.2} {\scriptsize($-$1\%)} \\
\midrule
\multicolumn{11}{l}{Training strategy (modes + CLIP heads)} \\
\rowcolor{gray!15} onset\_pre (ours) & \textit{75.1}{\scriptsize$\pm$12.5} & 47.9{\scriptsize$\pm$20.9} & \textbf{70.5}{\scriptsize$\pm$10.4} & \textbf{47.4}{\scriptsize$\pm$13.5} & \textbf{67.9}{\scriptsize$\pm$5.0} & \textbf{40.2}{\scriptsize$\pm$1.2} & \textbf{67.0}{\scriptsize$\pm$6.6} & \textbf{47.0}{\scriptsize$\pm$0.6} & \textbf{67.8}{\scriptsize$\pm$4.8} & \textbf{39.0}{\scriptsize$\pm$7.0} \\
onset\_traj & \textbf{77.6}{\scriptsize$\pm$6.5} {\scriptsize(+3\%)} & \textbf{52.0}{\scriptsize$\pm$14.9} {\scriptsize(+9\%)} & \textit{64.5}{\scriptsize$\pm$2.0} {\scriptsize($-$9\%)} & \textit{43.8}{\scriptsize$\pm$10.9} {\scriptsize($-$8\%)} & \textit{49.7}{\scriptsize$\pm$2.4} {\scriptsize($-$27\%)} & 38.4{\scriptsize$\pm$2.9} {\scriptsize($-$5\%)} & \textit{48.0}{\scriptsize$\pm$6.8} {\scriptsize($-$28\%)} & 31.3{\scriptsize$\pm$3.1} {\scriptsize($-$33\%)} & \textit{52.1}{\scriptsize$\pm$4.3} {\scriptsize($-$23\%)} & 21.6{\scriptsize$\pm$3.9} {\scriptsize($-$45\%)} \\
onset & 73.3{\scriptsize$\pm$3.5} {\scriptsize($-$2\%)} & 45.4{\scriptsize$\pm$14.6} {\scriptsize($-$5\%)} & 60.3{\scriptsize$\pm$7.7} {\scriptsize($-$15\%)} & 39.5{\scriptsize$\pm$10.7} {\scriptsize($-$17\%)} & 48.8{\scriptsize$\pm$1.4} {\scriptsize($-$28\%)} & \textit{39.9}{\scriptsize$\pm$6.6} {\scriptsize($-$1\%)} & 47.6{\scriptsize$\pm$11.8} {\scriptsize($-$29\%)} & \textit{36.6}{\scriptsize$\pm$5.7} {\scriptsize($-$22\%)} & 46.4{\scriptsize$\pm$5.4} {\scriptsize($-$32\%)} & \textit{25.5}{\scriptsize$\pm$9.0} {\scriptsize($-$35\%)} \\
onset\_post & 73.5{\scriptsize$\pm$4.3} {\scriptsize($-$2\%)} & \textit{50.2}{\scriptsize$\pm$11.0} {\scriptsize(+5\%)} & 37.4{\scriptsize$\pm$3.0} {\scriptsize($-$47\%)} & 41.8{\scriptsize$\pm$11.8} {\scriptsize($-$12\%)} & 24.1{\scriptsize$\pm$2.4} {\scriptsize($-$65\%)} & 31.3{\scriptsize$\pm$0.9} {\scriptsize($-$22\%)} & 28.3{\scriptsize$\pm$3.6} {\scriptsize($-$58\%)} & 25.0{\scriptsize$\pm$0.3} {\scriptsize($-$47\%)} & 30.6{\scriptsize$\pm$2.5} {\scriptsize($-$55\%)} & 17.0{\scriptsize$\pm$3.4} {\scriptsize($-$56\%)} \\
\bottomrule
\end{tabular}%
}
\end{table*}

\begin{table*}[t]
\centering
\caption{What the CLIP head and the score term add to failure-mode prediction: top-1 and top-3 accuracy (in \%), mean $\pm$ standard deviation over the splits. Our probe is shaded. Best in bold, second best in italics, per column and anchor.}
\label{tab:mode_ablation}
\setlength{\tabcolsep}{4pt}
\resizebox{1.0\textwidth}{!}{%
\begin{tabular}{lrrrrrrrrrrrrrrrr}
\toprule
 & \multicolumn{8}{c}{Top-1 accuracy (\%)} & \multicolumn{8}{c}{Top-3 accuracy (\%)} \\
\cmidrule(lr){2-9}\cmidrule(lr){10-17}
 & \multicolumn{2}{c}{task} & \multicolumn{2}{c}{signature} & \multicolumn{2}{c}{skill} & \multicolumn{2}{c}{Real robot} & \multicolumn{2}{c}{task} & \multicolumn{2}{c}{signature} & \multicolumn{2}{c}{skill} & \multicolumn{2}{c}{Real robot} \\
\cmidrule(lr){2-3}\cmidrule(lr){4-5}\cmidrule(lr){6-7}\cmidrule(lr){8-9}\cmidrule(lr){10-11}\cmidrule(lr){12-13}\cmidrule(lr){14-15}\cmidrule(lr){16-17}
Method & Seen & Unseen & Seen & Unseen & Seen & Unseen & Seen & Unseen & Seen & Unseen & Seen & Unseen & Seen & Unseen & Seen & Unseen \\
\midrule
\multicolumn{17}{l}{At the first alarm of our probe} \\
\rowcolor{gray!15} Score + modes + CLIP (ours) & \textbf{41.0{\scriptsize$\pm$1.4}} & \textbf{27.2{\scriptsize$\pm$2.5}} & \textit{37.3{\scriptsize$\pm$7.2}} & \textit{23.7{\scriptsize$\pm$0.7}} & 42.3{\scriptsize$\pm$5.3} & \textbf{32.2{\scriptsize$\pm$4.3}} & \textbf{57.9{\scriptsize$\pm$10.6}} & \textit{61.5{\scriptsize$\pm$44.0}} & \textit{76.0{\scriptsize$\pm$2.6}} & \textbf{65.9{\scriptsize$\pm$5.7}} & \textit{74.2{\scriptsize$\pm$1.0}} & \textit{62.8{\scriptsize$\pm$1.8}} & 74.6{\scriptsize$\pm$3.4} & \textbf{63.2{\scriptsize$\pm$5.6}} & \textit{93.3{\scriptsize$\pm$7.3}} & \textbf{100.0{\scriptsize$\pm$0.0}} \\
Score + modes & \textit{40.6{\scriptsize$\pm$3.7}} & \textit{25.8{\scriptsize$\pm$2.5}} & \textbf{44.0{\scriptsize$\pm$4.6}} & \textbf{25.6{\scriptsize$\pm$0.8}} & \textit{42.5{\scriptsize$\pm$5.3}} & 28.0{\scriptsize$\pm$7.0} & \textit{57.4{\scriptsize$\pm$11.8}} & \textbf{64.1{\scriptsize$\pm$45.4}} & \textbf{76.4{\scriptsize$\pm$1.5}} & 64.0{\scriptsize$\pm$2.8} & \textbf{75.6{\scriptsize$\pm$2.8}} & \textbf{63.3{\scriptsize$\pm$1.0}} & \textbf{75.7{\scriptsize$\pm$2.9}} & \textit{61.9{\scriptsize$\pm$7.9}} & 91.9{\scriptsize$\pm$7.0} & \textbf{100.0{\scriptsize$\pm$0.0}} \\
Modes + CLIP & 37.3{\scriptsize$\pm$5.8} & 23.7{\scriptsize$\pm$1.4} & 29.3{\scriptsize$\pm$11.1} & 17.7{\scriptsize$\pm$9.3} & \textbf{44.0{\scriptsize$\pm$6.4}} & \textit{30.5{\scriptsize$\pm$5.3}} & 55.0{\scriptsize$\pm$14.2} & 59.0{\scriptsize$\pm$42.8} & 72.3{\scriptsize$\pm$7.7} & \textit{64.9{\scriptsize$\pm$3.3}} & 62.6{\scriptsize$\pm$16.4} & 54.3{\scriptsize$\pm$13.4} & \textit{75.3{\scriptsize$\pm$5.2}} & 61.0{\scriptsize$\pm$5.6} & 90.4{\scriptsize$\pm$7.9} & \textbf{100.0{\scriptsize$\pm$0.0}} \\
Modes & 36.5{\scriptsize$\pm$6.0} & 23.7{\scriptsize$\pm$3.2} & 35.7{\scriptsize$\pm$5.7} & 20.6{\scriptsize$\pm$4.8} & 40.9{\scriptsize$\pm$6.7} & 30.1{\scriptsize$\pm$4.7} & 53.8{\scriptsize$\pm$12.5} & 56.4{\scriptsize$\pm$41.8} & 71.9{\scriptsize$\pm$7.8} & 63.6{\scriptsize$\pm$1.4} & 72.2{\scriptsize$\pm$6.6} & 57.3{\scriptsize$\pm$7.3} & 71.5{\scriptsize$\pm$3.4} & 61.6{\scriptsize$\pm$8.7} & \textbf{94.7{\scriptsize$\pm$7.2}} & \textbf{100.0{\scriptsize$\pm$0.0}} \\
\midrule
\multicolumn{17}{l}{At the true failure onset (upper bound)} \\
\rowcolor{gray!15} Score + modes + CLIP (ours) & \textit{49.2{\scriptsize$\pm$5.7}} & \textbf{40.6{\scriptsize$\pm$4.5}} & \textit{47.7{\scriptsize$\pm$9.0}} & \textbf{40.9{\scriptsize$\pm$3.1}} & 48.6{\scriptsize$\pm$3.5} & 39.6{\scriptsize$\pm$8.0} & \textit{61.9{\scriptsize$\pm$8.4}} & 23.7{\scriptsize$\pm$22.0} & \textit{80.3{\scriptsize$\pm$3.6}} & \textbf{73.6{\scriptsize$\pm$2.6}} & \textbf{81.7{\scriptsize$\pm$5.8}} & \textbf{77.5{\scriptsize$\pm$1.5}} & 77.4{\scriptsize$\pm$4.0} & \textbf{74.7{\scriptsize$\pm$6.7}} & \textit{95.9{\scriptsize$\pm$4.6}} & 62.4{\scriptsize$\pm$35.1} \\
Score + modes & \textbf{54.7{\scriptsize$\pm$3.2}} & \textit{39.6{\scriptsize$\pm$5.0}} & \textbf{49.1{\scriptsize$\pm$4.5}} & \textit{39.0{\scriptsize$\pm$0.4}} & 48.4{\scriptsize$\pm$3.7} & \textit{40.4{\scriptsize$\pm$8.0}} & 59.4{\scriptsize$\pm$5.0} & \textit{25.7{\scriptsize$\pm$21.1}} & \textbf{82.5{\scriptsize$\pm$3.5}} & \textit{72.6{\scriptsize$\pm$0.7}} & \textit{79.6{\scriptsize$\pm$3.8}} & \textit{74.5{\scriptsize$\pm$1.7}} & \textbf{77.9{\scriptsize$\pm$3.0}} & \textit{73.4{\scriptsize$\pm$8.1}} & \textbf{96.4{\scriptsize$\pm$3.2}} & \textit{62.8{\scriptsize$\pm$33.3}} \\
Modes + CLIP & 42.4{\scriptsize$\pm$5.3} & 37.0{\scriptsize$\pm$7.7} & 33.7{\scriptsize$\pm$15.9} & 31.2{\scriptsize$\pm$17.2} & \textit{49.4{\scriptsize$\pm$5.8}} & 39.8{\scriptsize$\pm$6.4} & \textbf{62.1{\scriptsize$\pm$8.8}} & 19.5{\scriptsize$\pm$19.8} & 77.3{\scriptsize$\pm$10.2} & 69.1{\scriptsize$\pm$5.3} & 69.2{\scriptsize$\pm$15.6} & 61.4{\scriptsize$\pm$23.2} & \textit{77.5{\scriptsize$\pm$4.7}} & 72.2{\scriptsize$\pm$5.9} & \textbf{96.4{\scriptsize$\pm$3.9}} & 59.5{\scriptsize$\pm$32.9} \\
Modes & 48.9{\scriptsize$\pm$5.6} & 38.1{\scriptsize$\pm$8.7} & 42.7{\scriptsize$\pm$7.2} & 35.0{\scriptsize$\pm$12.0} & \textbf{49.8{\scriptsize$\pm$4.9}} & \textbf{40.8{\scriptsize$\pm$5.5}} & 59.4{\scriptsize$\pm$7.6} & \textbf{32.4{\scriptsize$\pm$26.3}} & 76.4{\scriptsize$\pm$7.4} & 68.8{\scriptsize$\pm$5.4} & 74.2{\scriptsize$\pm$5.3} & 67.4{\scriptsize$\pm$13.2} & 77.4{\scriptsize$\pm$4.0} & 73.2{\scriptsize$\pm$7.5} & \textbf{96.4{\scriptsize$\pm$3.3}} & \textbf{66.1{\scriptsize$\pm$34.8}} \\
\bottomrule
\end{tabular}%
}
\end{table*}

\begin{table*}[t]
\centering
\caption{Early-detection area (in \%) on the unseen episodes when the failure time is moved later by 0, 1, and 2\,s; nothing is retrained. Ours is shaded. Best in bold, second best in italics.}
\label{tab:onset_shift}
\setlength{\tabcolsep}{4pt}
\resizebox{\textwidth}{!}{%
\begin{tabular}{lrrrrrrrrrrrrrrrrrr}
\toprule
 & \multicolumn{3}{c}{LIBERO-10 $\pi_{0.5}$} & \multicolumn{3}{c}{LIBERO-10 GR00T} & \multicolumn{3}{c}{LIBERO-X task} & \multicolumn{3}{c}{LIBERO-X signature} & \multicolumn{3}{c}{LIBERO-X skill} & \multicolumn{3}{c}{Real robot} \\
\cmidrule(lr){2-4}\cmidrule(lr){5-7}\cmidrule(lr){8-10}\cmidrule(lr){11-13}\cmidrule(lr){14-16}\cmidrule(lr){17-19}
Method & 0\,s & +1\,s & +2\,s & 0\,s & +1\,s & +2\,s & 0\,s & +1\,s & +2\,s & 0\,s & +1\,s & +2\,s & 0\,s & +1\,s & +2\,s & 0\,s & +1\,s & +2\,s \\
\midrule
SAFE-LSTM & 34.9 & 37.2 & 39.2 & \textbf{51.2} & \textbf{53.1} & \textbf{55.0} & \textit{40.4} & \textit{42.1} & \textit{44.4} & 41.2 & 42.8 & 44.9 & \textit{30.1} & \textit{31.7} & \textit{33.6} & \textit{25.1} & \textbf{26.8} & \textbf{28.9} \\
SAFE-MLP & 39.9 & 42.1 & 43.9 & 39.2 & 40.6 & 43.9 & \textbf{42.3} & \textbf{43.3} & \textbf{44.7} & \textit{46.2} & \textit{47.2} & \textit{47.9} & 23.5 & 24.3 & 25.7 & 20.2 & 21.4 & 22.8 \\
STAC & 36.4 & \textit{43.0} & \textit{53.2} & 17.5 & 22.0 & 27.6 & 28.0 & 29.5 & 32.2 & 28.7 & 30.6 & 34.0 & 21.5 & 22.6 & 25.1 & -- & -- & -- \\
ActProbe & 26.6 & 31.0 & 37.2 & 30.1 & 33.3 & 38.6 & 22.7 & 24.7 & 27.5 & 13.9 & 16.1 & 18.8 & 13.6 & 14.3 & 15.7 & -- & -- & -- \\
FailureSpot & 36.2 & 39.4 & 42.7 & 44.7 & 46.5 & 46.6 & 20.4 & 20.9 & 21.7 & 26.2 & 27.2 & 29.1 & 19.6 & 20.8 & 22.8 & -- & -- & -- \\
Mahalanobis & \textit{40.2} & 42.1 & 45.9 & 29.2 & 30.4 & 32.6 & 30.6 & 31.7 & 34.0 & 22.3 & 23.1 & 24.4 & 15.9 & 16.6 & 17.4 & 19.6 & 19.7 & 19.8 \\
Euclidean & 24.2 & 28.2 & 31.2 & 23.1 & 25.4 & 30.4 & 28.5 & 29.7 & 31.1 & 19.7 & 20.8 & 22.5 & 19.0 & 20.0 & 21.4 & 8.3 & 9.7 & 12.2 \\
PCA-KMeans & 12.9 & 12.9 & 13.8 & 20.0 & 20.0 & 20.0 & 25.0 & 25.3 & 25.9 & 32.5 & 34.0 & 36.4 & 20.7 & 21.5 & 23.1 & 19.4 & 19.9 & 20.3 \\
\midrule
\rowcolor{gray!15} \ours (Ours) & \textbf{47.9} & \textbf{52.5} & \textbf{55.4} & \textit{47.4} & \textit{49.1} & \textit{49.3} & 40.2 & 41.3 & 42.3 & \textbf{47.0} & \textbf{48.2} & \textbf{49.1} & \textbf{39.0} & \textbf{40.5} & \textbf{41.4} & \textbf{25.8} & \textit{26.3} & \textit{26.5} \\
\bottomrule
\end{tabular}%
}
\end{table*}

\section{Qualitative Results}
\label{app:qualitative}

\subsection{Latent Space}
\label{app:latent}

\ours{} encodes the VLA backbone embedding of each step with an LSTM into a latent state, which three MLP heads read: the failure score, the failure mode, and a CLIP embedding. The following figures show four of these spaces in two dimensions: the backbone embedding itself (the input), the LSTM state, and the hidden layers of the failure score head and of the failure mode head. Each is projected onto its first two principal components, fitted on the pre-failure steps of the training episodes only, and drawn for the training episodes and for the unseen tasks of the LIBERO-X \newtask{} split. The points are the steps up to the failure of failed episodes and all steps of successful episodes; the ellipses are the $1\sigma$ ellipses of a Gaussian fitted to each class, drawn over the points. The classes are the outcome (success in blue, failure in red) or the failure mode. The mode head is trained on failures only and has only a mode plot.

\begin{figure*}[p]
    \centering
    \includegraphics[width=0.9\textwidth]{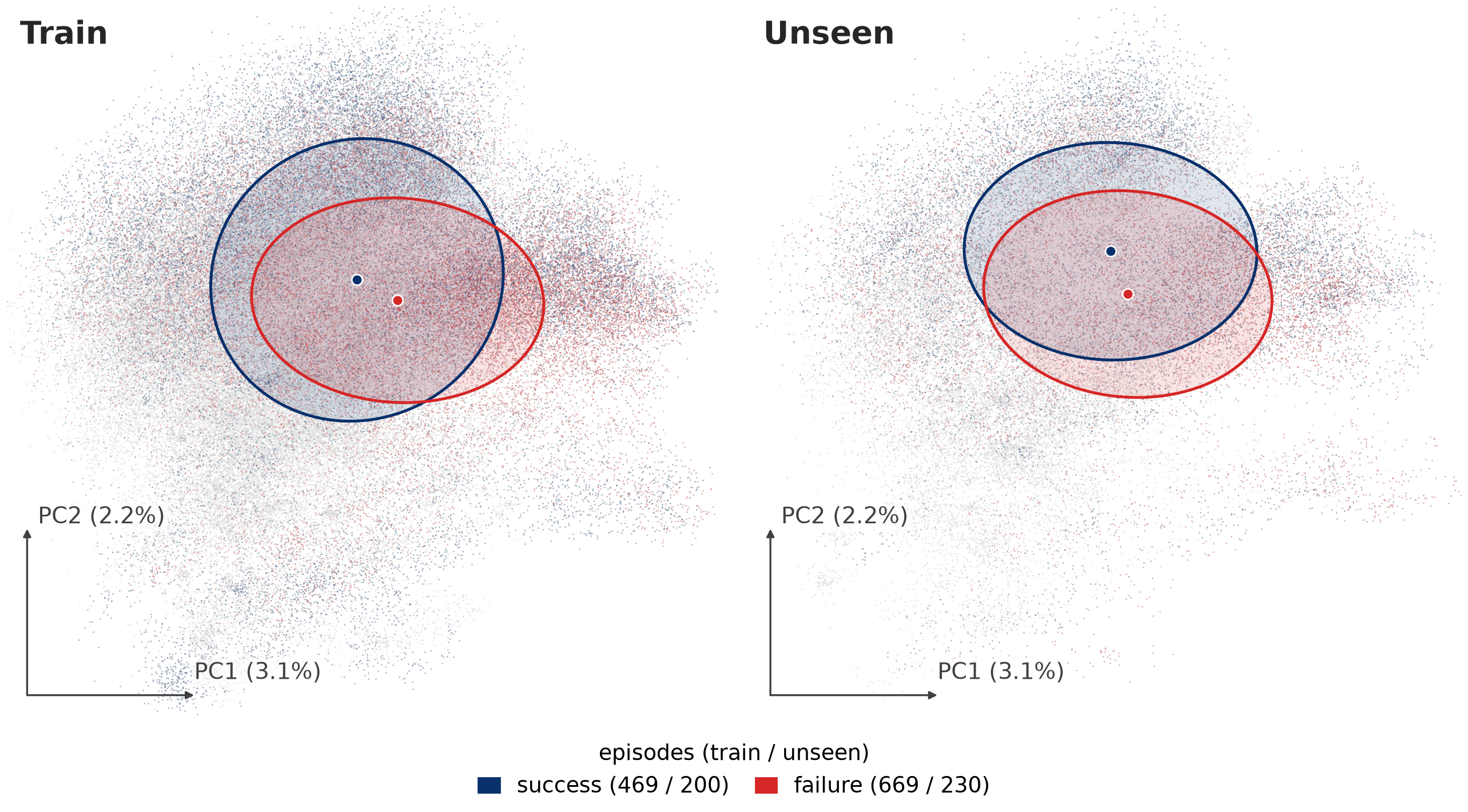}
    \caption{VLA backbone embedding, by outcome.}
    \label{fig:latent_vla_outcome}
\end{figure*}

\begin{figure*}[p]
    \centering
    \includegraphics[width=0.9\textwidth]{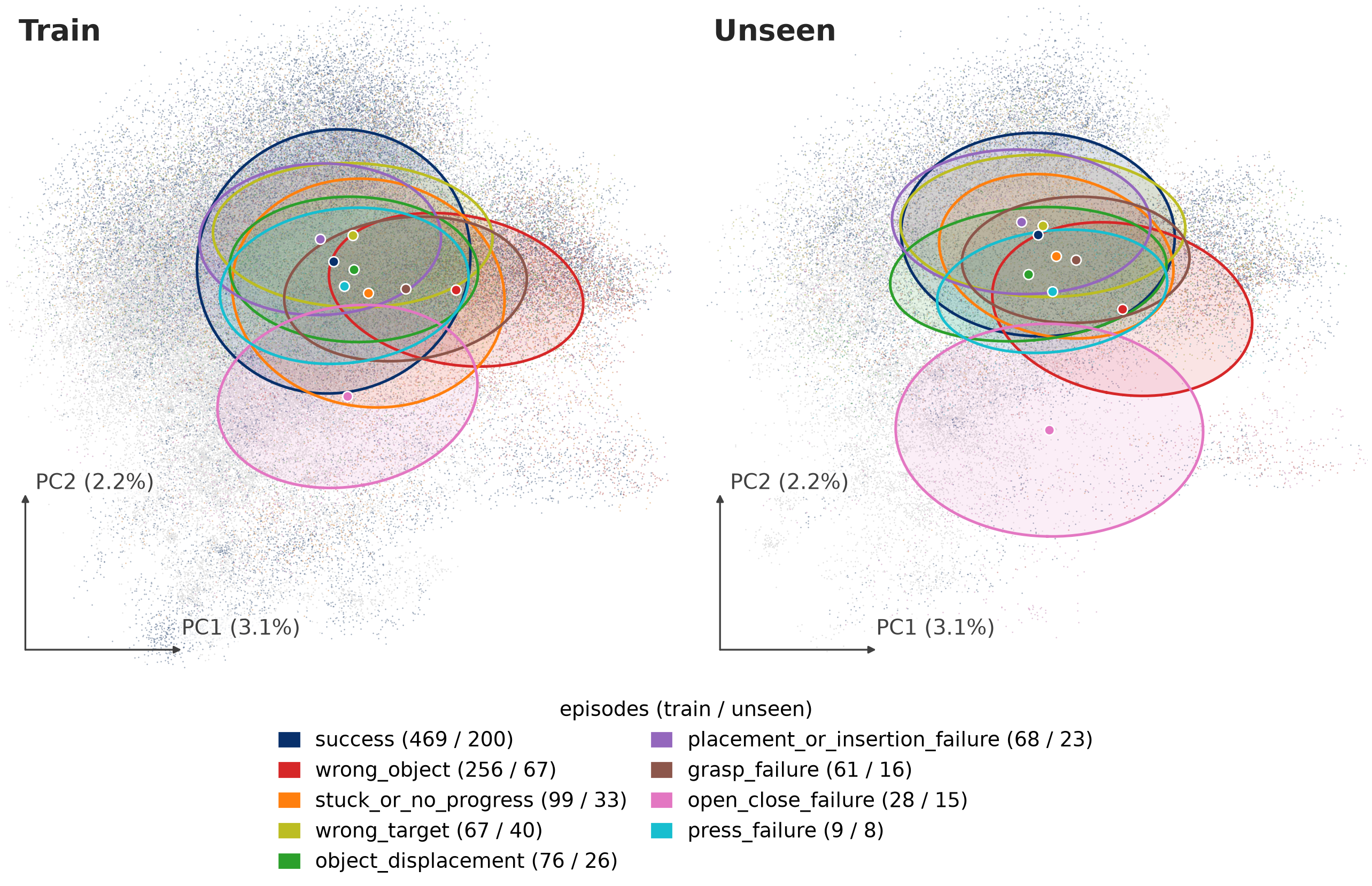}
    \caption{VLA backbone embedding, by failure mode.}
    \label{fig:latent_vla_mode}
\end{figure*}
\clearpage

\begin{figure*}[p]
    \centering
    \includegraphics[width=0.9\textwidth]{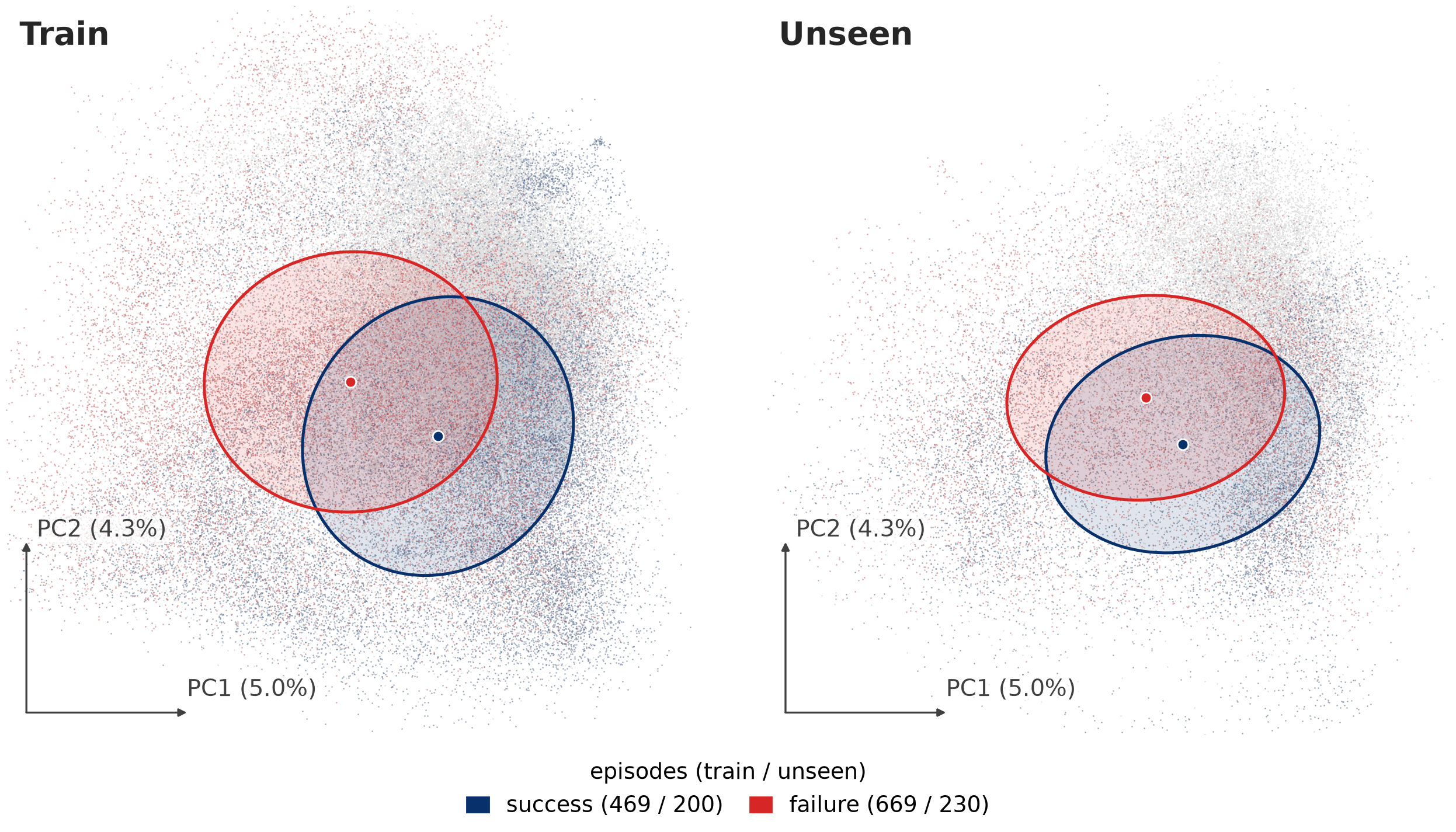}
    \caption{Probe LSTM state, by outcome.}
    \label{fig:latent_lstm_outcome}
\end{figure*}

\begin{figure*}[p]
    \centering
    \includegraphics[width=0.9\textwidth]{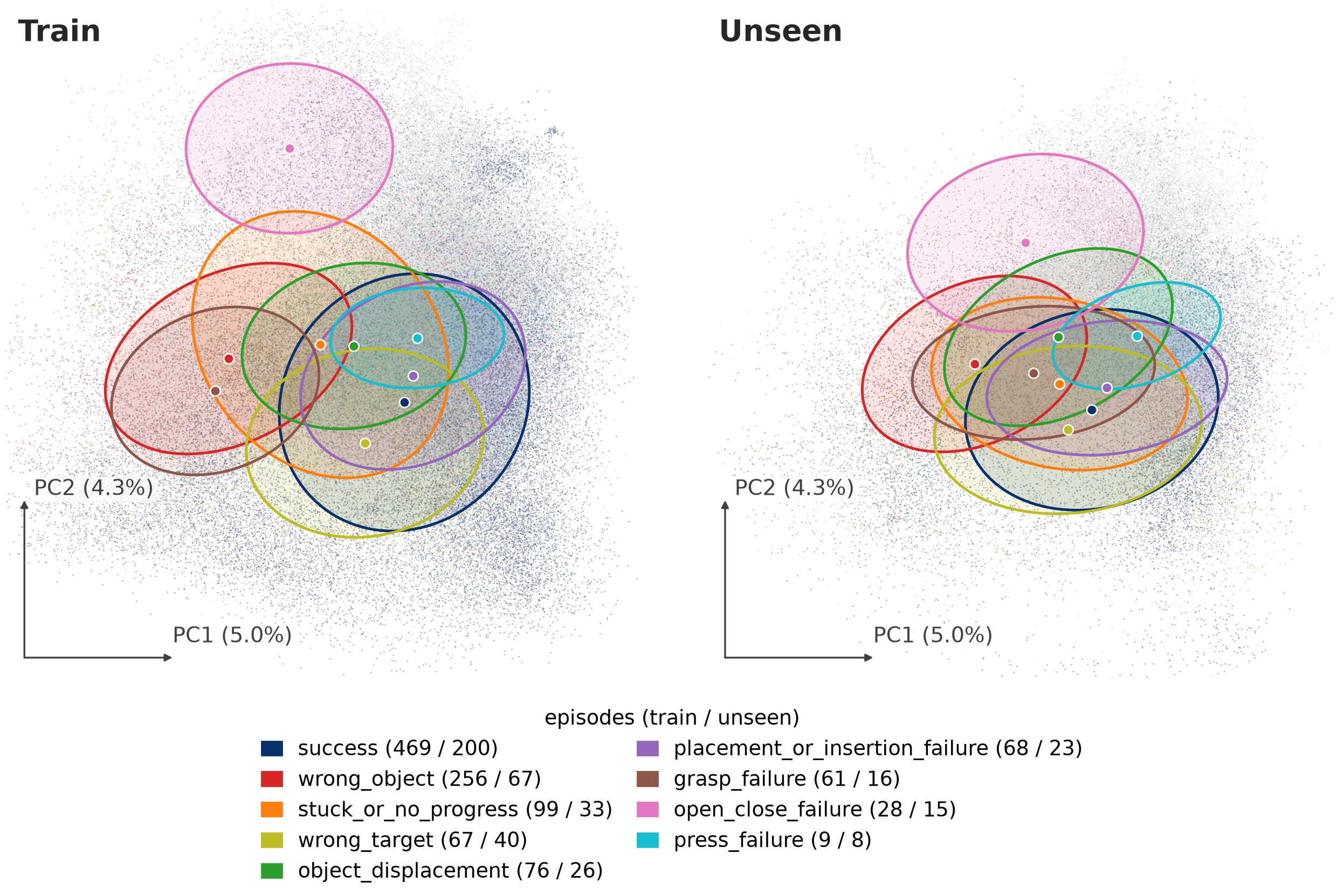}
    \caption{Probe LSTM state, by failure mode.}
    \label{fig:latent_lstm_mode}
\end{figure*}
\clearpage

\begin{figure*}[p]
    \centering
    \includegraphics[width=0.9\textwidth]{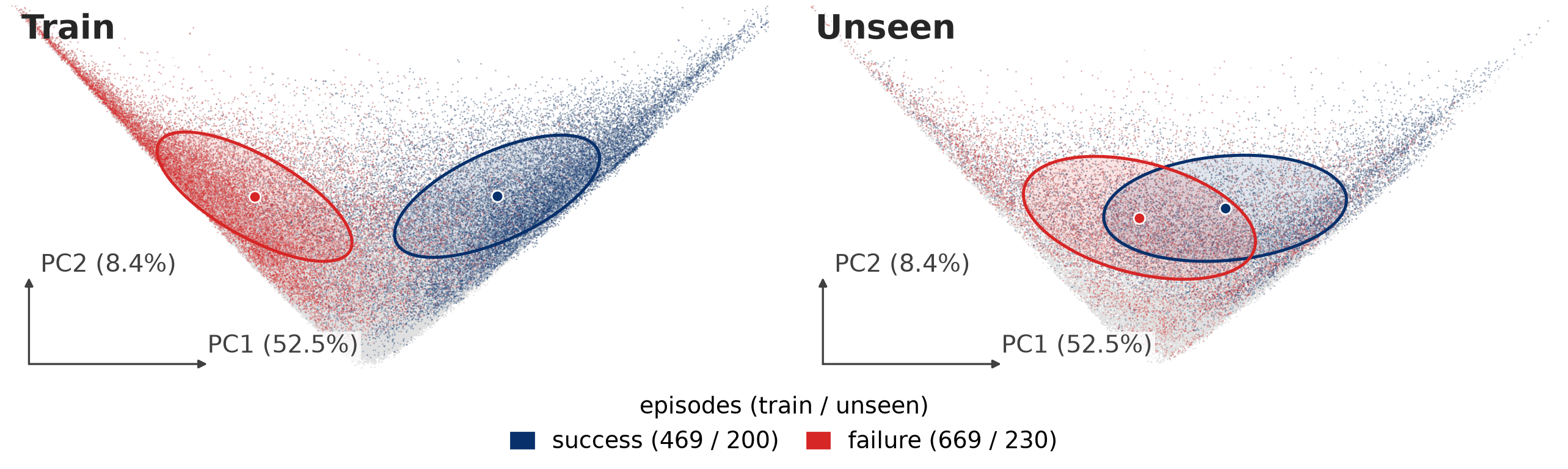}
    \caption{Failure score head layer, by outcome.}
    \label{fig:latent_score_outcome}
\end{figure*}

\begin{figure*}[p]
    \centering
    \includegraphics[width=0.9\textwidth]{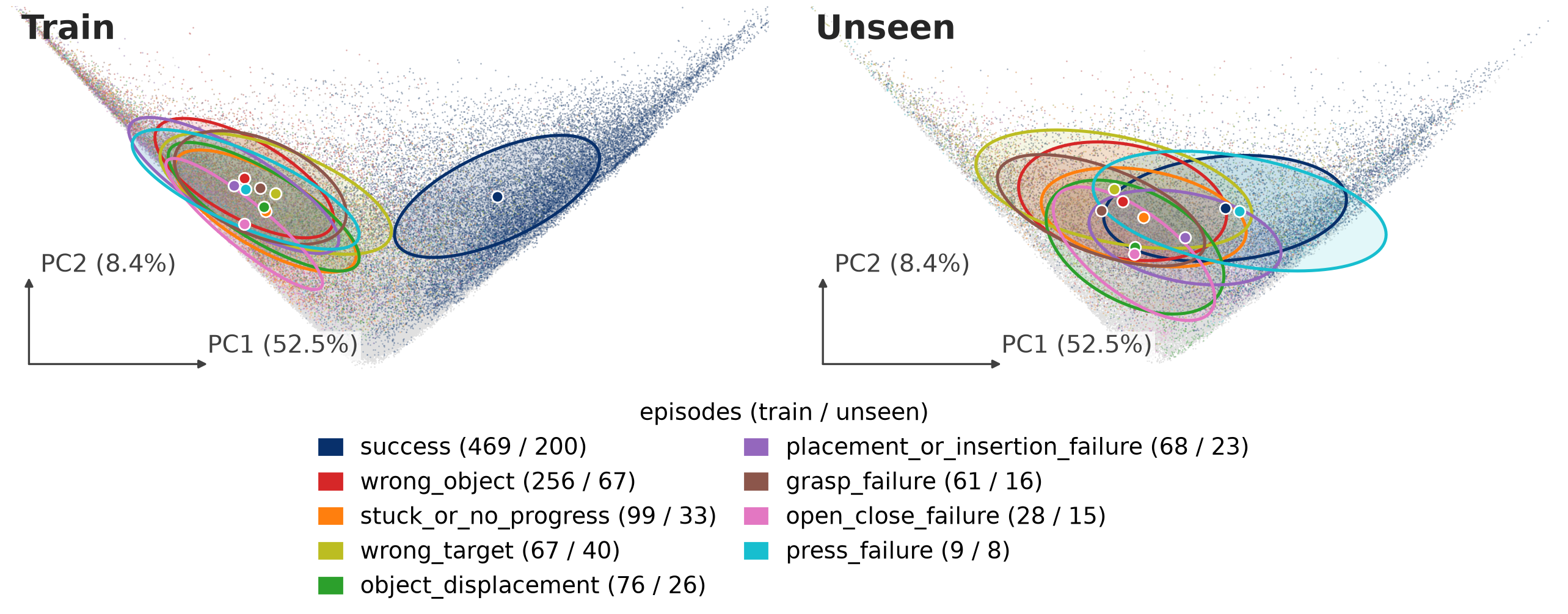}
    \caption{Failure score head layer, by failure mode.}
    \label{fig:latent_score_mode}
\end{figure*}
\clearpage

\begin{figure*}[p]
    \centering
    \includegraphics[width=0.9\textwidth]{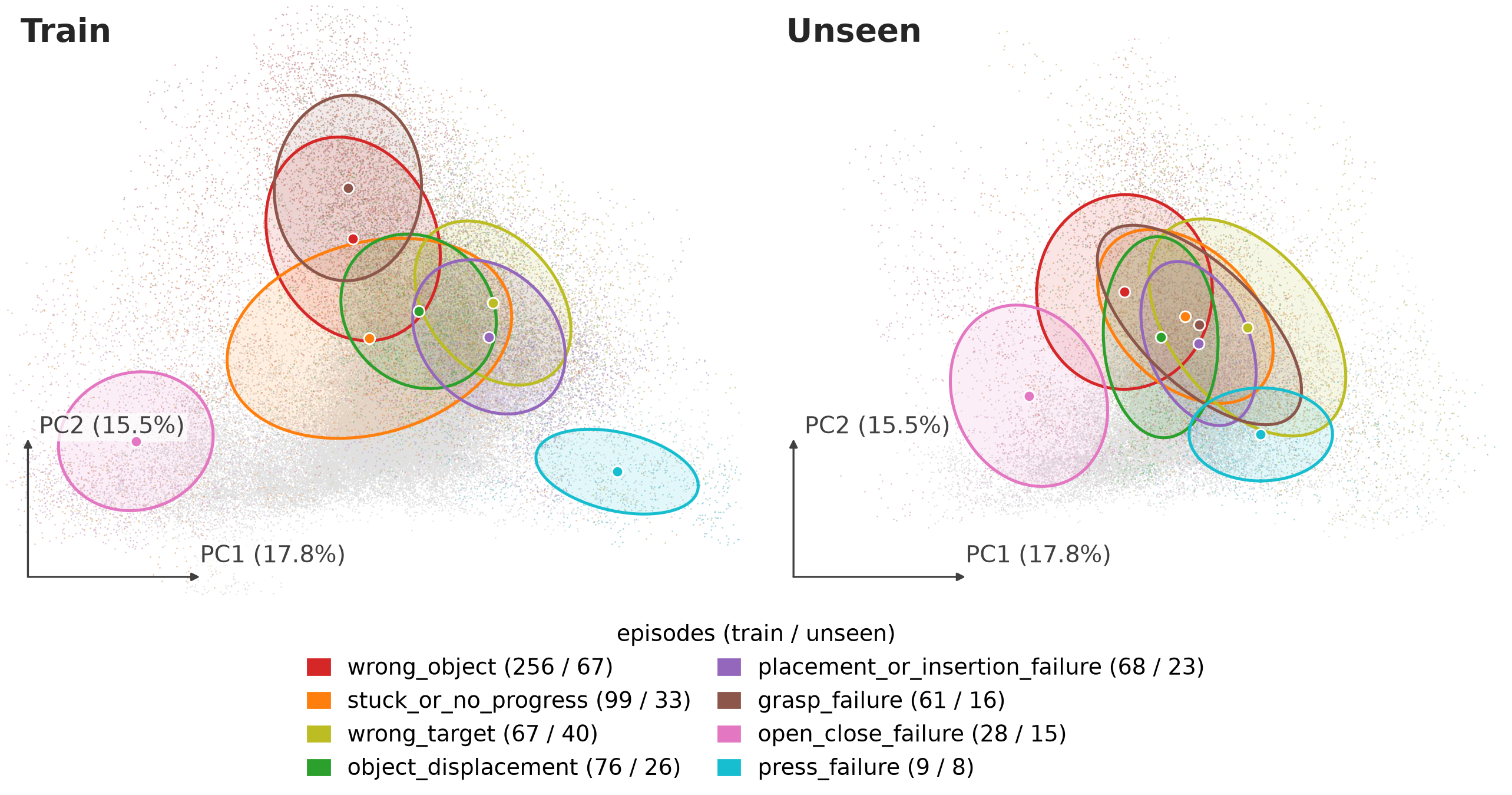}
    \caption{Failure mode head layer, by failure mode (failures only).}
    \label{fig:latent_modehead_mode}
\end{figure*}

\subsection{Latent Trajectories}
\label{app:trajectories}

\begin{figure*}[p]
    \centering
    \includegraphics[width=0.9\linewidth]{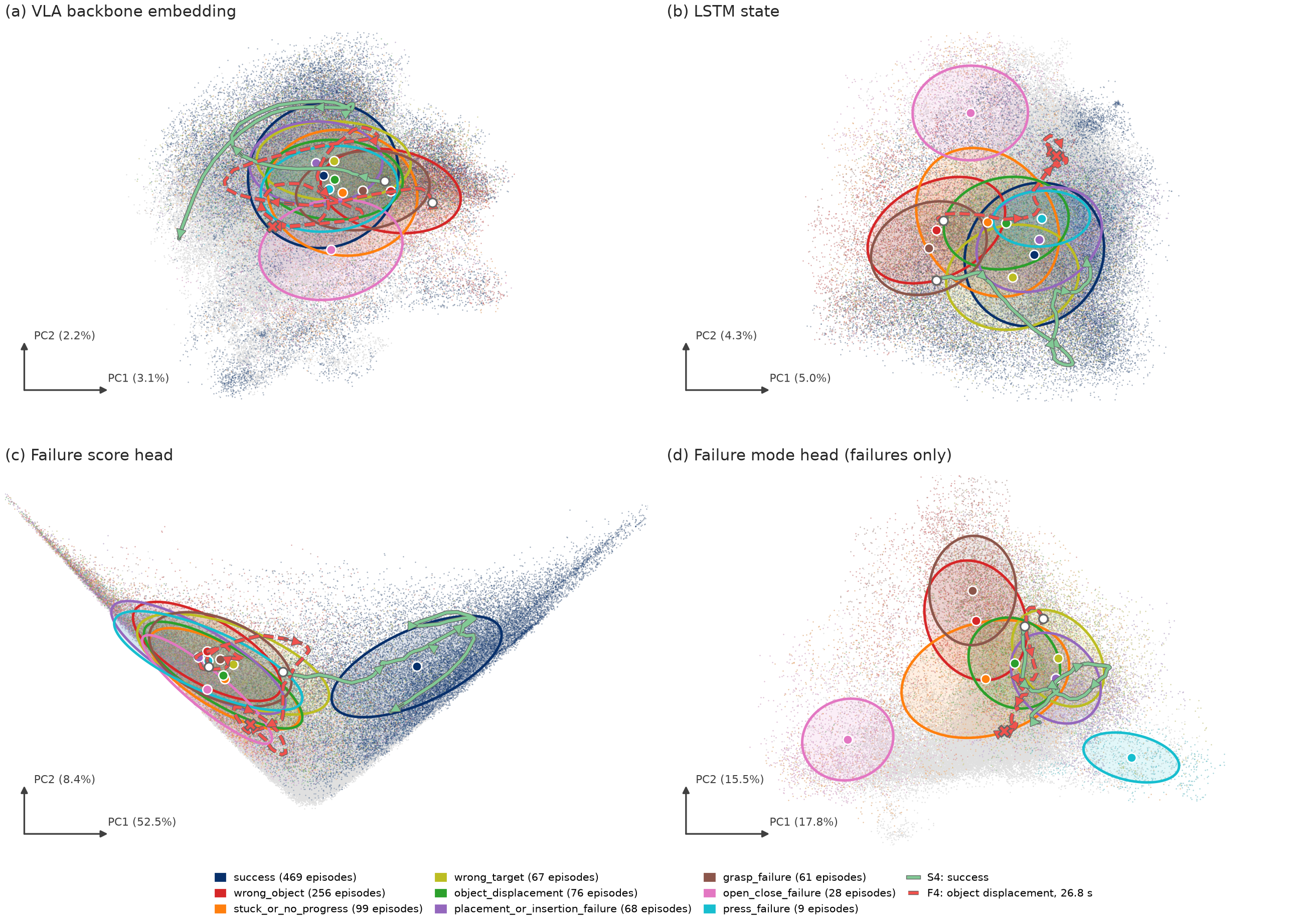}
    \caption{Seen tasks (training episodes). A success (green, ``pick up the bowl on the plate and place it to the left of the bowl drainer'') and an object displacement failure (red, ``place the striped book in the top region of the shelf'').}
    \label{fig:s4_f4}
\end{figure*}

\begin{figure*}[p]
    \centering
    \includegraphics[width=0.9\linewidth]{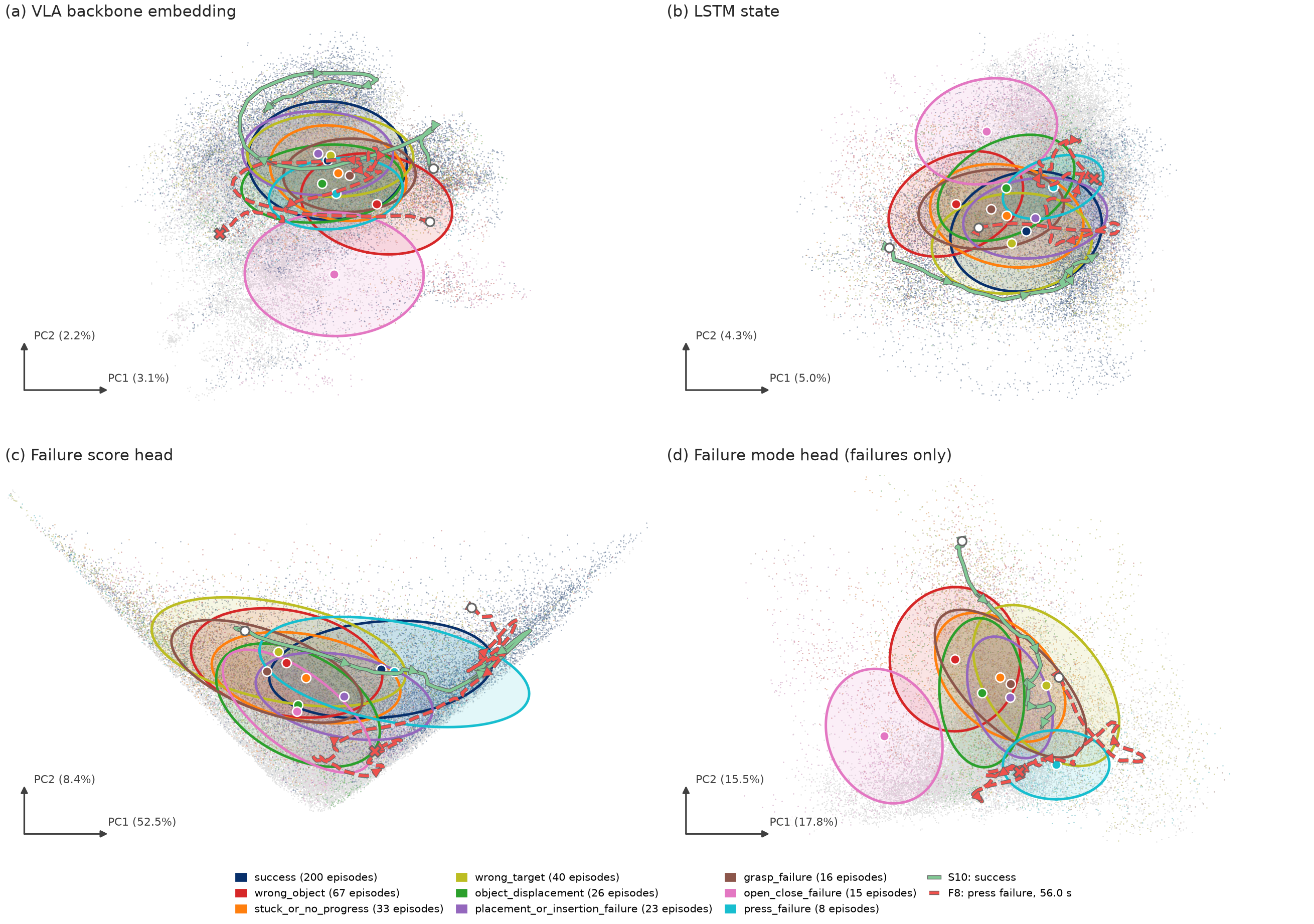}
    \caption{Unseen tasks. A success (green, ``place the red bowl in the basket'') and a press failure (red, ``turn on the flat stove'').}
    \label{fig:s10_f8}
\end{figure*}

To see how an episode moves through these spaces, \autoref{fig:s4_f4} and \autoref{fig:s10_f8} follow one success and one failure step by step, for episodes of seen and of unseen tasks. Each panel shows one space in its first two principal components: (a) the VLA backbone embedding, (b) the probe's LSTM state, and the hidden layers of (c) the failure score head and (d) the failure mode head. The points and ellipses are those of the episodes of the set shown (the training episodes, or the unseen tasks), grouped by outcome and failure mode, and the mode head is shown on failures only. The principal components are always those fitted on the training episodes. A trajectory starts at the ring; a failure is dashed and ends at the cross, at its failure step.

\subsection{Steering Demonstrations}
\label{app:steering}

We use steering only to show how the signal of \ours{} could be used; the heuristics are simple and are not evaluated. When the score crosses the calibrated threshold, the policy is paused, an intervention chosen from the predicted failure mode is applied, and the policy resumes.

\begin{figure*}[htbp]
    \centering
    \begin{subfigure}[b]{0.9\linewidth}
        \centering
        \includegraphics[width=\linewidth]{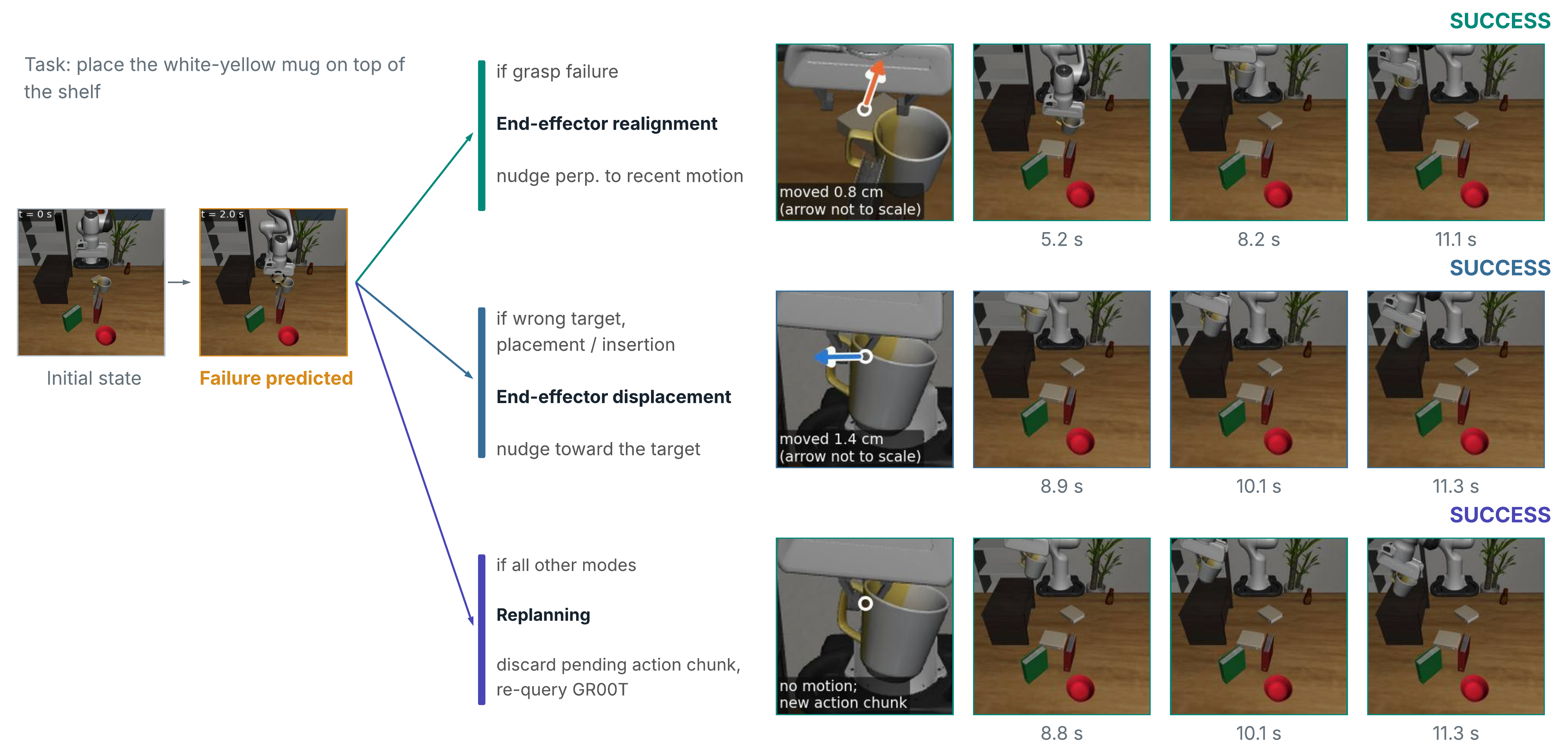}
        \caption{Simulation: ``place the white-yellow mug on top of the shelf.''}
        \label{fig:steering_strat}
    \end{subfigure}
    \hfill
    \begin{subfigure}[b]{0.9\linewidth}
        \centering
        \includegraphics[width=\linewidth]{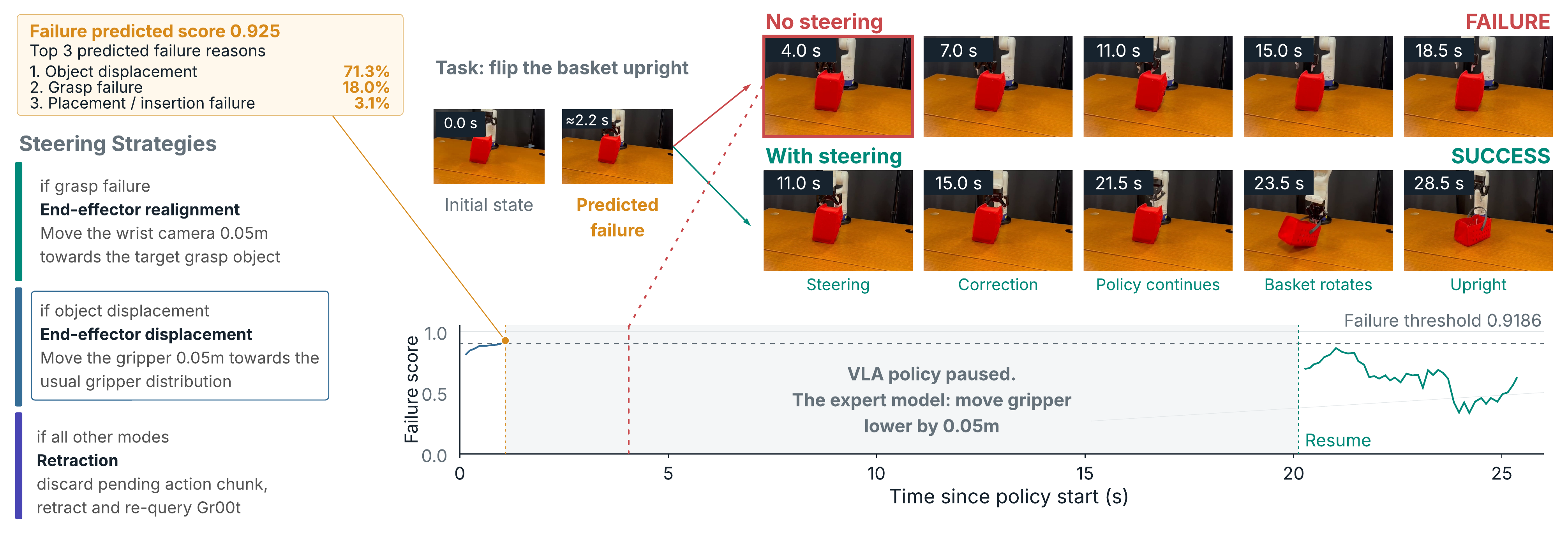}
        \caption{Real robot: ``flip the basket upright.''}
        \label{fig:basket_steering}
    \end{subfigure}
    \caption{Steering strategies, in simulation and on the real robot.}
    \label{fig:combined_steering_strat}
\end{figure*}

\paragraph{Simulation.} We implemented three heuristics for LIBERO-X. For a wrong-target or placement/insertion failure, the end effector is displaced by a small distance towards the destination. For a grasp failure, it is displaced laterally, perpendicular to its recent motion. For every other predicted mode, the policy replans from the current state. \autoref{fig:steering_strat} shows the three on the task ``place the white-yellow mug on top of the shelf'': after the alarm, the probe predicts the failure mode, the matching intervention is applied, and the policy resumes and succeeds, whereas without intervention the mug is placed on the table and the task fails. 

\paragraph{Real robot.} In the task ``put the mushroom on the plate'', the gripper misses the mushroom and hits the table without the probe; the probe predicts a grasp failure, and an expert would steer the wrist camera towards the grasp target and resume the policy. In the task ``flip the basket upright'', we start with the basket standing on its tall end, outside the training demonstrations; an expert might move the gripper along the $-z$ direction, so that the joint states are in the distribution of the flip motion, as the demonstrations are collected with the basket on its long side (\autoref{fig:basket_steering}).

\end{document}